\documentclass{article}

\usepackage{microtype}
\usepackage{graphicx}
\usepackage{subcaption} 
\usepackage{booktabs} 
\usepackage{cuted}
\usepackage{hyperref}
\usepackage{listings}
\usepackage{tcolorbox}
\tcbuselibrary{breakable}
\usepackage{tabularx}
\usepackage{svg}
\usepackage{multirow} 
\usepackage{pifont}
\usepackage[framemethod=TikZ]{mdframed}
\usepackage{color}
\usepackage{soul}
 
\usepackage[preprint]{icml2026}

\usepackage{amsmath}
\usepackage{amssymb}
\usepackage{mathtools}
\usepackage{amsthm}

\usepackage[capitalize,noabbrev]{cleveref}

\theoremstyle{plain}

\theoremstyle{definition}

\theoremstyle{remark}

\usepackage[disable,textsize=tiny]{todonotes}

\begin{document}

\twocolumn[
  \icmltitle{WideSeek: Advancing Wide Research via Multi-Agent Scaling}



  \icmlsetsymbol{equal}{*}

  \begin{icmlauthorlist}
    \icmlauthor{Ziyang Huang}{equal,casia,ucas}
    \icmlauthor{Haolin Ren}{equal,casia,ucas}
    \icmlauthor{Xiaowei Yuan}{casia,ucas}
    \icmlauthor{Jiawei Wang}{ustc}
    \icmlauthor{Zhongtao Jiang}{}
    \icmlauthor{Kun Xu}{}
    \icmlauthor{Shizhu He}{casia,ucas}
    \icmlauthor{Jun Zhao}{casia,ucas}
    \icmlauthor{Kang Liu}{casia,ucas}
  \end{icmlauthorlist}
  \begin{center}
  \url{https://wideseek-ai.github.io}
  \end{center}
  \icmlaffiliation{casia}{Institute of Automation, Chinese Academy of Sciences}
  \icmlaffiliation{ucas}{University of Chinese Academy of Sciences}
  \icmlaffiliation{ustc}{University of Science and Technology of China}

  \icmlcorrespondingauthor{Kang Liu}


  \vskip 0.3in
]



\printAffiliationsAndNotice{$^*$Equal contribution.}  

\begin{abstract}
Search intelligence is evolving from Deep Research to Wide Research, a paradigm essential for retrieving and synthesizing comprehensive information under complex constraints in parallel.
However, progress in this field is impeded by the lack of dedicated benchmarks and optimization methodologies for search breadth.
To address these challenges, we take a deep dive into Wide Research from two perspectives: \textbf{Data Pipeline} and \textbf{Agent Optimization}.
First, we produce WideSeekBench, a General Broad Information Seeking (GBIS) benchmark constructed via a rigorous multi-phase data pipeline to ensure diversity across the target information volume, logical constraints, and domains.
Second, we introduce WideSeek, a dynamic hierarchical multi-agent architecture that can autonomously fork parallel sub-agents based on task requirements.
Furthermore, we design a unified training framework that linearizes multi-agent trajectories and optimizes the system using end-to-end RL.
Experimental results demonstrate the effectiveness of WideSeek and multi-agent RL, highlighting that scaling the number of agents is a promising direction for advancing the Wide Research paradigm.
\end{abstract}
\begin{figure}[t]
    \centering
    \includegraphics[width=\linewidth]{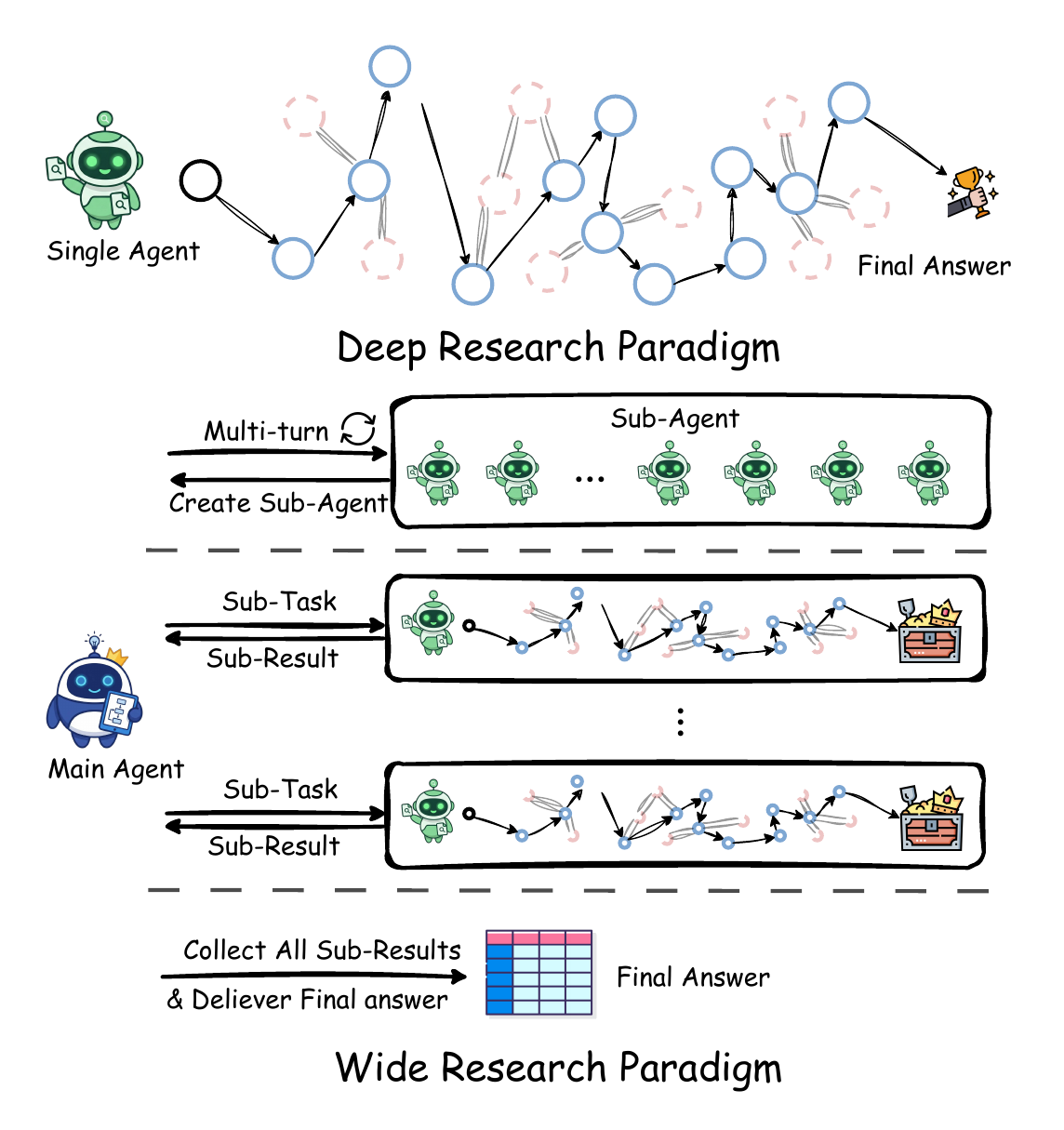}
    \caption{Deep Research paradigm vs. Wide Research paradigm.}
    \label{fig:intro}
\end{figure}
\section{Introduction}

Search Intelligence constitutes the cornerstone of Agentic AI \cite{shi2025deepresearchsystematicsurvey, Abou_Ali_2025}. Moving beyond a mere substitute for conventional search engines, it serves as an essential module for complex, real-world applications, including repository-level code generation \cite{jimenez2024swebench}, enterprise data intelligence \cite{lei2025dacompbenchmarkingdataagents}, and general GUI manipulation \cite{xie2024osworld}. 

Existing research has predominantly focused on \textbf{Deep Research} \cite{wei2025browsecompsimplechallengingbenchmark}, which employs complex, multi-step reasoning and action sequences to locate a single hard-to-find piece of information. 
As AI enters its \textit{Second Half}~\cite{yao2025second}, the research community is increasingly shifting its focus toward real-world and utility scenarios. This transition necessitates a move toward \textbf{Wide Research}~\cite{manus2025wide}, as shown in Figure~\ref{fig:intro}, which replaces sequential reasoning with a parallel orchestration paradigm. By prioritizing high-breadth synthesis and structural comprehensiveness, Wide Research enhances productivity and scales the effectiveness of industrial AI deployment.


Wide Research focuses on systematic retrieval across expansive search spaces, transitioning from deep-but-narrow chains to high-breadth parallelized frameworks. Aligning with Kimi Agent-Swarm~\cite{moonshot2026kimi25}, this paradigm employs a sophisticated orchestrator to decompose complex global objectives into granular, parallel sub-tasks, which are then concurrently executed by autonomous agents capable of iterative deep research and mutual cross-validation. A representative application is the generation of Competitor Analysis Tables, as exemplified by systems such as Manus \cite{manus2025wide}, which synthesize information from thousands of sources into comprehensive comparative tables, substantially reducing labor costs of Human Data Analyst while enhancing productivity at scale.


Despite its promise, the advancement of Wide Research is hindered by three primary challenges: (1) \textbf{Limitations in Benchmarks:} Existing benchmarks \cite{wong2025widesearchbenchmarkingagenticbroad, lan2025deepwidesearchbenchmarkingdepthwidth} are largely constructed by human experts, which limits their scale, diversity, and categorization depth.
Furthermore, they typically provide only test sets, lacking the training data necessary for model optimization; (2) \textbf{Deficiencies in Data Synthesis:} Current data synthesis methods for search agents focus on sampling complex graph topologies to simulate multi-step reasoning paths \cite{li2025websailornavigatingsuperhumanreasoning, tao2025webshaperagenticallydatasynthesizing}. While these approaches effectively optimize for search depth, they lack the capacity to efficiently synthesize a large scale of atomic information under complex constraint, which is critical for search width; and (3) \textbf{Optimization Gaps:} Previous approaches often rely on closed-source models within static multi-agent frameworks \cite{smolagents}, or concentrates on enhancing the depth of single-agent reasoning \cite{lu2025deepdiveadvancingdeepsearch}. There is a notable lack of exploration into the end-to-end optimization of systems capable of autonomously broadening their search paths.
To address these challenges, we investigate the Wide Research paradigm through two perspectives: data pipeline construction and agent optimization. 

\textbf{Data Pipeline \& Benchmark.}
While conventional methods construct information graphs from web pages to emulate reasoning paths toward a single answer, our approach utilizes large-scale Knowledge Graphs (KGs) \cite{schmelzeisen2021wikidated10evolvingknowledge} to extract clusters of interconnected world knowledge. Specifically, we initialize the process with seed entities and a set of sampled seed constraints. By applying formal set operations (including intersection, union, and difference), we construct complex constraints that resolve into a target entity set. Simultaneously, we sample high-coverage attributes of these entities to define the target attribute set. Next, we fetch all atomic information from Knowledge Graph to form the answer table and construct the input task based on the complex constraints. For convenient evaluation, this pipeline produces column-wise rubrics for reward system. To ensure the quality of data, all tasks will be evaluated by a hybrid filtering system.

Based on this pipeline, we introduce \textbf{WideSeekBench}, a benchmark for General Broad Information Seeking (GBIS) comprising both training and test sets. To ensure rigorous and multi-dimensional evaluation, the test set is strictly sampled and balanced across target information volume, operator complexity, and domains. 

\textbf{Agent Optimization.}
The Wide Research paradigm requires agents to acquire and synthesize target information from a large volume of sources. This necessitates a reasoning architecture that supports both parallel and serial execution, typically involving ultra-long-context reasoning and extensive tool invocation. To expand the search scope, enable robust cross-validation, and reduce execution complexity, we propose \textbf{WideSeek}, a system built on a dynamic multi-agent architecture. Following a \textit{Planner-Executor} pattern, the main agent is responsible for planning, task decomposition, and self-reflection, while sub-agents reason and execute tool calls to complete the sub-task. In contrast to previous methods that pre-define the roles and quantity of agents, which often degenerate into rigid workflows, WideSeek empowers the main agent with complete autonomy. It allows the system to dynamically instantiate any number of sub-agents at any step based on task requirements. Building on this flexible architecture, we collect all trajectories of the main agent and sub-agents and linearize them into a unified trajectory. Based on this, we optimize the system using end-to-end Reinforcement Learning (RL).

In conclusion, our experiments and analysis demonstrate that the transition from Deep to Wide Research requires a fundamental shift in agentic design, transitioning from sequential to dynamic, parallel orchestration. Moreover, our work not only establishes a rigorous benchmark for the field but also provides compelling evidence that specialized end-to-end multi-agent optimization can enable models to search at scale in complex scenarios. 
\begin{figure*}[th]
    \centering
    \includegraphics[width=\linewidth]{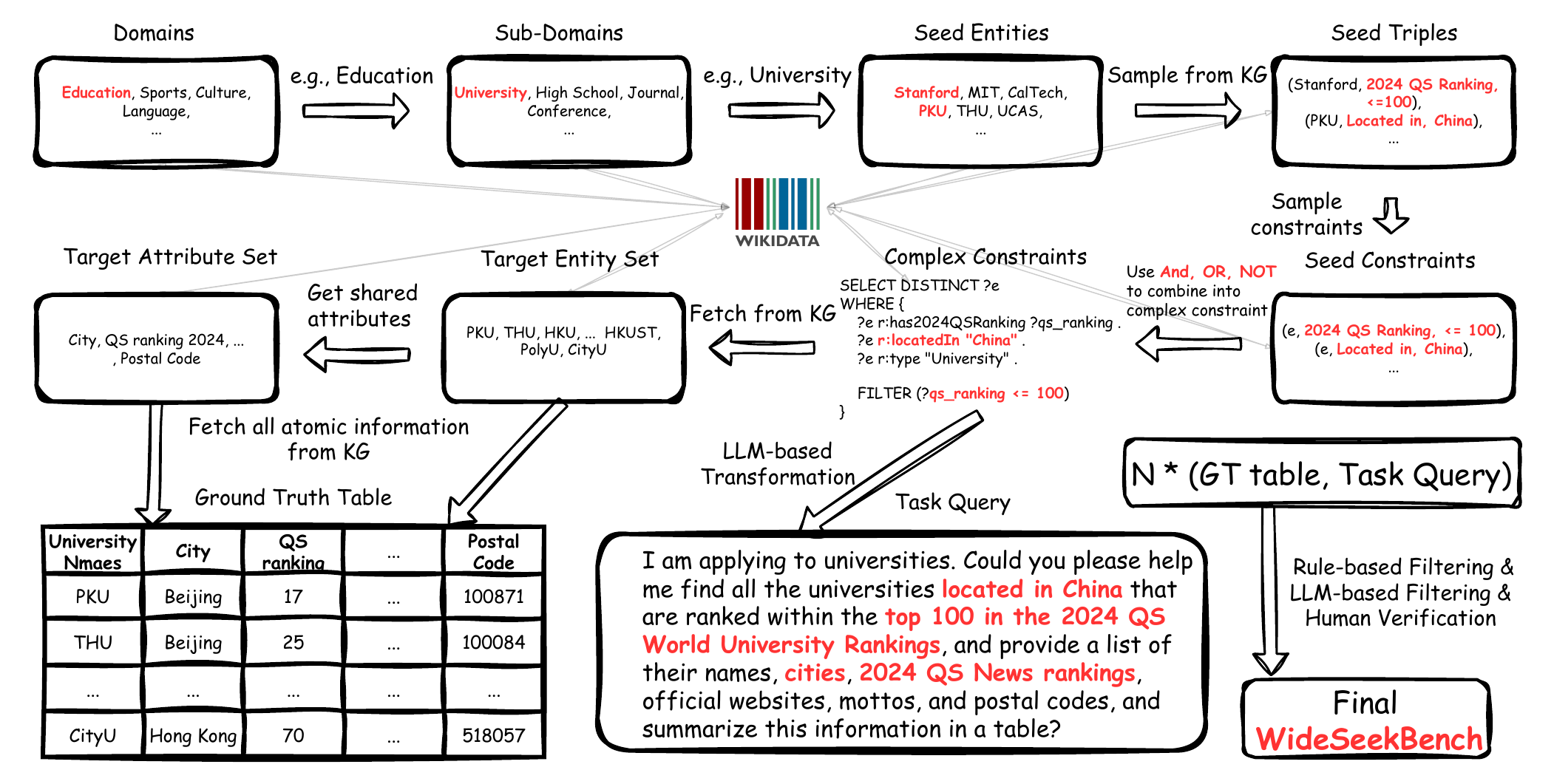}
    \caption{The data pipeline of WideSeekBench construction, which mines a set of target information under complex constraints.}
    \label{fig:bench}
\end{figure*}

\section{Data Pipeline \& Benchmark}

In contrast to Deep Research, Wide Research represents an application that is more closely aligned with real-world productivity scenarios. It aims to retrieve a collection of relevant information that satisfies complex constraints. Specifically, we can compile all relevant information into a table for comparative analysis. We define this task as General Broad Information Seeking (GBIS). To systematically evaluate models' Wide Research capabilities and to further investigate how post-training can enhance these capabilities in base models, we propose a rigorous multi-stage data pipeline and thus construct the \textbf{WideSeekBench}.

\subsection{Task Definition}
We define the GBIS task over a universe of entities $\mathcal{E}$ within a world knowledge space $\mathcal{W}$. A task instance is formally defined as a tuple $\mathcal{T} = (\mathcal{Q}, \mathcal{A})$, where $\mathcal{Q}$ is a task query encoding a complex semantic constraint, and $\mathcal{A} = \{a_1, a_2, \dots, a_m\}$ is the set of required attributes.

The query $\mathcal{Q}$ maps to a latent semantic filter function $\Phi: \mathcal{E} \to \{0, 1\}$. The objective is to construct a ground truth table $\mathbf{T^*}$ corresponding to the target entity set $\mathbf{E}^* = \{e \in \mathcal{E} \mid \Phi(e) = 1\}$. Formally, $\mathbf{T}^*$ is a table of size $|\mathbf{E}^*| \times m$:

{
\small
\begin{equation}
\mathbf{T}^* = \begin{bmatrix}
v_{1,1} & \cdots & v_{1,m} \\
\vdots & \ddots & \vdots \\
v_{|\mathbf{E}^*|,1} & \cdots & v_{|\mathbf{E}^*|,m}
\end{bmatrix},  v_{i,j} = \text{Value}(e_i, a_j)
\end{equation}
}

GBIS requires the agent to comprehensively synthesize $\mathbf{T}^*$. This requires not only the precision of the search but also the recall.

\subsection{Data Pipeline}
We employ a multi-phase approach on a knowledge graph $\mathcal{K}$ to synthesize complete benchmark instances of the form $(\mathcal{Q}, \mathcal{A}, \mathbf{T}^*, 
\mathcal{R})$, where $\mathcal{R}$ denotes the evaluation rubrics.
We provide more details in Appendix \ref{app:bench_details}.

\textbf{Phase 1: Seed Constraint Construction.}
To ensure comprehensive coverage and diversity, we adopt a top-down sampling strategy. 
(a) \textit{Domain Definition \& Sampling:} We start with a human-defined set of high-level domains $\mathcal{D}_{domain}$ (e.g., Education, Sports). From each high-level domain, we sample specific sub-domains $\mathcal{D}_{sub}$ (e.g., University, Basketball). 
(b) \textit{Seed Sampling:} Within each sub-domain, we sample seed entities $e_{seed}$ and extract their relations (triples) $\mathcal{R}_{seed} = \{(e_{seed}, p, v)\}$ from $\mathcal{K}$. 
This process yields a diverse pool of atomic constraints 
$\mathcal{C}_{atom}^{(e_{seed})} = \{ (p, v) \}$ associated with each seed entity.

\textbf{Phase 2: Logical Composition \& Schema Extension.}
We compose atomic constraints into complex constraints and extend the attribute schema. 
(a) \textit{Logical Composition:} Using operators $\mathcal{O} = \{\land, \lor, \neg\}$, we recursively define the composite filter $\Phi$ as:
\begin{equation}
    \Phi(e) := c(e) \mid \neg \Phi(e) \mid \Phi_1(e) \land \Phi_2(e) \mid \Phi_1(e) \lor \Phi_2(e)
\end{equation}
where $c(\cdot)$ denotes a boolean predicate induced by an atomic constraint $(p, v) \in \mathcal{C}_{atom}^{(e_{seed})}$, 
and $c(e)=1$ if entity $e$ satisfies property $p$ with value $v$.
We execute $\Phi$ over $\mathcal{K}$ to retrieve the target entity set $\mathbf{E}^*$.
(b) \textit{Schema Extension:}
Given the validated entity set $\mathbf{E}^*$, we construct a candidate attribute set
$\mathcal{A}_{cand} = \bigcup_{e \in \mathbf{E}^*} \text{Attributes}(e)$,
from which we select target attributes $\mathcal{A} \subset \mathcal{A}_{cand}$
by enforcing entity coverage and sufficient value diversity,
and retrieve all corresponding values to populate $\mathbf{T}^*$. This phase yields approximately 30,000 candidate tasks.

\textbf{Phase 3: Agent Task Synthesis.}
This phase converts complex constraints and target attributes into user-facing tasks using LLMs. 
(a) \textit{Self-Refining Query Synthesis:} We treat query generation as an iterative, self-refining process. An LLM generator $\mathcal{M}_{gen}$ converts $\Phi$ into a query $\mathcal{Q}$, while a LLM verifier $\mathcal{M}_{ver}$ extracts logic $\hat{\Phi}$ back from $\mathcal{Q}$. Discrepancies ($\hat{\Phi} \not\equiv \Phi$) trigger feedback loops for $\mathcal{M}_{gen}$ to regenerate $\mathcal{Q}$ until consistency is achieved. The consistency is also evaluated by $\mathcal{M}_{ver}$.
(b) \textit{Column-wise Rubric Generation:} For each attribute $a_j$, we generate a specific evaluation rubric $\mathcal{R}_j$ based on column semantics and cell values $\mathbf{T}^*_{\cdot, j}$, defining acceptance criteria for formats and tolerances. This phase yields approximately 15,000 candidate tasks.

\textbf{Phase 4: Multi-Stage Filtering.}
To ensure high quality, we apply a three-level filtering protocol: 
(a) \textit{Rule-based Filter:} We perform web searches to discard tasks where entities in $\mathbf{E}^*$ are not grounded in a web page. Moreover, we discard tasks where some cells lack natural language descriptions or $\mathbf{T}^*$ is sparse ($>50\%$ empty cells). 
(b) \textit{LLM-based Filter:} An LLM scores tasks against five dimensions: Human-Likeness, Solvability, Common Sense, Temporal Stability, and Rubric Rationality. A final task passes all these standards.
(c) \textit{Human Verification:} A final manual review removes subtle semantic irrationalities. This phase yields 5156 final tasks.

\subsection{WideSeekBench}
We introduce \textbf{WideSeekBench}, a comprehensive benchmark designed to evaluate Wide Research capabilities. The dataset comprises a total of 5,156 tasks, which are partitioned into a training set of 4,436 tasks $\mathcal{D}_{train}$ and a held-out test set of 720 tasks $\mathcal{D}_{test}$. The comparison of different search agent benchmarks is shown in Table~\ref{tab:benchmark_comparison}.

To enable fine-grained evaluation, we meticulously controlled the distribution of the test set. This allows for a multi-dimensional task classification and detailed analysis. Specifically, the test tasks are categorized based on three distinct dimensions:
(1) \textbf{Volume of Target Information:} We quantify the volume based on the total number of cells in the ground truth table. Based on this, tasks are divided into 10 distinct intervals to assess performance across varying information volume. The specific distribution is illustrated in Figure~\ref{fig:combined_stats}b.
(2) \textbf{Constraint Complexity:} To evaluate how agents handle complex tasks, we classify the tasks into 7 types based on the nature of the constraints involved. The distribution of these constraint types is presented in Table~\ref{tab:test_details}.
(3) \textbf{Domain Diversity:} We categorize the tasks into 18 distinct domains to ensure broad topical coverage. The domain-wise distribution is shown in Figure~\ref{fig:combined_stats}d.

Furthermore, we ensure that all entities in ground truth tables correspond to existing real-world web pages via search. To guarantee a fair, transparent, and reproducible evaluation, we constructed a standalone Simulated Environment. This environment includes a local document corpus and a local search engine. Detailed specifications of the simulated environment are provided in the Appendix \ref{app:envir}. Following WideSearch \cite{wong2025widesearchbenchmarkingagenticbroad}, we use Success Rate, Row F1, and Item F1 as the evaluation metrics. We show the details of evaluation in Appendix \ref{app:eval}.

\begin{figure*}[t]
    \centering
    \includegraphics[width=\linewidth]{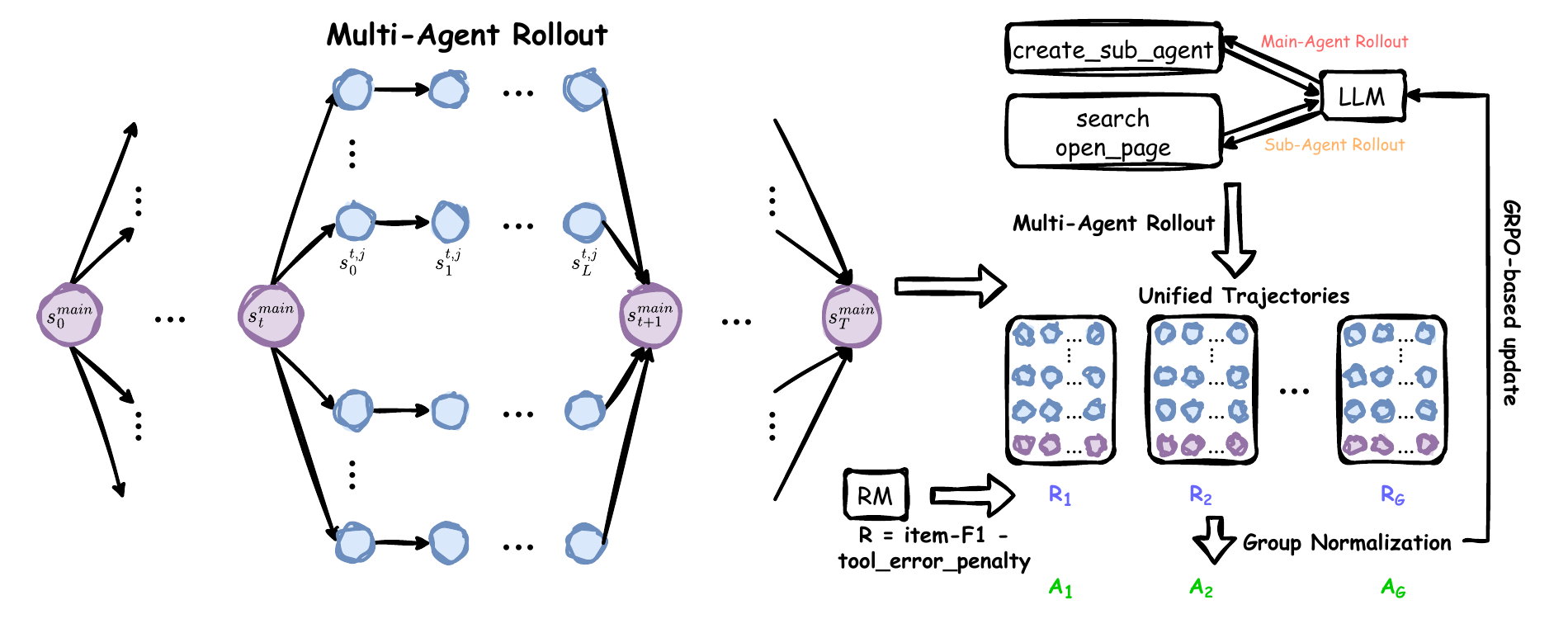}
    \caption{An illustration of Multi-Agent Reinforcement Learning. As shown on the left, the main agent can fork any number of sub-agents at any step. The trajectories of the main agent and sub-agents are unified for RL training.}
    \label{fig:method}
\end{figure*}

\section{WideSeek}
 Given a task $(\mathcal{Q}, \mathcal{A})$, the objective is to retrieve related information to construct a structured table $\hat{\mathbf{T}}$ containing a set of entities $\hat{\mathbf{E}} = \{e_1, e_2, \dots, e_N\}$ and their corresponding attribute values $v(\hat{\mathbf{E}}, \mathcal{A})$, satisfying a complex semantic constraint $\Phi$ derived from $\mathcal{Q}$.
To address the complexity of this task, which often exceeds the context and reasoning limits of a single serial trajectory, we propose \textbf{WideSeek}. WideSeek operates as a dynamic, hierarchical multi-agent system governed by a unified policy $\pi_\theta$.

\subsection{Multi-Agent Rollout}

The inference process, as shown in the left of Figure \ref{fig:method}, is modeled as a hierarchical Markov Decision Process (MDP) \cite{luo2025agentlightningtrainai}. Unlike static multi-agent architectures with fixed roles, WideSeek employs a centralized Main Agent (Planner) that dynamically forks variable instances of Sub-Agents (Executors) at any step.

\textbf{Hierarchical State Transition.}
At the top level, the Main Agent operates at time steps $t$. Let $s_t^{main}$ denote the global state, encompassing the user query $\mathcal{Q}$ and the history of high-level thoughts and sub-results. The Main Agent's policy $\pi_\theta(a_t^{main} | s_t^{main})$ selects an action $a_t^{main}$ from a hierarchical action space $\mathbf{A} = \mathbf{A}_{\text{planning}} \cup \mathbf{A}_{\text{termination}}$.

If $a_t^{main} \in \mathbf{A}_{\text{planning}}$, the agent invokes the function \texttt{create\_sub\_agent}$(q_{sub}^{(1)}, \dots, q_{sub}^{(k)})$. This action triggers the parallel instantiation of $k$ Sub-Agents, where $k$ is dynamically determined by the policy rather than a hyperparameter.
Each Sub-Agent $j$ ($j \in \{1, \dots, k\}$) operates in its own local MDP defined by the sub-task $q_{sub}^{(j)}$. It generates a trajectory $\mathcal{T}_{sub}^{(j)} = (s_{0}^j, a_{0}^j, s_{1}^j\dots)$\footnote{We reuse ${\mathcal{T}}$ to represent trajectories.} using the same unified policy $\pi_\theta$, utilizing atomic search tools (e.g., \texttt{search}, \texttt{open\_page}). Each action execution receives an observation $o^{j}_{t}$ from the environment and updates the sub-agent state: $s_{t+1}^{j} \leftarrow s_t^{j} \cup o^{j}_{t}$. Upon completion, the sub-agent returns a textual sub-result $r_j$, which updates the global state: $s_{t+1}^{main} \leftarrow s_t^{main} \cup \{r_1, \dots, r_k\}$.
If $a_t^{main} \in \mathbf{A}_{\text{termination}}$, the agent synthesizes the accumulated information in $s_t^{main}$ to produce the final answer $\mathbf{T}_{ans}$ and terminates the rollout.

This hierarchical execution generates a composite trajectory $\boldsymbol{\mathcal{T}}$ that interleaves the planner's reasoning traces with the execution traces of all dynamically created sub-agents.

\subsection{Cold Start}
Given the complexity of the task, we distill high-quality trajectories from multiple teacher models and fine-tune the policy via SFT (Supervised Fine-Tuning). Further details are provided in the Appendix \ref{app:sft}.

\subsection{Multi-Agent Reinforcement Learning}

Standard single-agent RL optimizes a sequential trajectory. However, WideSeek's execution graph is a dynamic tree structure. We propose a \textbf{Unified Multi-Agent RL} framework that models the entire system as a single generative process optimized via Group Relative Policy Optimization (GRPO) \cite{shao2024deepseekmathpushinglimitsmathematical}.

\textbf{Unified Trajectory Modeling.}
We model the multi-agent interaction as a unified joint distribution. Since all agents share the same LLM checkpoint $\pi_\theta$, we linearize the hierarchical execution trace into a single sequence.
First, we define the trajectory of the $j$-th Sub-Agent forked at the Main Agent's time step $t$ as a complete sequence of local state-action pairs:

\begin{equation}
    \mathcal{T}_{\text{sub}}^{(t,j)} = \left[ (s_{0}^{t,j}, a_{0}^{t,j}), (s_{1}^{t,j}, a_{1}^{t,j}), \dots, (s_{L}^{t,j}, r_{t,j}) \right]
\end{equation}

The global unified trajectory $\boldsymbol{\mathcal{T}}$ is then constructed by interleaving each Main Agent step $(s_t^{\text{main}}, a_t^{\text{main}})$ with the set of trajectories from all $K_t$ Sub-Agents forked at that step:

\begin{equation}
\begin{split}
\boldsymbol{\mathcal{T}} = \Bigg[ & (s_0^{\text{main}}, a_0^{\text{main}}), \bigcup_{j=1}^{K_0} \mathcal{T}_{\text{sub}}^{(0,j)}, \dots, \\
& (s_t^{\text{main}}, a_t^{\text{main}}), \underbrace{\bigcup_{j=1}^{K_t} \mathcal{T}_{\text{sub}}^{(t,j)}}_{\text{Executors at step } t}, \dots, (s_{T}^{\text{main}}, Y) \Bigg]
\end{split}
\end{equation}

\textbf{Reward Function Design.}
To guide the policy toward both accurate information retrieval and robust tool usage, we define a comprehensive global reward $R(\boldsymbol{\mathcal{T}})$ that serves as the sparse training signal. The reward is composed of a correctness score based on Item-F1 and a penalty for format violations.

To discourage structural degradation, we impose a format penalty. Let $n_{err}$ be the total count of format errors (e.g., invalid tool calls) in trajectory $\boldsymbol{\mathcal{T}}$, and $N_{max}$ be a predefined maximum tolerance for errors. The final reward function is defined as:

\begin{equation}
    R(\boldsymbol{\mathcal{T}}) = \text{Item-F1}(\mathbf{T}_{ans}, \mathbf{T}^*) - \lambda \cdot \underbrace{\left( \frac{n_{err}}{N_{max}} \right)}_{\text{Format Penalty}}
\end{equation}

where $\lambda$ is a balancing coefficient. This ensures that the agent is penalized proportionally to the frequency of format hallucinations relative to the tolerance threshold.

\textbf{Optimization via Unified GRPO.}
We optimize $\pi_\theta$ to maximize the expected reward of the unified trajectory. For each query $\mathcal{Q}$, we sample a group of $G$ unified trajectories $\{\boldsymbol{\mathcal{T}}_1, \dots, \boldsymbol{\mathcal{T}}_G\}$.
The Global GRPO objective is formally defined as:

\begin{equation}
\begin{split}
    \mathcal{J}(\theta) = \mathbb{E}_{\mathcal{Q} \sim \mathcal{D}, \{\boldsymbol{\mathcal{T}}_g\} \sim \pi_{\theta_{old}}} \Bigg[ \frac{1}{G} \sum_{g=1}^G \frac{1}{|\boldsymbol{\mathcal{T}}_g|} 
    \sum_{u=1}^{|\boldsymbol{\mathcal{T}}_g|}
    \frac{1}{|a_{u,k}|} \\ \sum_{k=1}^{|a_{u,k}|}
    \min \left( \rho_{g,u,k} \hat{A}_{g}, \text{clip}(\rho_{g,u,k}, 1-\epsilon, 1+\epsilon) \hat{A}_{g} \right) \Bigg]
\end{split}
\end{equation}

\begin{table*}[thbp]
\centering
\caption{Experiment results on WideSeekBench. We run each task for 4 times.}
\label{tab:results}
\resizebox{\linewidth}{!}{
\begin{tabular}{lccccccc}
\toprule
\multirow{2}{*}{\textbf{Model}} & \textbf{Success Rate (\%)} & \multicolumn{2}{c}{\textbf{Row F1 Score (\%)}} & \multicolumn{2}{c}{\textbf{Item F1 Score (\%)}} & \multirow{2}{*}{\textbf{\# Sub-Agents}} & \multirow{2}{*}{\textbf{\# Tool Calls}} \\
\cmidrule(lr){2-2} \cmidrule(lr){3-4} \cmidrule(lr){5-6}
 & Pass@4 & Mean@4 & Max@4 & Mean@4 & Max@4 & & \\
\midrule
\multicolumn{8}{l}{\textit{Proprietary Models}} \\
GPT-5.2 & 0.00 & \textbf{4.45} & 6.75 & \textbf{21.03} & \textbf{26.88} & 11.21 & \textbf{408.64} \\ 
GPT-5.1 & 0.00 & 4.11 & 6.75 & 20.44 & 27.88 & 6.02 & 121.36 \\ 
DeepSeek-v3.2 & 0.00 & 4.34 & \textbf{6.85} & 20.51 & 27.09 & \textbf{31.25} & 326.41 \\ 
Kimi-K2-Thinking & 0.00 & 3.17 & 5.86 & 17.48 & 25.19 & 8.74 & 85.36 \\ 
Seed-1.8 & 0.14 & 3.44 & 5.92 & 17.88 & 25.23 & 7.93 & 88.36 \\ 
\midrule
\multicolumn{8}{l}{\textit{Open-Sourced Models}} \\
Qwen3-8B-Thinking & 0.00 & 0.53 & 1.51 & 7.37 & 12.71 & 4.18 & 9.50 \\ 
Qwen3-30B-A3B-Thinking & 0.00 & 1.26 & 3.00 & 10.11 & 16.51 & 7.53 & 17.15 \\ 
WideSeek-8B-RL & 0.00 & 1.09 (\textcolor{red}{+0.56}) & 2.59 (\textcolor{red}{+1.08}) & 10.86 (\textcolor{red}{+3.49}) & 16.61 (\textcolor{red}{+3.90}) & 9.57 (\textcolor{red}{$\times$2.29}) & 41.09 (\textcolor{red}{$\times$4.33}) \\ 
WideSeek-8B-SFT & 0.14 & 1.74 (\textcolor{red}{+1.21}) & 3.66 (\textcolor{red}{+2.15}) & 11.35 (\textcolor{red}{+3.98}) & 18.92 (\textcolor{red}{+6.21}) & 13.16 (\textcolor{red}{$\times$3.15}) & 121.98 (\textcolor{red}{$\times$12.84}) \\ 
WideSeek-8B-SFT-RL & 0.00 & \textbf{1.95 (\textcolor{red}{+1.42})} & \textbf{3.88 (\textcolor{red}{+2.37})} & \textbf{12.87 (\textcolor{red}{+5.50})} & \textbf{19.73 (\textcolor{red}{+7.02})} & \textbf{26.60 (\textcolor{red}{$\mathbf{\times}$6.36})} & \textbf{273.75 (\textcolor{red}{$\mathbf{\times}$28.82})} \\ 
\bottomrule
\end{tabular}
}
\end{table*}

Here, $k$ indexes the action tokens generated by the model across the each step in linearized unified trajectory $\boldsymbol{\mathcal{T}}_g$, covering both Main Agent planning steps and Sub-Agent execution steps. The term $\rho_{g,u,k} = \frac{\pi_\theta(a_{u,k} | s_{u}, a_{u, <k})}{\pi_{\theta_{old}}(a_{u,k} | s_{u,k}, a_{u, <k})}$ represents the importance sampling ratio for the $k$-th token in the $u$-th action. The group-relative advantage $\hat{A}_g$ is computed using the global reward $R(\boldsymbol{\mathcal{T}}_g)$ as $\hat{A}_g = (R(\boldsymbol{\mathcal{T}}_g) - \mu_R) / \sigma_R$, where $\mu_R$ and $\sigma_R$ are the mean and standard deviation of rewards within the sampled group, respectively.

\section{Experiment}

\subsection{Setting}
We test proprietary models and open-sourced models on WideSeekBench. We use Qwen3-8B \cite{yang2025qwen3technicalreport} as the base for agent optimization. For more training settings, we show in Appendix \ref{app:setting}. To test the generalization to the Deep Research dataset, we test the agent on Browsecomp-plus \cite{chen2025browsecompplusfairtransparentevaluation}. We also show WideSeek trajectory example in Appendix \ref{app:case_study} for better understanding.

\subsection{Main Results}

\textbf{Scalability Gaps.}
As shown in Table~\ref{tab:results}, current state-of-the-art proprietary models, including GPT-5.2, exhibit limited success on the challenging WideSeekBench, with Mean@4 Item-F1 remaining only 21.03. This underscores the difficulty of conducting search at scale. Moreover, a distinct behavioral gap exists between proprietary and open-sourced models. Proprietary models spontaneously instantiate more sub-agents (e.g., DeepSeek-v3.2 forks 31.25) and execute significantly more tool calls (e.g., GPT-5.2 executes 408). This suggests that while current frontier models possess the potential for parallel task orchestration, they fail to effectively coordinate these actions to satisfy complex, high-breadth constraints without specialized optimization.

\textbf{Efficacy of WideSeek Optimization.}
We analyze the impact of our optimization method on the Qwen3-8B-Thinking as presented in Table~\ref{tab:results}. Distilling high-quality trajectories via SFT results in a strong performance boost, with WideSeek-8B-SFT achieving a 12.84$\times$ increase in tool usage and a 3.15$\times$ increase in sub-agent instantiation compared to the base model, indicating successful learning of multi-agent scaling. Further end-to-end optimization via RL yields the highest performance, where WideSeek-8B-SFT-RL achieves an Item F1 score of 12.87\% (+5.50\% over base) and a Max Row F1 of 3.88\%. The system learns to scale its search effort aggressively, increasing tool calls by a factor of 28.82$\times$ and sub-agents by 6.36$\times$. RL from scratch (WideSeek-RL) also learns to scale the number of sub-agents and tool calls, thus yielding better performance. While performance gains are substantial, they remain bounded by the 8B parameter size, suggesting that the reasoning bottleneck persists even with extensive retrieval. Additionally, Figure~\ref{fig:reward_tool_call} illustrates the training dynamics, revealing a strong correlation between the rising reward curve and increasing tool calls, confirming that the model discovers broader information seeking as the optimal policy.

\begin{figure*}[t]
    \centering
    \includegraphics[width=\linewidth]{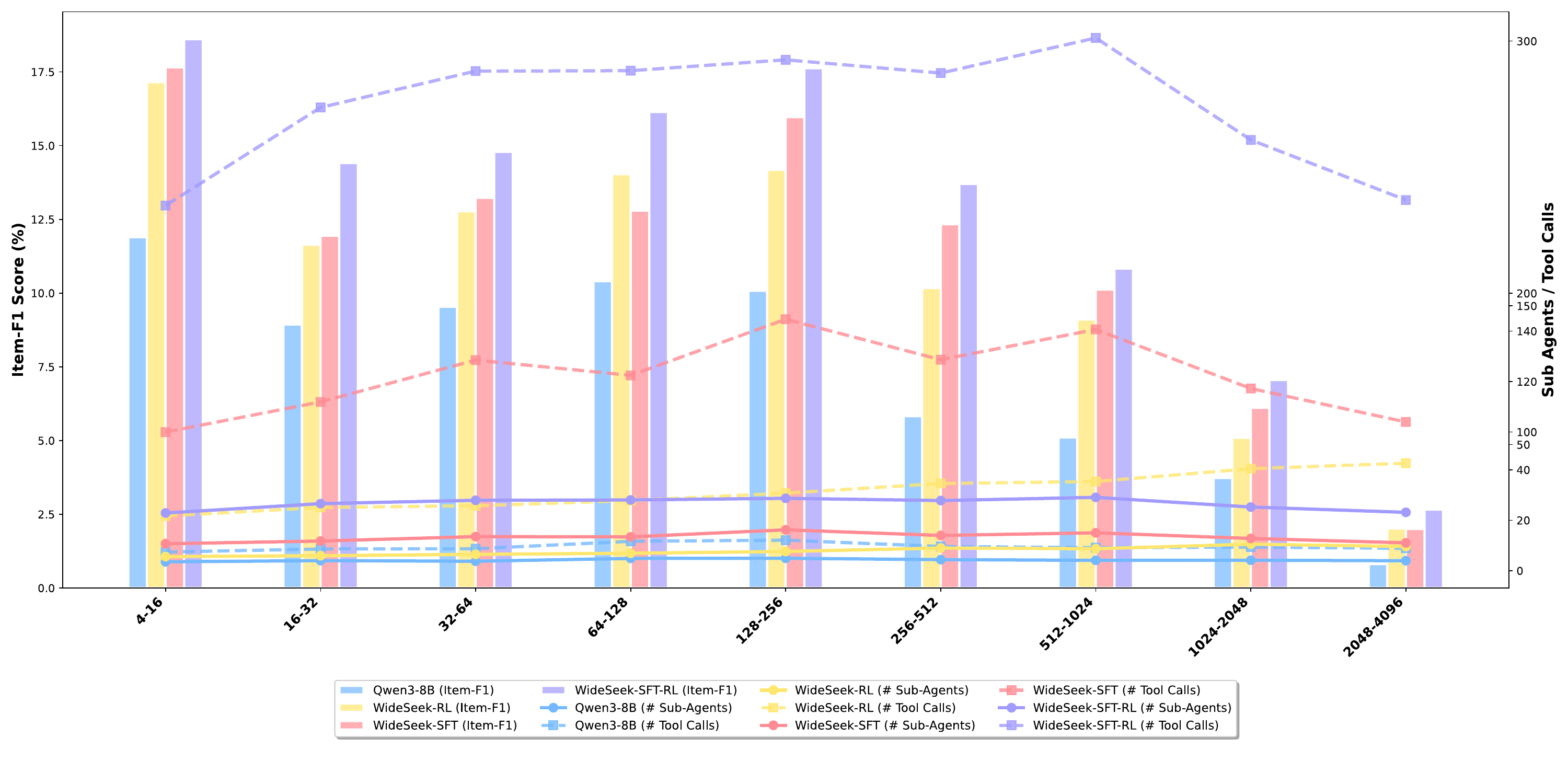}
    \caption{Item-F1 score, the number of sub-agents, and the number of tool calls on different task sets with different volume of target information.}
    \label{fig:volume_analysis}
\end{figure*}

\begin{table}[t]
    \centering
    \caption{Browsecomp-Plus performance. We test the generalization of WideSeek to Deep Research dataset.}
    \label{tab:model_acc}
    
    \resizebox{\linewidth}{!}{%
        \begin{tabular}{lcc}
            \toprule
            \textbf{Model} & \textbf{Scaffold} & \textbf{Acc} \\
            \midrule
            Gemini-2.5-Pro & ReAct & 29.52 \\ 
            GPT-OSS-120B-Low & ReAct & 25.54 \\
            DeepSeek-R1-0528 & ReAct & 16.39 \\
            Search-R1-32B & ReAct & 11.08 \\
            Qwen3-32B & ReAct & 10.72 \\
            \midrule 
            Qwen3-30B-A3B & WideSeek & 14.82 \\
            Qwen3-8B & WideSeek & 14.22 \\
            WideSeek-8B-SFT & WideSeek & 23.61 \\
            WideSeek-8B-SFT-RL & WideSeek & 23.61 \\
            WideSeek-8B-RL & WideSeek & \textbf{26.42 (\textcolor{red}{+12.20}) } \\
            \bottomrule
        \end{tabular}%
    }
\end{table}

\textbf{Generalization to Deep Research.}
To assess whether the capabilities transfer to deep research tasks, we evaluate our models on the BrowseComp-Plus (Table~\ref{tab:model_acc}). Even without any training, the WideSeek scaffold provides a structural advantage; the base Qwen3-8B utilizing WideSeek's dynamic multi-agent framework (14.22\%) outperforms significantly larger models like Qwen3-32B (10.72\%) that rely on ReAct. This suggests that decomposing complex queries into parallel sub-tasks effectively mitigates the context management burden. Furthermore, training on WideSeekBench confers robust generalization capabilities, with WideSeek-8B-RL achieving an accuracy of 26.42\%, a +12.20\% improvement over the base model. Despite being trained solely on wide research tasks, the agent's ability transfers effectively to deep research tasks.

\section{Analysis}

WideSeekBench facilitates a granular evaluation of agent capabilities through multi-dimensional task classification. Overall, our experimental results indicate that multi-agent RL consistently enhances performance across all analyzed dimensions, demonstrating the robustness of our method.

\textbf{Volume of Target Information.}
We categorize tasks based on the total count of atomic information in the ground truth table, ranging from small-scale intervals ([4, 16]) to massive-scale intervals ([2048, 4096]). As shown in Figure~\ref{fig:volume_analysis}, across all intervals, a consistent performance hierarchy is observed: WideSeek-8B-SFT-RL $>$ WideSeek-8B-SFT $>$ WideSeek-8B-RL. In the lower volume range ([4, 128]), performance gaps are minimal as the retrieval load remains manageable. However, in the range of [128, 4096], performance significantly degrades as the volume increases, confirming that massive-scale information seeking remains a formidable challenge. Notably, in the extreme interval ([2048, 4096]), both WideSeek-8B-SFT and WideSeek-8B-SFT-RL exhibit a counter-intuitive drop in tool call frequency alongside low success rates. This phenomenon suggests an "early stopping" behavior, likely stemming from the refusal tendencies distilled from the teacher model (frontier LLMs), which often assess such high-volume tasks as infeasible and reject them. Conversely, the WideSeek-8B-RL model, trained from scratch without SFT initialization, does not exhibit this bias; instead, its tool usage scales positively with atomic information volume, indicating that the agent has autonomously learned to deploy more extensive search actions to maximize recall in data-heavy scenarios.

\begin{figure}[h]
    \centering
    \includegraphics[width=\linewidth]{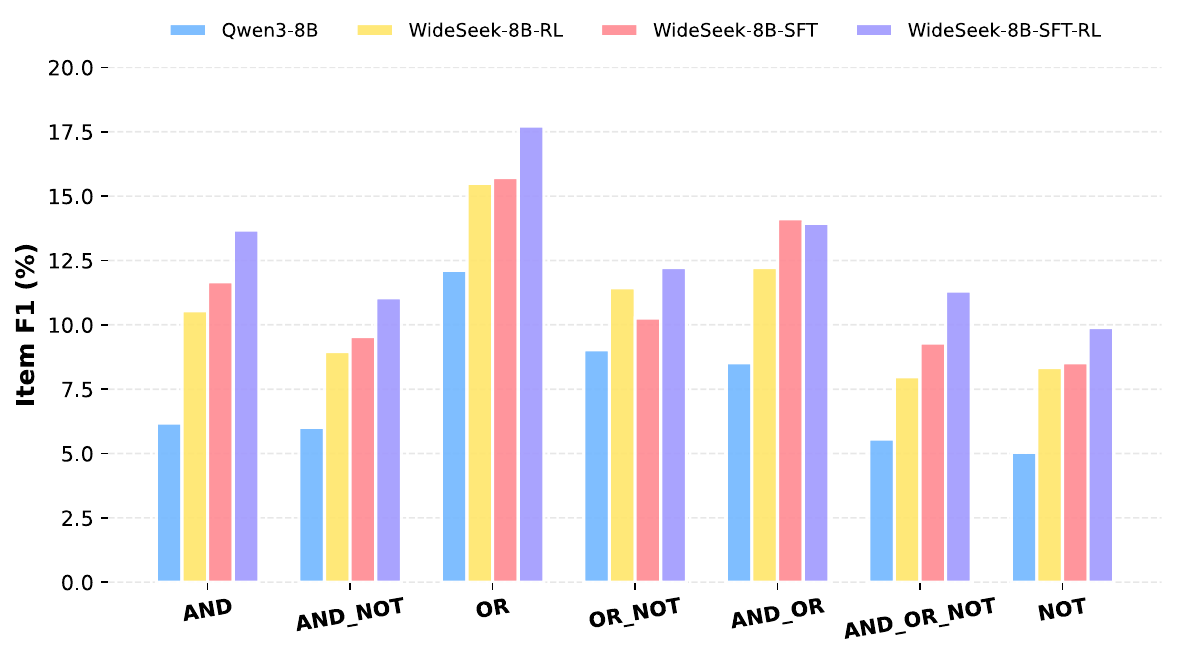}
    \caption{Item-F1 score on different constraint types.}
    \label{fig:constraint_analysis}
\end{figure}

\textbf{Constraint Type.}
We classify tasks into seven distinct logical constraint types corresponding to set operations in SPARQL (e.g., \texttt{AND}, \texttt{OR}, \texttt{NOT}), which represent the logic required to filter information sets (see Appendix \ref{app:logic_synthesis}). As illustrated in Figure~\ref{fig:constraint_analysis}, our analysis reveals that models generally achieve higher performance on `\texttt{OR}' type constraints. This is likely because disjunctive logic inherently aligns with parallel execution, allowing the system to easily decompose the query into independent sub-agents for concurrent search. In contrast, the `\texttt{NOT}' constraint type yields the lowest performance. Furthermore, compounding other constraints with negation (e.g., \texttt{OR\_NOT}) invariably leads to significant performance drops. This highlights that set difference operations (requiring the agent to exclude a specific entity set from the results) constitute a distinct reasoning bottleneck for current search agents.

\begin{figure}[h]
    \centering
    \includegraphics[width=\linewidth]{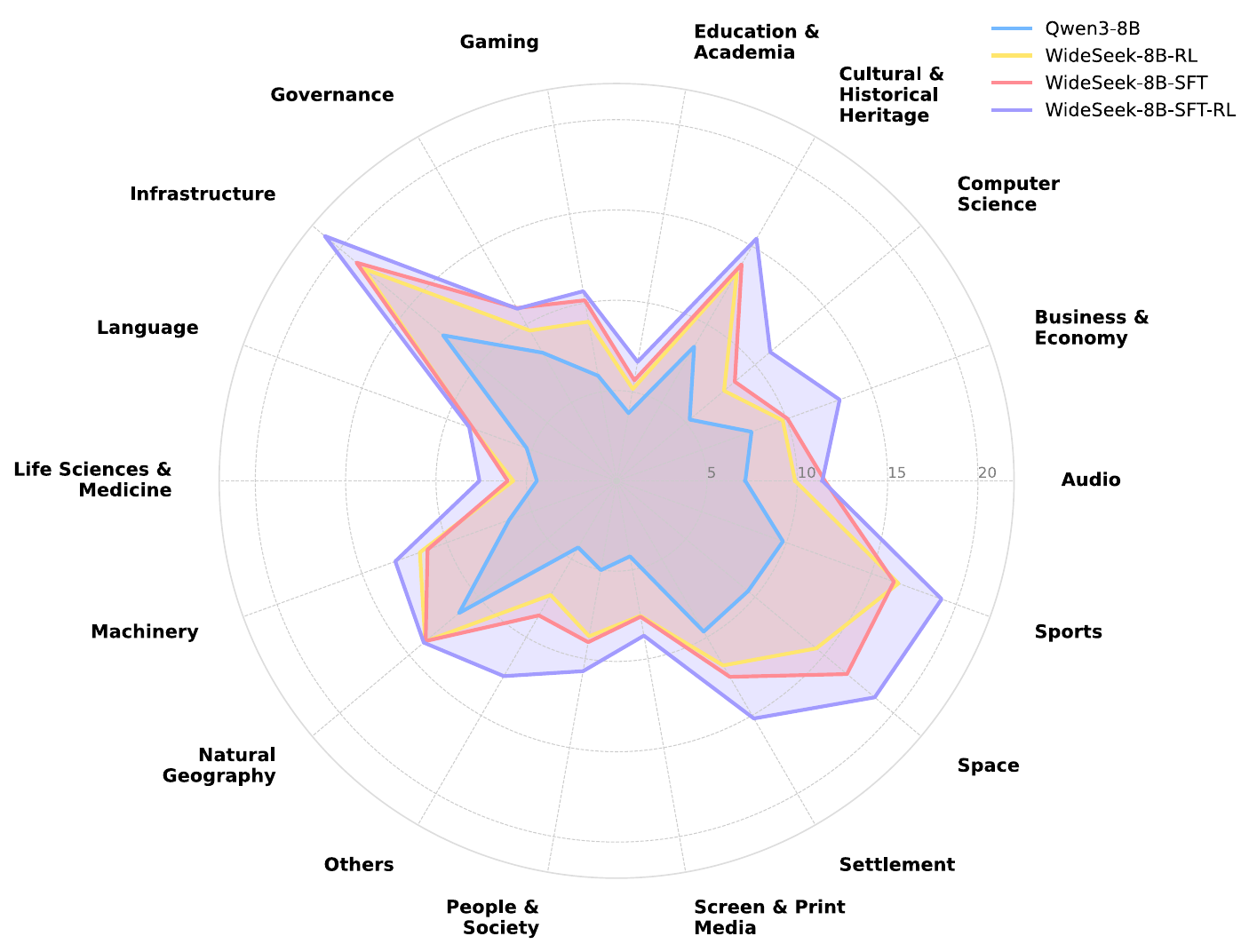}
    \caption{Item-F1 score on different domains.}
    \label{fig:domain_analysis}
\end{figure}

\textbf{Domain.}
We evaluate agent performance across 18 distinct domains. As shown in Figure~\ref{fig:domain_analysis}, the results demonstrate that our agent optimization strategy yields robust improvements universally, maintaining the trend WideSeek-8B-SFT-RL $>$ WideSeek-8B-SFT $>$ WideSeek-8B-RL across all categories. This validates the effectiveness of our method in enabling models to learn superior multi-agent coordination strategies during exploration to retrieve more comprehensive information. Simultaneously, the models exhibit consistent domain sensitivity; for instance, performance is notably higher in \textit{Infrastructure} compared to \textit{Education \& Academia}. 

\section{Related Work}
\subsection{Data Synthesis for Search Agent}
The training of search agents has shifted towards high-quality synthetic data to overcome the scale and diversity limits of human-curated benchmarks \cite{li2025websailornavigatingsuperhumanreasoning,tao2025webshaperagenticallydatasynthesizing,tongyidr}. Early synthesis efforts predominantly adopted an information-driven paradigm, focusing on simulating web navigation paths. For instance, WebWalkerQA \cite{wu2025webwalker} constructs linear information chains to emulate human browsing, while WebDancer \cite{wu2025webdancer} and WebSailor \cite{li2025websailornavigatingsuperhumanreasoning} leverage external information aggregation and entity coreference networks to generate complex QA pairs. However, these methods primarily optimize for search depth, focusing on the retrieval of specific reasoning paths to reach a single answer.
To enhance structural consistency and logical rigour, formalization-driven synthesis has gained attention, especially in the mathematical domain \cite{xin2024deepseek,ren2025deepseek} and the knowledge base question answering domain \cite{xia2025improving}. Most recently, WebShaper \cite{tao2025webshaperagenticallydatasynthesizing} pioneered the use of set-theoretic constructs (Knowledge Projections) to model information-seeking tasks. However, WebShaper still focuses on augmenting the reasoning structure to handle complex multi-step depth.

In contrast, our work introduces a formalization grounded in set theory specifically designed for search width. Unlike path-based or reasoning-oriented methods, we use Knowledge Graphs to extract clusters of interconnected world knowledge and define target entity sets within expansive search spaces using set operators. This allows us to precisely regulate task breadth and constraint complexity, addressing the ``Wide Research'' requirements that traditional information-driven \cite{wu2025webwalker, li2025websailornavigatingsuperhumanreasoning} or depth-oriented formalization \cite{tao2025webshaperagenticallydatasynthesizing} paradigms do not fully cover.

\subsection{LLM-based Multi-Agent Reinforcement Learning}

Traditional Large Language Model (LLM)-based multi-agent systems primarily rely on static, heuristic-driven architectures with pre-defined roles, often lacking parameter-level optimization for specific collaborative tasks \cite{qian2024chatdev, hong2023metagpt}. Recently, the research community has shifted toward cooperative MARL to enable more effective coordination. For instance, MAGRPO \cite{liu2025llm} introduces a multi-agent group relative policy optimization to fine-tune multiple LLMs for writing and coding tasks, moving beyond individual rewards toward collective efficiency. Similarly, the Optimized Workforce Learning (OWL) framework \cite{hu2025owl} utilizes reinforcement learning to optimize a domain-agnostic planner for complex task decomposition. While these works demonstrate the potential of RL in multi-agent coordination, they either focus on general-purpose cooperation or decouple planning from execution to maintain transferability, often leaving the specialized executors as black-box modules.  M-GRPO \cite{hong2025multiagentdeepresearchtraining} and Fold-GRPO \cite{sun2025scalinglonghorizonllmagent} use the branch-return paradigm, but they usually fork a fixed number of sub-agents (i.e., 1) for sub-tasks execution at each step of the main agent.

The industry has also seen the emergence of advanced agentic products, such as Kimi K2.5 Agent Swarm \cite{moonshot2026kimi25}, which achieves impressive performance by optimizing the orchestrator while treating sub-agents as static parameters. However, such "orchestration-only" optimization may limit the system's ability to refine the interaction granularity between the planner and executors. In contrast, we propose an end-to-end reinforcement learning approach that simultaneously optimizes both the main planner agent and the sub-agents (executors). Unlike OWL's decoupling or Kimi 2.5's static sub-agent paradigm, our work enables the entire system to co-evolve, allowing the main agent to autonomously broaden search paths while the sub-agents adapt their retrieval and synthesis strategies for industrial-scale "Wide Research." This joint optimization ensures that the planning of breadth and the execution of tool-calling are aligned toward maximizing final search utility.

\section{Conclusion}
To address the paradigm shift from Deep to Wide Research, we introduce WideSeekBench to formalize the General Broad Information Seeking (GBIS) task. We construct it via a rigorous multi-phase data pipeline that mines intersected world knowledge from KGs. We propose WideSeek, a dynamic hierarchical multi-agent architecture optimized via an end-to-end reinforcement learning framework. Our results demonstrate that WideSeek effectively leverages agent scaling to solve complex, parallel retrieval tasks, significantly advancing Wide Research capabilities.








\section*{Impact Statement}
This paper presents work whose goal is to advance the field of Machine
Learning. There are many potential societal consequences of our work, none
which we feel must be specifically highlighted here.





\bibliography{example_paper}
\bibliographystyle{icml2026}

\newpage
\appendix
\onecolumn
\section{The Details of WideSeekBench}
\label{app:bench_details}

\subsection{Benchmark Comparison}
\begin{table*}[h]
\centering
\caption{Comparison of WideSeekBench with existing information-seeking benchmarks. \textbf{Task Type} distinguishes between finding specific hidden info (Deep) vs. collecting broad structured info (Wide). \textbf{Auto Gen.} indicates if the data pipeline is automated. \textbf{Multi-dim.} indicates if tasks are classified by fine-grained dimensions (i.e., constraints, domains.).}

\label{tab:benchmark_comparison}

\resizebox{\textwidth}{!}{
\begin{tabular}{lcccccc}
\toprule
\textbf{Benchmark} & \textbf{Size} & \textbf{Domains} & \textbf{Task Type} & \textbf{Train Set} & \textbf{Auto Gen.} & \textbf{Multi-dim Class.} \\
\midrule
GAIA (Text-Only Split) & 103 & - & Deep & \ding{55} & \ding{55} & \ding{55} \\
BrowseComp  & 1,266 & 9 & Deep & \ding{55} & \ding{55} & \ding{55}  \\
BrowseComp-ZH  & 289 & 11 & Deep & \ding{55} & \ding{55} & \ding{55}  \\
WideSearch  & 200 & 14 & Wide & \ding{55} & \ding{55} & \ding{55}  \\
DeepWideSearch  & 220 & 15 & Wide & \ding{55} & MIX & \ding{55}  \\
xbench-DeepSearch & 100 & - & Deep & \ding{55} & \ding{55} & \ding{55}  \\
WebShaper & 5,000 & - & Deep& \ding{51} & \ding{51} & \ding{55} \\
\midrule
\textbf{WideSeekBench (Ours)} & \textbf{5,156} & \textbf{18} & \textbf{Wide} & \ding{51} & \ding{51} & \ding{51} \\
\bottomrule
\end{tabular}
}
\end{table*}

\subsection{Knowledge Graph Source and Infrastructure}
\label{app:infrastructure}

We ingest the Wikidata Truthy Dump (October 1, 2025) into a local QLever \cite{qlever} SPARQL engine
 to support efficient, rate-limit-free execution of complex SPARQL queries over the full knowledge graph.

\subsection{Seed Constraint Construction}
\label{app:sampling}

We construct a diverse set of seed entities to serve as the semantic basis for downstream constraint construction and task synthesis.

\paragraph{Domain Taxonomy.}
We define 18 high-level domains (e.g., \textit{Computer Science, Life Sciences, Governance}). Each domain is mapped to a set of Wikidata classes, which are treated as domain-specific sub-domains. In total, this mapping yields 200 sub-domains across all domains. These sub-domains jointly define a controlled search scope $\mathcal{S}_{sub-domain}$ (refer to Appendix~\ref{app:domain_scale} for details).

\paragraph{Retrieval and Ranking.}
For each sub-domain, we identify 80 informative seed entities from the knowledge base $\mathcal{K}$ using a three-stage SPARQL-based workflow.
(1) \textit{Retrieval:} Given a sub-domain class, we retrieve a candidate entity set $\mathbf{E}_{cand}$ by recursively querying the class and all its subclasses via the transitive closure of the \texttt{wdt:P279} (subclass of) relation (Listing~\ref{lst:retrieval_query}).
(2) \textit{Ranking:} Each candidate entity $e \in \mathbf{E}_{cand}$ is ranked by its information density, approximated by the number of outgoing RDF triples associated with $e$ (Listing~\ref{lst:ranking_query}). Entities with higher information density are preferred, as they support the construction of richer constraints and attribute schemas.
(3) \textit{Filtering:} We remove non-entity artifacts and structurally uninformative entries, including entities whose labels begin with \texttt{"List of"} or \texttt{"Category:"}. The remaining entities constitute the seed entity set $\mathbf{E}_{seed}$.

\begin{lstlisting}[
    language=SPARQL, 
    caption={Candidate entity retrieval via recursive subclass matching, where \texttt{wdt:P31} and \texttt{wdt:P279} represent the 'instance of' and 'subclass of' relations in Wikidata, respectively.}, 
    label={lst:retrieval_query},
    basicstyle=\ttfamily\small,
    frame=single
]
SELECT DISTINCT ?entity WHERE {
  ?entity (wdt:P31/wdt:P279*) wd:TARGET_ID .
}
\end{lstlisting}

\begin{lstlisting}[
    language=SPARQL, 
    caption={Ranking entities by information density (triple count).}, 
    label={lst:ranking_query},
    basicstyle=\ttfamily\small,
    frame=single
]
SELECT ?entity ?label (COUNT(?p) AS ?count) WHERE {
  VALUES ?entity { wd:Q_CANDIDATE_1 ... }
  ?entity ?p ?o .
  OPTIONAL { ?entity rdfs:label ?label . FILTER(LANG(?label) = "en") }
}
GROUP BY ?entity ?label
ORDER BY DESC(?count)
\end{lstlisting}


\subsection{Logical Composition and Task Synthesis}
\label{app:logic_synthesis}

We describe the procedures for composing atomic constraints into executable queries, executing and validating the resulting retrievals, and constructing bounded tables. For each seed entity, we generate up to 200 composite constraints. To control redundancy and dataset balance, each seed contributes at most 4 validated tables.

\paragraph{Query Formulation.}
Given a sampled seed entity $e_{\text{seed}} \in \mathbf{E}_{\text{seed}}$ and its associated relations
$\mathcal{R}_{\text{seed}} = \{(e_{\text{seed}}, p, v)\}$, retrieved from $\mathcal{K}$ via property-seeking SPARQL queries,
we define the associated atomic constraint set $\mathcal{C}_{atom}^{(e_{\text{seed}})} = \{(p, v)\}$.
We then sample atomic constraints $c \in \mathcal{C}_{atom}^{(e_{\text{seed}})}$ and compose them into composite SPARQL filters
using seven predefined logical patterns (Table~\ref{tab:logic_sparql}), yielding a composite constraint $\Phi$.
Apart from the domain constraint, each composite constraint $\Phi$ is required to contain at least
1 and at most 8 atomic constraints.

\paragraph{Execution and Verification.}
Each composite filter $\Phi$ is executed against the knowledge base $\mathcal{K}$ to retrieve a candidate entity set $\mathbf{E}^*$. We restrict the cardinality of the candidate entity set $\mathbf{E}^*$ to the interval $[1, 1024]$. As shown in Listing~\ref{lst:cardinality}, a verification step enforces this constraint prior to attribute retrieval, discarding any queries where $|\mathbf{E}^*|$ falls outside the bound.

\paragraph{Table Construction and Quality Control.} 
Given the validated entity set $\mathbf{E}^*$, we first collect a candidate attribute set
$\mathcal{A}_{cand} = \bigcup_{e \in \mathbf{E}^*} \text{Attributes}(e)$,
and dynamically select target attributes $\mathcal{A} \subset \mathcal{A}_{cand}$
by retaining only those with at least 50\% coverage across entities and sufficient value diversity
(Listing~\ref{lst:prop_freq}). Next, we compute the potential table size $N_{\text{cells}} = |\mathbf{E}^*| \times |\mathcal{A}|$, and retain tasks satisfying $N_{\text{cells}} \in [8, 8192]$. Entities that fail to resolve to valid labels are removed, resulting in the cleaned entity set $\mathbf{E}_{\text{clean}}$. Finally, we perform batch SPARQL queries to retrieve all cell values (Listing~\ref{lst:fetch_values}) and populate the table $\mathbf{T}^*$. To avoid redundancy, we further deduplicate tables by discarding those with identical entity sets and attribute schemas, retaining only one representative table per equivalence class.

\begin{table*}[t]
\centering
\setlength{\abovecaptionskip}{0.1in}
\setlength{\belowcaptionskip}{0.1in}
\caption{Logical patterns in WideSeekBench. $\mathcal{D}$ denotes the domain constraint.}
\label{tab:logic_sparql}
\begin{small}
\begin{tabular}{l c p{4.5cm} p{8cm}}
\toprule
\textbf{Pattern} & \textbf{Prob.} & \textbf{Formulation $\Phi(e)$} & \textbf{SPARQL Implementation} \\
\midrule
\texttt{AND} & 20\% & $\mathcal{D}(e) \land (\bigwedge_{i} c_i(e))$ & 
\begin{minipage}{\linewidth}
\begin{lstlisting}[language=SPARQL, basicstyle=\ttfamily\tiny, breaklines=true, numbers=none, frame=none, aboveskip=2pt, belowskip=0pt]
SELECT ?item WHERE {
  ?item wdt:P31 wd:Q_dom . 
  ?item wdt:P1 wd:Q1 . ?item wdt:P2 wd:Q2 .
}
\end{lstlisting}
\end{minipage} \\ \midrule

\texttt{OR} & 20\% & $\mathcal{D}(e) \land (\bigvee_{i} c_i(e))$ & 
\begin{minipage}{\linewidth}
\begin{lstlisting}[language=SPARQL, basicstyle=\ttfamily\tiny, breaklines=true, numbers=none, frame=none, aboveskip=2pt, belowskip=0pt]
SELECT ?item WHERE {
  ?item wdt:P31 wd:Q_dom .
  { { ?item wdt:P1 wd:Q1 } UNION { ?item wdt:P2 wd:Q2 } }
}
\end{lstlisting}
\end{minipage} \\ \midrule

\texttt{NOT} & 15\% & $\mathcal{D}(e) \land c_{base}(e) \land \neg (\bigvee_{i} c_{ex_i}(e))$ & 
\begin{minipage}{\linewidth}
\begin{lstlisting}[language=SPARQL, basicstyle=\ttfamily\tiny, breaklines=true, numbers=none, frame=none, aboveskip=2pt, belowskip=0pt]
SELECT ?item WHERE {
  ?item wdt:P31 wd:Q_dom . ?item wdt:P_base wd:Q_base .
  FILTER NOT EXISTS {
    { ?item wdt:P_ex1 wd:Q_ex1 } UNION { ?item wdt:P_ex2 wd:Q_ex2 }
  }
}
\end{lstlisting}
\end{minipage} \\ \midrule

\texttt{AND\_OR} & 15\% & $\mathcal{D}(e) \land [(\bigwedge_{i} c_{i}(e)) \lor (\bigwedge_{j} c_{j}(e))]$ & 
\begin{minipage}{\linewidth}
\begin{lstlisting}[language=SPARQL, basicstyle=\ttfamily\tiny, breaklines=true, numbers=none, frame=none, aboveskip=2pt, belowskip=0pt]
SELECT ?item WHERE {
  ?item wdt:P31 wd:Q_dom .
  { { ?item wdt:P1 wd:Q1 . ?item wdt:P2 wd:Q2 } 
    UNION 
    { ?item wdt:P3 wd:Q3 . ?item wdt:P4 wd:Q4 } }
}
\end{lstlisting}
\end{minipage} \\ \midrule

\texttt{AND\_NOT} & 15\% & $\mathcal{D}(e) \land (\bigwedge_{i} c_{in_i}(e)) \land \neg (\bigvee_{j} c_{ex_j}(e))$ & 
\begin{minipage}{\linewidth}
\begin{lstlisting}[language=SPARQL, basicstyle=\ttfamily\tiny, breaklines=true, numbers=none, frame=none, aboveskip=2pt, belowskip=0pt]
SELECT ?item WHERE {
  ?item wdt:P31 wd:Q_dom .
  ?item wdt:P1 wd:Q1 . ?item wdt:P2 wd:Q2 .
  FILTER NOT EXISTS {
    { ?item wdt:P_ex1 wd:Q_ex1 } UNION { ?item wdt:P_ex2 wd:Q_ex2 }
  }
}
\end{lstlisting}
\end{minipage} \\ \midrule

\texttt{OR\_NOT} & 10\% & $\mathcal{D}(e) \land (\bigvee_{i} c_{in_i}(e)) \land \neg (\bigvee_{j} c_{ex_j}(e))$ & 
\begin{minipage}{\linewidth}
\begin{lstlisting}[language=SPARQL, basicstyle=\ttfamily\tiny, breaklines=true, numbers=none, frame=none, aboveskip=2pt, belowskip=0pt]
SELECT ?item WHERE {
  ?item wdt:P31 wd:Q_dom .
  { { ?item wdt:P1 wd:Q1 } UNION { ?item wdt:P2 wd:Q2 } }
  FILTER NOT EXISTS {
    { ?item wdt:P_ex1 wd:Q_ex1 } UNION { ?item wdt:P_ex2 wd:Q_ex2 }
  }
}
\end{lstlisting}
\end{minipage} \\ \midrule

\texttt{AND\_OR\_NOT} & 5\% & $\mathcal{D}(e) \land [(\bigwedge_{i} c_{i}(e)) \lor (\bigwedge_{j} c_{j}(e))] \land \neg (\bigvee_{k} c_{ex_k}(e))$ & 
\begin{minipage}{\linewidth}
\begin{lstlisting}[language=SPARQL, basicstyle=\ttfamily\tiny, breaklines=true, numbers=none, frame=none, aboveskip=2pt, belowskip=0pt]
SELECT ?item WHERE {
  ?item wdt:P31 wd:Q_dom .
  { { ?item wdt:P1 wd:Q1 . ?item wdt:P2 wd:Q2 }
    UNION { ?item wdt:P3 wd:Q3 . ?item wdt:P4 wd:Q4 } }
  FILTER NOT EXISTS {
    { ?item wdt:P_ex1 wd:Q_ex1 } UNION { ?item wdt:P_ex2 wd:Q_ex2 }
  }
}
\end{lstlisting}
\end{minipage} \\ 
\bottomrule
\end{tabular}
\end{small}
\end{table*}

\begin{lstlisting}[language=SPARQL, caption={Pre-flight cardinality check.}, label={lst:cardinality}, basicstyle=\ttfamily\scriptsize, frame=single]
SELECT (COUNT(DISTINCT ?item) AS ?count) WHERE {
  # Synthesized Logical Constraints (see Table 4)
  ?item wdt:P31 wd:Q_domain .
  ...
}
\end{lstlisting}

\begin{table}[h]
\centering
\setlength{\abovecaptionskip}{0.1in}
\setlength{\belowcaptionskip}{0.1in}
\caption{Property ID Blacklist.}
\label{tab:property_blacklist}
\begin{small}
\begin{tabular}{l l}
\toprule
\textbf{Property ID} & \textbf{Label} \\
\midrule
\texttt{P31}   & Instance of \\
\texttt{P106}  & Occupation \\
\texttt{P108}  & Employer \\
\texttt{P248}  & Stated in \\
\texttt{P279}  & Subclass of \\
\texttt{P361}  & Part of \\
\texttt{P373}  & Commons category \\
\texttt{P460}  & Said to be the same as \\
\texttt{P527}  & Has part \\
\texttt{P646}  & Freebase ID \\
\texttt{P910}  & Topic's main category \\
\texttt{P1001} & Applies to jurisdiction \\
\texttt{P1343} & Described by source \\
\texttt{P1709} & Equivalent class \\
\texttt{P1754} & Category related to list \\
\texttt{P1889} & Different from \\
\texttt{P2671} & Google Knowledge Graph ID \\
\texttt{P3876} & Category for alumni of educational institution \\
\texttt{P6104} & Maintained by WikiProject \\
\bottomrule
\end{tabular}
\end{small}
\end{table}

\begin{minipage}{\linewidth}
\begin{lstlisting}[language=SPARQL, caption={Attribute frequency analysis.}, label={lst:prop_freq}, basicstyle=\ttfamily\scriptsize, frame=single]
SELECT ?prop (COUNT(DISTINCT ?item) AS ?cnt) WHERE {
  VALUES ?item { wd:Q_sample1 ... }
  ?item ?prop ?value .
  FILTER(STRSTARTS(STR(?prop), "http://www.wikidata.org/prop/direct/"))
} GROUP BY ?prop
\end{lstlisting}
\end{minipage}

\begin{lstlisting}[language=SPARQL, caption={Batch value retrieval.}, label={lst:fetch_values}, basicstyle=\ttfamily\scriptsize, frame=single]
SELECT ?item ?prop ?value ?valueLabel WHERE {
  VALUES ?item { wd:Q_e1 ... }
  VALUES ?directProp { wdt:P1 ... }
  ?item ?directProp ?value .
  ?prop wikibase:directClaim ?directProp . # Map back to direct predicate
  OPTIONAL { ?value rdfs:label ?valueLabel . FILTER(LANG(?valueLabel) = "en") }
}
\end{lstlisting}

\subsection{Agent Task Synthesis and Multi-Stage Filtering}
\label{app:nl_generation}

We implement a cyclic generation-verification pipeline to transform structured logical filters $\Phi$ into diverse, human-like search tasks $Q$, followed by a rigorous quality assurance protocol. In this subsection, all LLM-based operations are powered by GPT-5.

\paragraph{Self-Refining Query Synthesis.}
The transformation process employs a dual-model architecture to ensure both linguistic diversity and logical fidelity. First, raw constraints are mapped into a structured f-string template (e.g., \texttt{"Find all \{sub-domain\} that \{prop\} is \{val\}..."}). A generator model $M_{gen}$ then transforms this template into natural language using a style randomization protocol, sampling a syntactic mode $s \sim U(1, 10)$ from a predefined set (\textit{Action}, \textit{Question}, \textit{Imperative}, \textit{Need}, \textit{Context}, \textit{Interest}, \textit{Description}, \textit{Casual}, \textit{Professional} and \textit{Task}) for each task. To ensure semantic accuracy, a critic model $M_{ver}$ extracts the logic $\hat{\Phi}$ back from the generated query $Q$ and performs a constraint-by-constraint equivalence check $S(\Phi, \hat{\Phi})$. The verifier rigorously compares entity preservation, operator logic ($\land, \lor, \neg$), filtering scope, and output schema consistency. Any discrepancy triggers a feedback loop with specific error correction instructions, capped at $k=5$ iterations.

\paragraph{Data-Driven Rubric Synthesis.}
We leverage an LLM to synthesize adaptive evaluation criteria $R_j$ by analyzing the data distribution of each ground truth column $\mathbf{T}^*_{\cdot,j}$. Unlike rigid string matching, the model generates semantic compliance standards tailored to the specific data type: (1) \textit{Entities} explicitly accept aliases and naming variations; (2) \textit{Dates} enforce semantic exactness regardless of format; (3) \textit{Numerics} require value equality within defined tolerances; and (4) \textit{Sets} enforce equality independent of item order.

\paragraph{Quality Assurance Protocol.}
We apply a three-tier filtering mechanism. (1) \textit{Rule-Based Filtering} discards tasks with sparse ground truth ($>50\%$ empty cells) or weak web grounding where target entities lack verifiable search API hits, as determined by their English sitelink counts in Wikidata, where entities with zero English sitelinks are strictly filtered out. (2) \textit{LLM-Based Filtering} employs a judge model to evaluate tasks on a 5-point scale across \textit{Human-Likeness}, \textit{Solvability}, \textit{Common Sense}, \textit{Temporal Stability}, and \textit{Rubric Rationality}, a violation in any category results in immediate rejection.(3) \textit{Human Verification} removes subtle semantic irrationalities (e.g., logical contradictions) that automated filters might overlook.

\begin{tcolorbox}[
    colback=gray!5, 
    colframe=black!75, 
    title=\textbf{Prompt: LLM-Based Filtering}, 
    fonttitle=\bfseries\sffamily,
    breakable,
    label={lst:auditor_prompt} 
]
\begin{lstlisting}[
    basicstyle=\ttfamily\scriptsize, 
    breaklines=true, 
    numbers=none, 
    frame=none, 
    aboveskip=0pt, 
    belowskip=0pt
]
You are an expert Data Quality Auditor for a Search Agent Benchmark.
Your goal is to validate synthesized widesearch tasks to ensure they are high-quality, solvable, and fairly evaluated.

## Input Data provided for each request:
1. Task Info: The natural language user query and basic metadata.
2. Rubrics: The evaluation criteria for each column (preprocess, metric, criterion).
3. Ground Truth Stats & Sample: Statistics of the expected answer table and a preview of the data.
4. Intermediate Logic: The SPARQL query and constraints used to generate the task.

## Evaluation Criteria (Pass/Fail)

Analyze the task against the following 5 dimensions. If it fails on ANY dimension, mark as INVALID.

### 1. Human-Likeness & Phrasing
* Goal: Ensure the query sounds like a real human request.
* Reject if phrasing is unnatural, overly pedantic, or a raw SPARQL translation.
    * Fail: "Find entities where mass is exactly 70.53 kg."
    * Fail: "Find items with property P31 equal to Q5." (Leaked IDs).
    * Pass: "Find all public hospitals in Portugal founded before 1950."

### 2. Task Solvability
* Goal: Ensure the task is logically possible.
* Reject if logic contains contradictions (e.g., "born > 2000 AND died < 1999").
* Reject if it asks for private/internal info not publicly available.

### 3. Answer Common Sense
* Goal: Detect "Hallucinated" or "Incomplete" Ground Truth.
* Reject if GT clearly contradicts general world knowledge.
    * Fail: "All US States" -> GT has only 5 rows (Recall too low).
    * Fail: "Current UK Monarchs" -> GT has 10 rows (Recall too high/Wrong).
    * Fail: GT has >50% empty cells for key attributes.

### 4. Temporal Stability & Wiki-Fit
* Goal: Ensure facts are static and suitable for Wikipedia search.
* Reject if query relies on "Current/Now" without a time anchor (e.g., "Current CEO") UNLESS the GT explicitly handles specific dates.
* Reject if attributes are purely database metadata (e.g., "Wikidata QID", "GND ID") unlikely to be found in Wikipedia text.

### 5. Rubric Rationality (Criterion Quality Check)
* Goal: Ensure the criterion explicitly instructs the judge to handle the variability of web data (formats, aliases, precision).
* Reject if the criterion enforces rigid constraints that are unrealistic for web extraction.
    * Fail (False Strictness on Non-Strings): The criterion demands "exact string match" or "no formatting differences" for Dates, Coordinates, or Floating Point Numbers.
        * Bad: "Accept only exact string match." (for a Date like '2020-01-01').
        * Good: "Accept dates representing the same day, regardless of format (e.g., 'Jan 1st, 2020')."
    * Fail (Unrealistic Precision): The criterion for Coordinates or Statistics requires matching to extreme precision (e.g., >4 decimal places) which rarely appears in web text.
        * Bad: "Reject if coordinate differs by 0.000001."
        * Good: "Accept if coordinates are geographically close (within ~1km)."
    * Fail (Missing Alias Logic): The criterion for Entities (e.g., Universities, Journals) fails to mention allowing "Aliases" or "Common Abbreviations".
        * Bad: "Accept only 'University of California, Los Angeles'."
        * Good: "Accept 'UCLA' or 'Univ. of Calif. LA' as valid matches."
    * Fail (Vague Instructions): The criterion is too generic to ensure consistency.
        * Bad: "Check if the answer is correct." (Too open-ended).
        * Good: "Accept exact matches or minor spelling variations; reject completely different entities."

## Output Format
Output a JSON object ONLY:
```json
{
  "status": "VALID" | "INVALID",
  "reasoning": "Concise explanation...",
  "flags": {
    "unnatural_phrasing": boolean,
    "logically_unsolvable": boolean,
    "gt_contradicts_commonsense": boolean,
    "temporal_or_wiki_issue": boolean,
    "bad_rubric": boolean
  },
  "suggestion": "Optional fix"
}
\end{lstlisting} \end{tcolorbox}
 
\subsection{WideSeekBench Statistics}

\paragraph{Scale of Target Information}
\label{app:domain_scale}

\begin{figure}[htbp]
    \centering
    \begin{subfigure}[b]{0.48\textwidth}
        \centering
        \includegraphics[width=\textwidth]{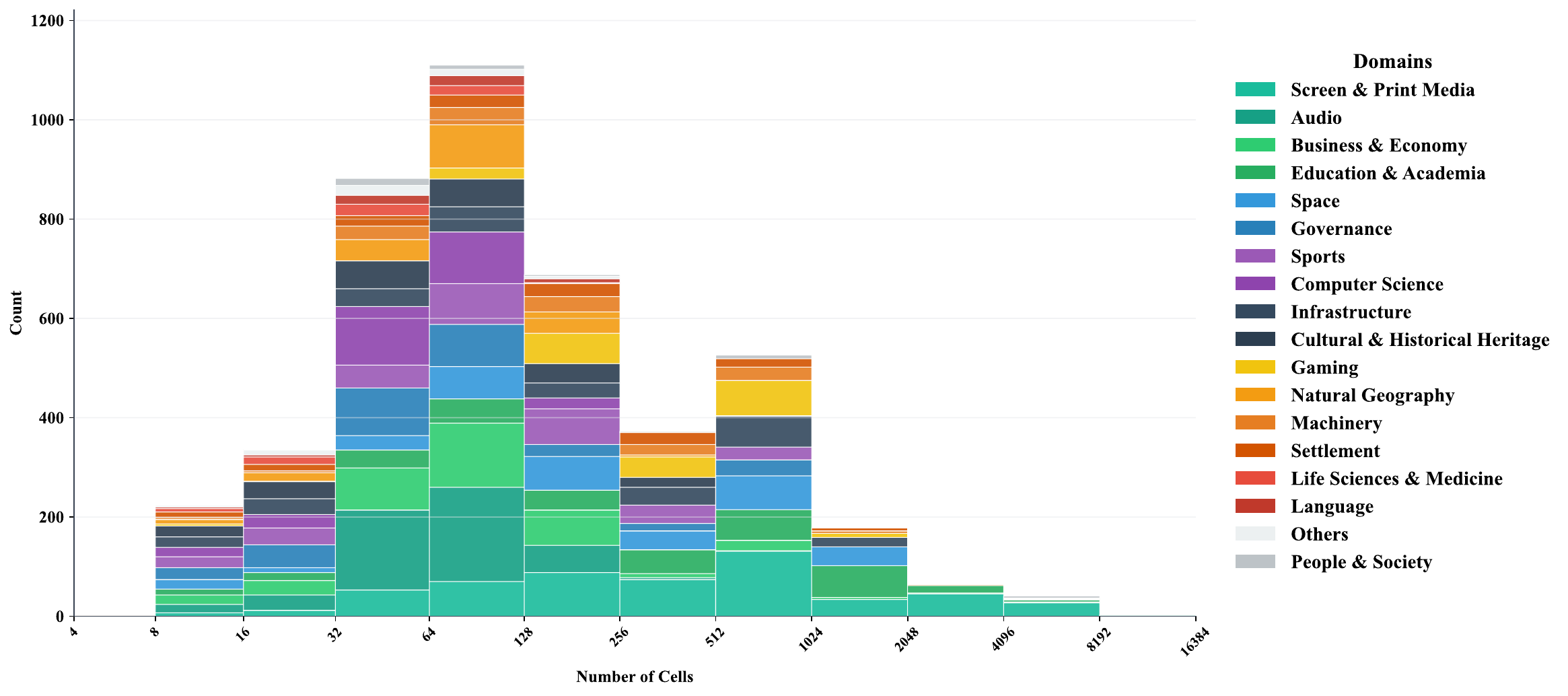}
        \caption{Train Set Scale}
        \label{fig:train_scale}
    \end{subfigure}
    \hfill
    \begin{subfigure}[b]{0.48\textwidth}
        \centering
        \includegraphics[width=\textwidth]{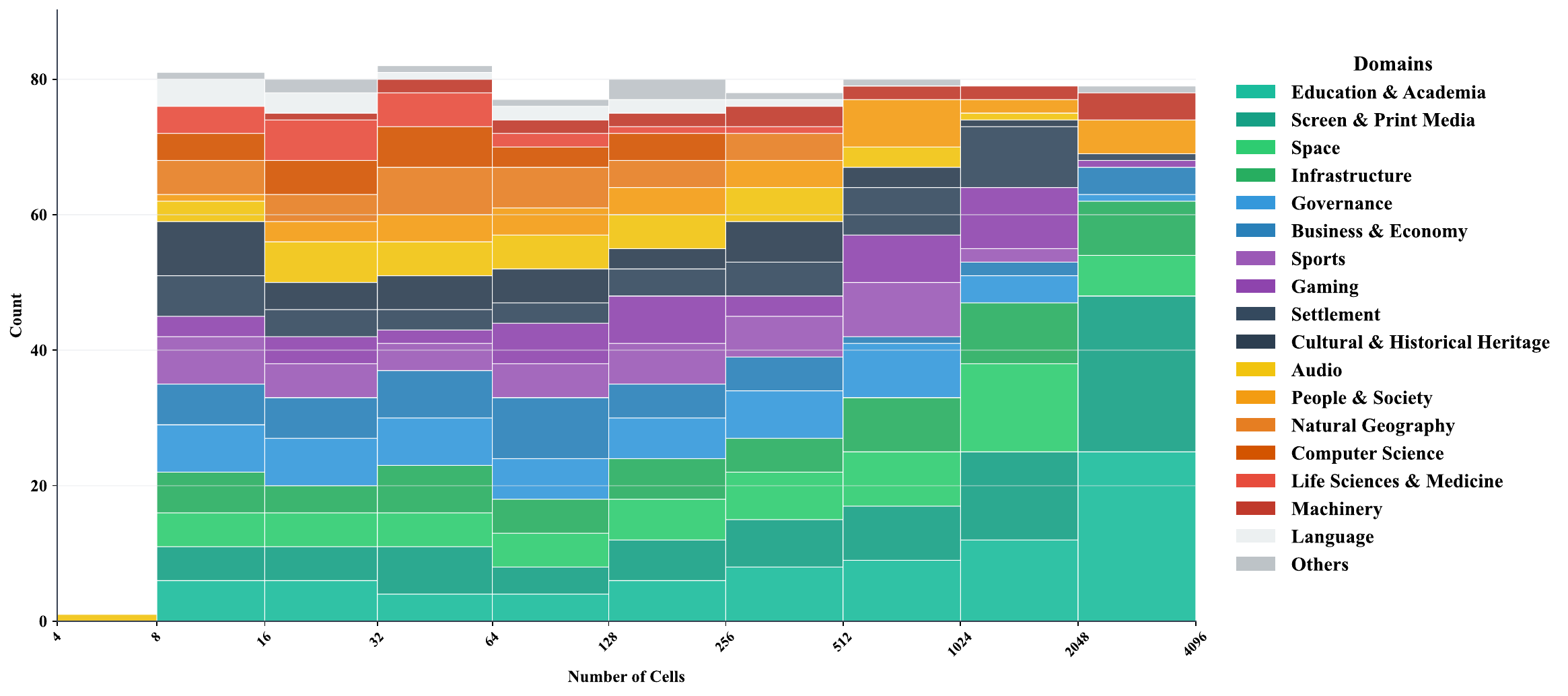}
        \caption{Test Set Scale}
        \label{fig:test_scale}
    \end{subfigure}
    
    \par\bigskip 

    \begin{subfigure}[b]{0.48\textwidth}
        \centering
        \includegraphics[width=\textwidth]{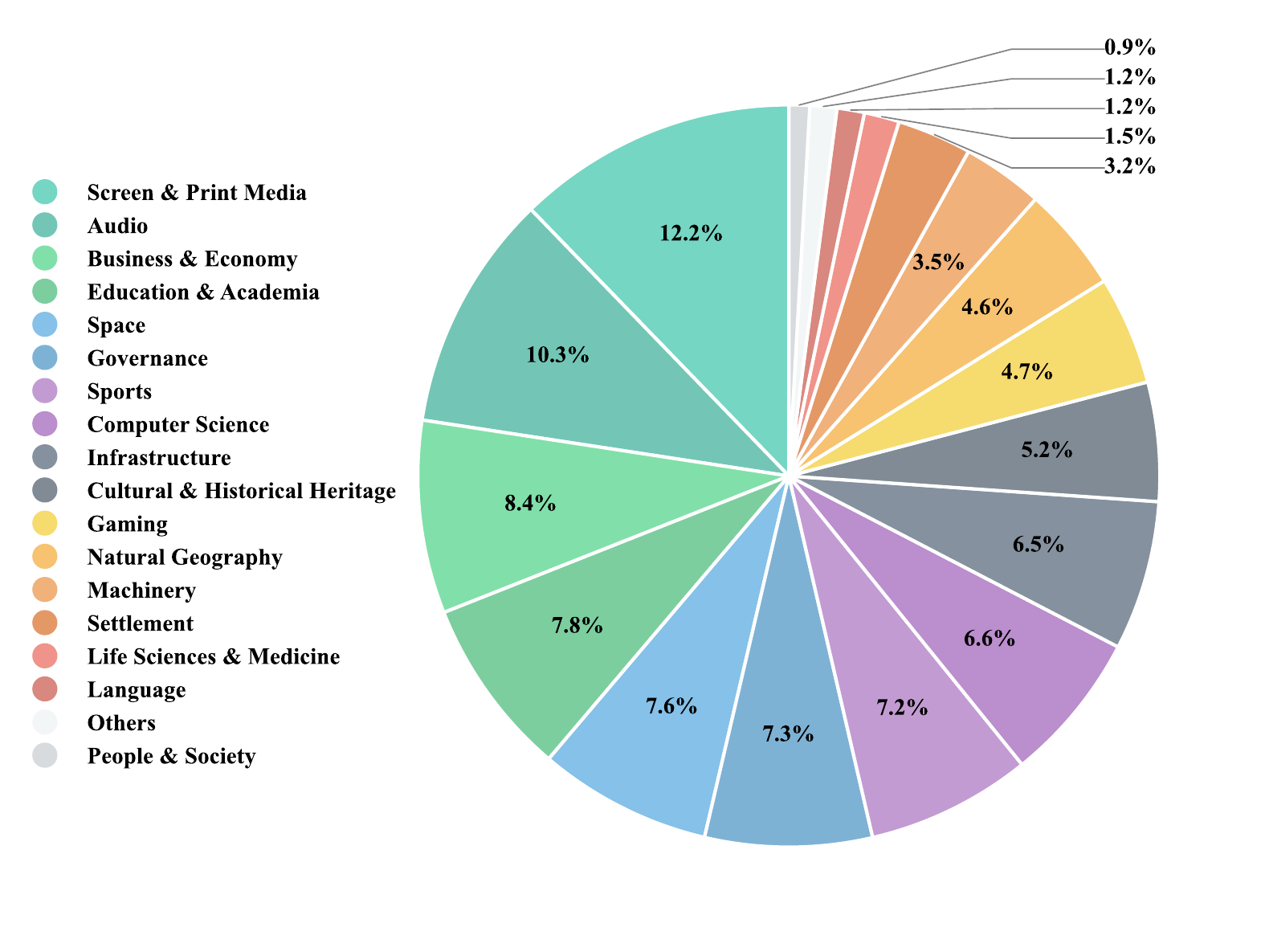}
        \caption{Train Set Domain Distribution}
        \label{fig:train_domain}
    \end{subfigure}
    \hfill
    \begin{subfigure}[b]{0.48\textwidth}
        \centering
        \includegraphics[width=\textwidth]{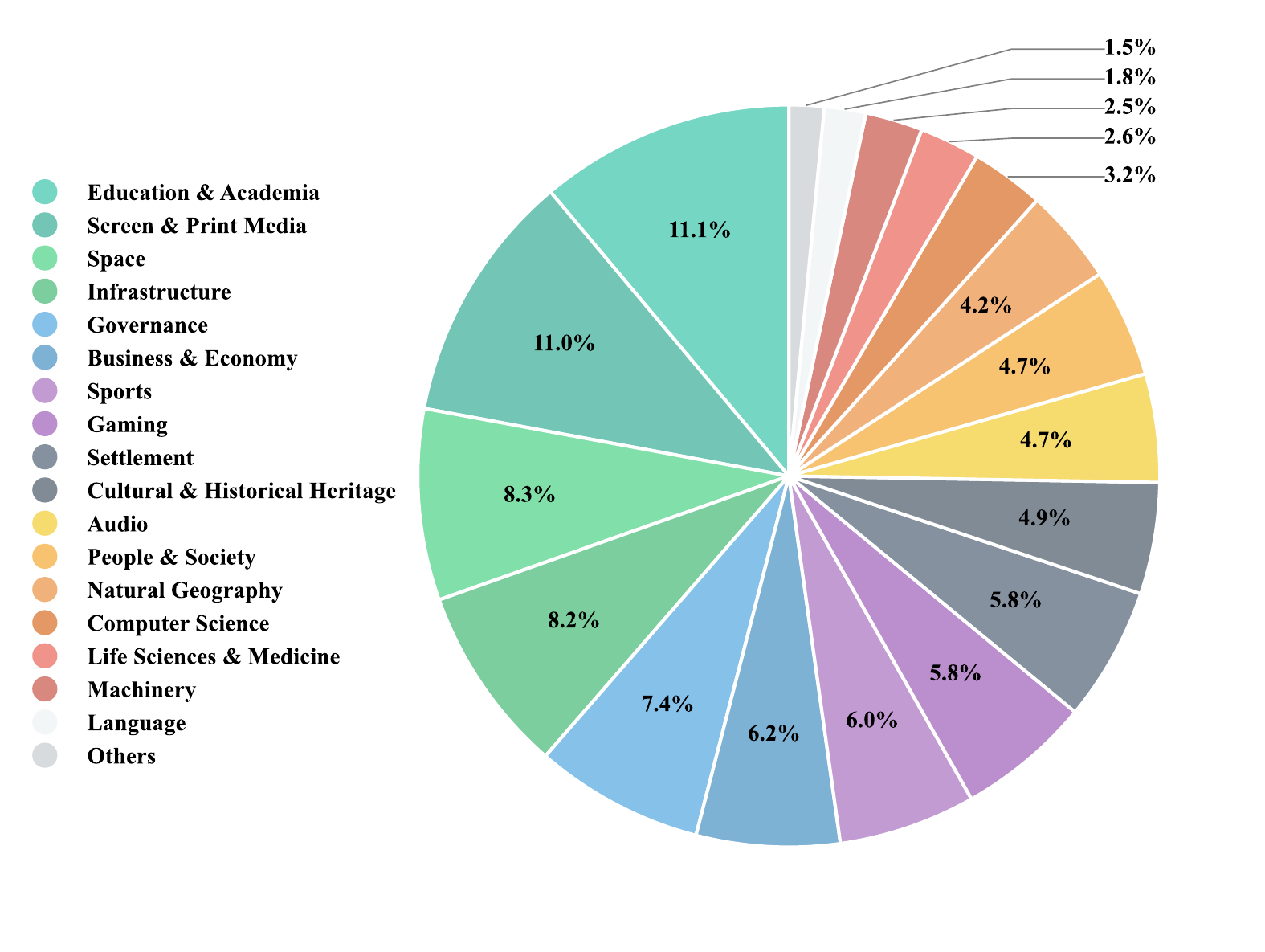}
        \caption{Test Set Domain Distribution}
        \label{fig:test_domain}
    \end{subfigure}

    \caption{Comprehensive statistics of WideSeekBench. Top row (a, b) illustrates the scale of target information. Bottom row (c, d) depicts the distribution of domains across Training and Test sets.}
    \label{fig:combined_stats}
\end{figure}

Figure~\ref{fig:combined_stats} depicts the scale of target information across diverse top-level domains. The dataset contains 4,436 training instances and 720 test instances, covering 18 domains. Tables~\ref{tab:train_details} and~\ref{tab:test_details} provide detailed distributions of subdomains. High-frequency categories in the training set include \textit{film} (252), \textit{video game} (197), and \textit{airport} (176). The test set preserves a similar distribution (e.g., \textit{film} 41, \textit{video game} 38).

\begin{table}[htbp]
\centering
\caption{Training set domain and sub-domain summary.}
\footnotesize
\resizebox{\textwidth}{!}{%
\begin{tabular}{c c c | c c c | c c c | c c c}
\toprule
\textbf{Domain} & \textbf{Subdomain} & \textbf{Count} & \textbf{Domain} & \textbf{Subdomain} & \textbf{Count} & \textbf{Domain} & \textbf{Subdomain} & \textbf{Count} & \textbf{Domain} & \textbf{Subdomain} & \textbf{Count} \\
\midrule
Screen \& Print Media & film & 252 & Space & planetary nebula & 1 & Infrastructure & railway station & 26 & Machinery & vehicle & 42 \\
\cline{4-6}
 & short film & 109 & Governance & political party & 122 &  & controlled-access highway & 21 &  & vehicles and vehicle parts product & 26 \\
 & television series & 92 &  & charitable organization & 28 &  & lighthouse & 19 &  & tool & 17 \\
 & literary work & 32 &  & non-governmental organization & 18 &  & hotel & 13 &  & equipment & 11 \\
 & television program & 31 &  & polity & 17 &  & road & 10 &  & automobile model & 9 \\
 & publisher & 7 &  & government agency & 15 &  & power station & 5 &  & ship & 5 \\
 & comics & 6 &  & armed organization & 14 &  & wind farm & 3 &  & physical tool & 4 \\
\cline{10-12}
 & magazine & 5 &  & political organization & 13 &  & house & 2 & Settlement & town & 39 \\
 & episode & 5 &  & battle & 13 &  & building & 1 &  & municipality & 32 \\
 & periodical & 2 &  & international organization & 12 &  & industrial building & 1 &  & village & 22 \\
\cline{7-9}
 & poem & 2 &  & war & 11 & Cultural \& Historical Heritage & historical country & 56 &  & city & 20 \\
\cline{1-3}
Audio & single & 146 &  & former administrative territorial entity & 10 &  & tomb & 40 &  & neighborhood & 10 \\
 & album & 89 &  & treaty & 8 &  & ceremony & 15 &  & district & 9 \\
 & rock band & 86 &  & legal case & 7 &  & church building & 15 &  & province & 8 \\
 & song & 74 &  & organization & 6 &  & cultural heritage & 14 &  & human settlement & 3 \\
 & musical group & 40 &  & administrative territorial entity & 5 &  & museum & 12 &  & region & 1 \\
\cline{10-12}
 & orchestra & 18 &  & firearm & 5 &  & archaeological site & 10 & Life Sciences \& Medicine & taxon & 17 \\
 & Musical Work & 3 &  & public election & 5 &  & heritage & 10 &  & protein family & 9 \\
 & concert & 2 &  & executive branch & 3 &  & cultural property & 10 &  & hospital & 8 \\
 & rock & 1 &  & conflict & 3 &  & architectural heritage monument & 9 &  & mammal & 6 \\
\cline{1-3}
Business \& Economy & bank & 73 &  & association & 3 &  & shrine & 9 &  & Chordata & 6 \\
 & public company & 72 &  & legal norm & 3 &  & heritage site & 9 &  & fungi & 6 \\
 & goods & 70 &  & crime & 2 &  & temple & 6 &  & Vertebrata & 4 \\
\cline{4-6}
 & manufactured good & 56 & Sports & sporting event & 76 &  & location of worship & 6 &  & medication & 3 \\
 & enterprise & 18 &  & sports season & 64 &  & funerary structure & 4 &  & anatomical structure & 2 \\
 & stock exchange & 17 &  & competition stage & 36 &  & structure of worship & 3 &  & bird & 2 \\
 & business & 17 &  & association football club & 28 &  & chapel & 1 &  & disease & 1 \\
 & brewery & 13 &  & competition & 22 &  & cemetery & 1 &  & enzyme & 1 \\
\cline{7-9}
 & brand & 11 &  & recurring sporting event edition & 17 & Gaming & video game & 197 &  & insect & 1 \\
 & company & 9 &  & recurring sporting event & 16 &  & electronic game & 12 &  & plant & 1 \\
 & trademark & 8 &  & racing & 12 &  & board game & 1 &  & anomaly & 1 \\
\cline{7-9} \cline{10-12}
 & currency & 8 &  & Olympic Games & 11 & Natural Geography & national park & 38 & Language & language & 43 \\
 & farm & 1 &  & physical activity & 7 &  & mountain & 32 &  & languoid & 9 \\
\cline{1-3}
Education \& Academia & university & 143 &  & sports venue & 5 &  & island & 27 &  & language variety & 1 \\
\cline{10-12}
 & college & 117 &  & sports competition & 5 &  & lake & 18 & Others & visual artwork & 13 \\
 & scientific journal & 25 &  & association football match & 5 &  & protected area & 18 &  & flag & 11 \\
 & research institute & 25 &  & tennis tournament & 5 &  & canal & 14 &  & dish & 8 \\
 & academic journal & 13 &  & sport & 4 &  & park & 11 &  & data & 4 \\
 & educational institution & 6 &  & nation at sport competition & 4 &  & disaster & 7 &  & artificial physical object & 3 \\
 & laboratory & 6 &  & baseball player & 1 &  & glacier & 7 &  & physical process & 2 \\
 & school & 6 &  & sports club & 1 &  & landform & 6 &  & philosophy & 2 \\
\cline{4-6}
 & library & 5 & Computer Science & programming language & 114 &  & earthquake & 6 &  & knowledge organization system & 2 \\
\cline{1-3}
Space & airport & 176 &  & operating system & 94 &  & natural heritage & 5 &  & sculpture & 2 \\
 & space mission & 47 &  & free software & 35 &  & hill & 3 &  & communications media & 2 \\
 & artificial satellite & 34 &  & computer & 28 &  & valley & 3 &  & assembly & 1 \\
 & rocket launch & 31 &  & computer network protocol & 10 &  & forest & 3 &  & chemical process & 1 \\
 & asteroid & 25 &  & software & 7 &  & nature reserve & 3 &  & disposable product & 1 \\
\cline{10-12}
 & aircraft model & 10 &  & database & 3 &  & watercourse & 2 & People \& Society & human & 30 \\
\cline{4-6}
 & exoplanet & 8 & Infrastructure & metro station & 146 &  & mineral & 1 &  & ethnic group & 9 \\
\cline{7-9}
 & variable star & 3 &  & dam & 42 & Machinery & machine & 43 &  & occupation & 1 \\
\midrule
\multicolumn{12}{c}{\textbf{Total: 4436}} \\
\bottomrule
\end{tabular}
}
\label{tab:train_details}
\end{table}
\begin{table}[htbp]
\centering
\caption{Test set domain and sub-domain summary.}
\footnotesize
\resizebox{\textwidth}{!}{%
\begin{tabular}{c c c | c c c | c c c | c c c}
\toprule
\textbf{Domain} & \textbf{Subdomain} & \textbf{Count} & \textbf{Domain} & \textbf{Subdomain} & \textbf{Count} & \textbf{Domain} & \textbf{Subdomain} & \textbf{Count} & \textbf{Domain} & \textbf{Subdomain} & \textbf{Count} \\
\midrule
Education \& Academia & college & 34 & Governance & former administrative territorial entity & 3 & Settlement & human settlement & 4 & Natural Geography & natural heritage & 1 \\
 & university & 26 &  & charitable organization & 2 &  & village & 3 &  & lake & 1 \\
 & research institute & 5 &  & battle & 2 &  & city & 3 &  & forest & 1 \\
\cline{10-12}
 & laboratory & 4 &  & political organization & 2 &  & province & 2 & Computer Science & operating system & 11 \\
 & school & 3 &  & government agency & 2 &  & neighborhood & 2 &  & programming language & 8 \\
 & academic journal & 3 &  & non-governmental organization & 1 &  & region & 2 &  & free software & 1 \\
 & educational institution & 3 &  & conflict & 1 &  & district & 1 &  & computer network protocol & 1 \\
\cline{7-9}
 & scientific journal & 1 &  & international organization & 1 & Cultural \& Historical Heritage & historical country & 11 &  & computer & 1 \\
 & library & 1 &  & war & 1 &  & church building & 6 &  & archive & 1 \\
\cline{1-3} \cline{4-6} \cline{10-12}
Screen \& Print Media & film & 41 & Business \& Economy & bank & 9 &  & ceremony & 4 & Life Sciences \& Medicine & hospital & 6 \\
 & short film & 16 &  & public company & 8 &  & historical event & 3 &  & protein family & 4 \\
 & television series & 12 &  & manufactured good & 7 &  & architectural heritage monument & 2 &  & fungi & 2 \\
 & television program & 6 &  & stock exchange & 5 &  & cultural heritage & 2 &  & symptom & 2 \\
 & literary work & 2 &  & brewery & 4 &  & tomb & 2 &  & Vertebrata & 1 \\
 & photograph & 1 &  & goods & 4 &  & heritage site & 2 &  & taxon & 1 \\
 & magazine & 1 &  & enterprise & 3 &  & museum & 1 &  & plant & 1 \\
\cline{1-3}
Space & airport & 31 &  & company & 2 &  & heritage & 1 &  & mammal & 1 \\
 & space mission & 12 &  & currency & 1 &  & cultural property & 1 &  & bird & 1 \\
\cline{7-9} \cline{10-12}
 & artificial satellite & 7 &  & business & 1 & People \& Society & human & 31 & Machinery & automobile model & 6 \\
 & aircraft model & 3 &  & brand & 1 &  & ethnic group & 3 &  & vehicle & 3 \\
\cline{4-6} \cline{7-9}
 & asteroid & 3 & Sports & sports season & 16 & Audio & musical group & 7 &  & machine & 3 \\
 & exoplanet & 2 &  & association football club & 6 &  & album & 6 &  & ship & 2 \\
 & rocket launch & 1 &  & sporting event & 6 &  & rock band & 6 &  & equipment & 2 \\
 & astronomical object & 1 &  & competition stage & 3 &  & single & 6 &  & vehicles and vehicle parts product & 1 \\
\cline{1-3}
Infrastructure & metro station & 35 &  & recurring sporting event & 3 &  & song & 5 &  & tool & 1 \\
\cline{10-12}
 & railway station & 7 &  & association football match & 2 &  & orchestra & 3 & Language & language & 8 \\
 & controlled-access highway & 5 &  & recurring sporting event edition & 2 &  & musician & 1 &  & human language & 4 \\
\cline{7-9}
 & hotel & 4 &  & sports competition & 2 & Natural Geography & national park & 6 &  & language variety & 1 \\
\cline{10-12}
 & dam & 3 &  & racing & 1 &  & island & 5 & Others & visual artwork & 4 \\
 & power station & 2 &  & Olympic Games & 1 &  & mountain & 5 &  & dish & 2 \\
 & lighthouse & 1 &  & sports venue & 1 &  & canal & 3 &  & flag & 2 \\
\cline{4-6}
 & wind farm & 1 & Gaming & video game & 38 &  & hill & 2 &  & unit of measurement & 1 \\
 & road & 1 &  & board game & 3 &  & landform & 2 &  & science & 1 \\
\cline{1-3}
Governance & political party & 31 &  & electronic game & 1 &  & protected area & 2 &  & artificial physical object & 1 \\
\cline{4-6} \cline{10-12}
 & armed organization & 4 & Settlement & town & 19 &  & watercourse & 1 &  & & \\
 & polity & 3 &  & municipality & 6 &  & park & 1 &  & & \\
\midrule
\multicolumn{12}{c}{\textbf{Total: 720}} \\
\bottomrule
\end{tabular}
}
\label{tab:test_details}
\end{table}

\paragraph{Constraint Complexity}
Table~\ref{tab:constraint_complexity} shows the distribution of logical patterns in the dataset, which directly reflects the distribution of constraints. The training set is dominated by single-type patterns, with pure conjunctions (\texttt{AND}) accounting for $37.8\%$, followed by \texttt{AND\_NOT} ($19.5\%$). The test set exhibits a more balanced distribution, with simple \texttt{AND} patterns reduced to $20.0\%$ and complex composite patterns substantially increased. The most complex combination, \texttt{AND\_OR\_NOT}, constitutes $11.5\%$ of the test set (compared to $5.1\%$ in training), and other high-complexity patterns such as \texttt{AND\_OR} and \texttt{OR\_NOT} are also more evenly represented.

\begin{table}[htbp]
\centering
\caption{Distribution of logical patterns in WideSeekBench.}
\small
\begin{tabular}{l | c c | c c}
\toprule
\multirow{2}{*}{\textbf{Patterns}} & \multicolumn{2}{c|}{\textbf{Training Set}} & \multicolumn{2}{c}{\textbf{Test Set}} \\
& \textbf{Count} & \textbf{Percentage} & \textbf{Count} & \textbf{Percentage} \\
\midrule
\texttt{AND} & 1,676 & 37.8\% & 144 & 20.0\% \\
\texttt{AND\_NOT} & 866 & 19.5\% & 119 & 16.5\% \\
\texttt{AND\_OR} & 704 & 15.9\% & 104 & 14.4\% \\
\texttt{OR} & 502 & 11.3\% & 93 & 12.9\% \\
\texttt{NOT} & 233 & 5.3\% & 94 & 13.1\% \\
\texttt{OR\_NOT} & 229 & 5.2\% & 83 & 11.5\% \\
\texttt{AND\_OR\_NOT} & 226 & 5.1\% & 83 & 11.5\% \\
\midrule
\textbf{Total} & \textbf{4,436} & \textbf{100.0\%} & \textbf{720} & \textbf{100.0\%} \\
\bottomrule
\end{tabular}
\label{tab:constraint_complexity}
\end{table}

\begin{figure}[htbp]
    \centering
    \begin{subfigure}[b]{0.48\textwidth}
        \centering
        \includegraphics[width=\textwidth]{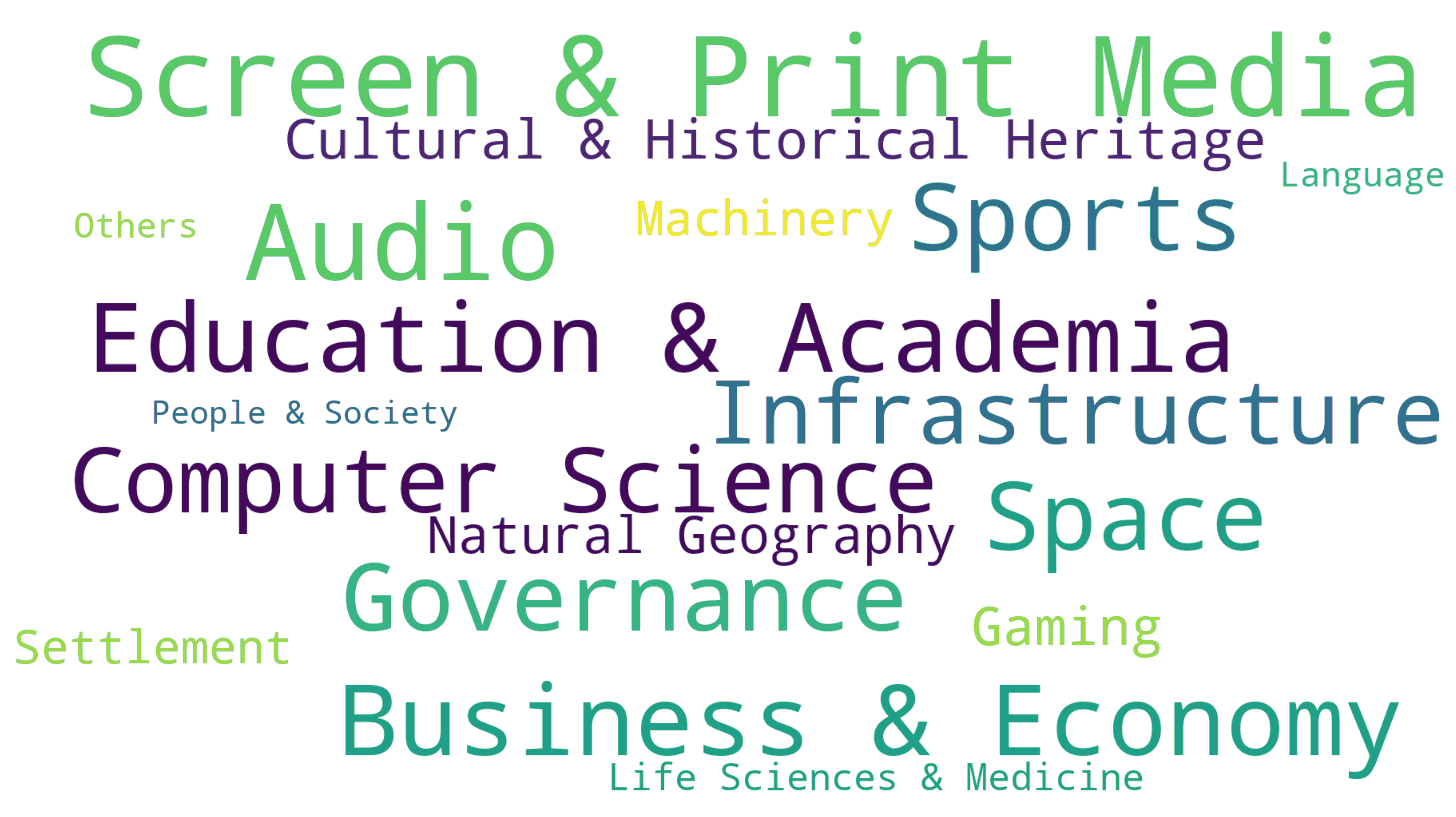}
        \caption{Train: Domain Distribution}
    \end{subfigure}
    \hfill
    \begin{subfigure}[b]{0.48\textwidth}
        \centering
        \includegraphics[width=\textwidth]{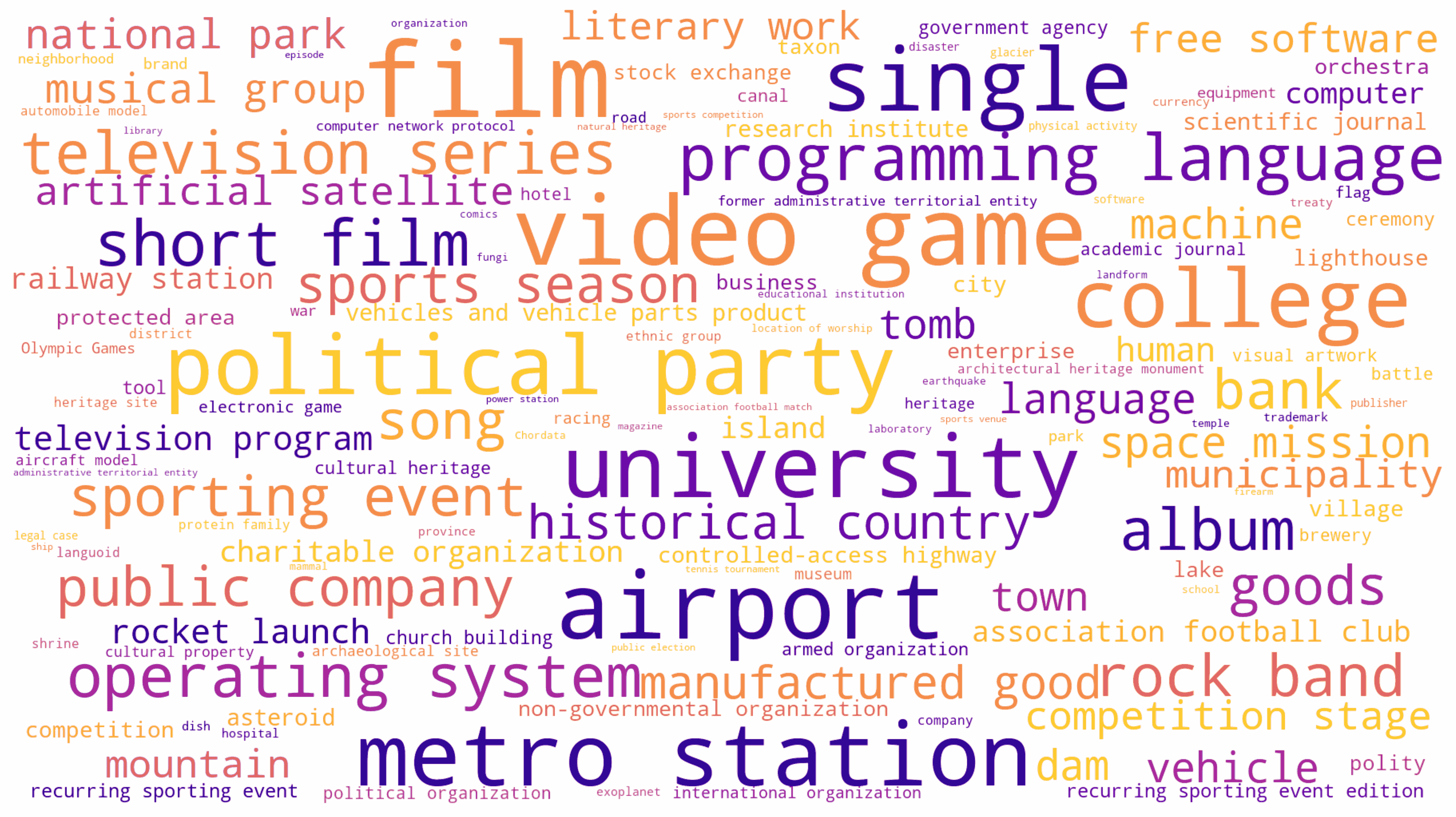}
        \caption{Train: Subdomain Distribution}
    \end{subfigure}
    
    \par\bigskip 
    
    \begin{subfigure}[b]{0.48\textwidth}
        \centering
        \includegraphics[width=\textwidth]{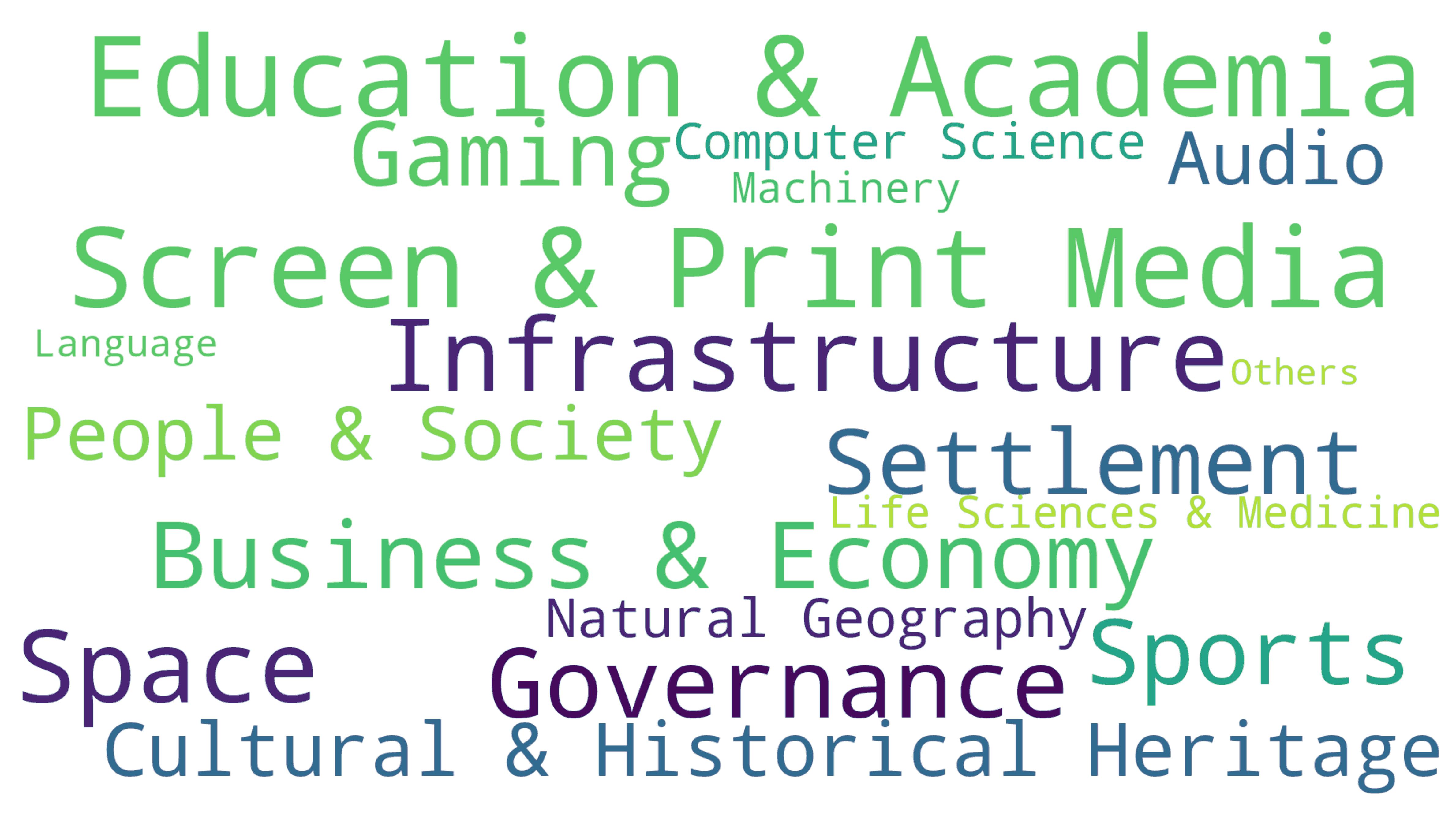}
        \caption{Test: Domain Distribution}
    \end{subfigure}
    \hfill
    \begin{subfigure}[b]{0.48\textwidth}
        \centering
        \includegraphics[width=\textwidth]{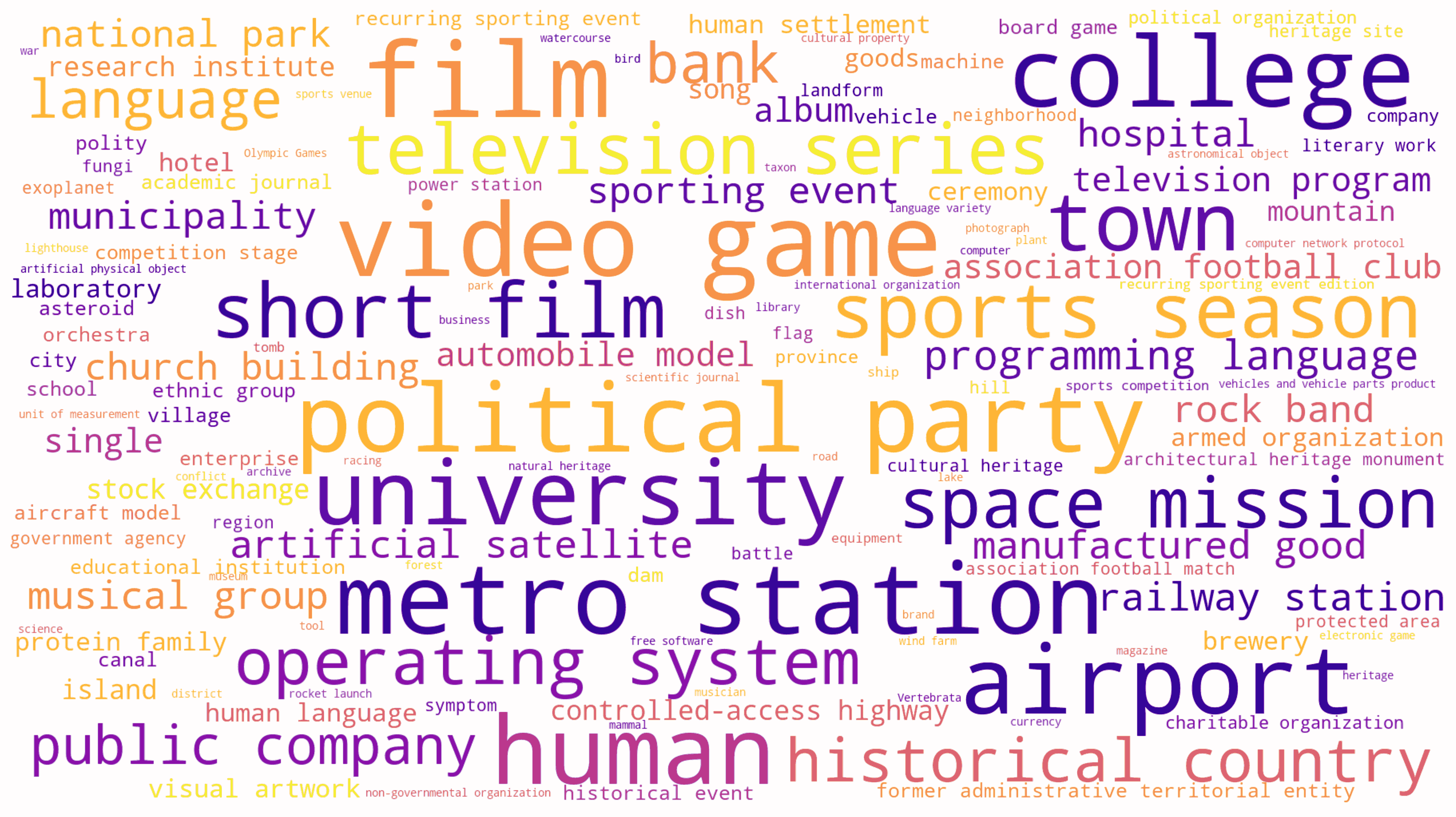}
        \caption{Test: Subdomain Distribution}
    \end{subfigure}
    
    \caption{Word clouds illustrating the diversity of WideSeekBench. Top row (a, b) shows the Training set, and bottom row (c, d) shows the Test set. The size of each term corresponds to its frequency.}
    \label{fig:wordclouds}
\end{figure}

\paragraph{Domain Diversity}
Figure~\ref{fig:wordclouds} shows the distribution of topics in WideSeekBench across training and test sets. Dominant domains such as \textit{Screen \& Print Media} and \textit{Gaming} are represented by subdomains including \textit{film} and \textit{video game}. Scientific and technical sectors are also covered, notably \textit{Space} (e.g., \textit{airport}, \textit{space mission}) and \textit{Infrastructure} (e.g., \textit{metro station}). The dataset exhibits a long-tailed distribution that includes specialized concepts ranging from \textit{Life Sciences} (e.g., \textit{protein family}, \textit{enzyme}) to \textit{Natural Geography} features (e.g., \textit{planetary nebula}, \textit{glacier}). The test set (Figures~\ref{fig:wordclouds}c and \ref{fig:wordclouds}d) maintains a similar distribution across domains.

\subsection{Task Cases}
\begin{tcolorbox}[
    colback=blue!5!white,
    colframe=blue!75!black,
    title={Example 1: \scalebox{1.3}{\textbf{\texttt{AND}}}},
    breakable,
    fonttitle=\bfseries,
    boxsep=3pt,
    left=5pt, right=5pt, top=5pt, bottom=5pt
]

\textbf{Domain:} Education \& Academia \\
\textbf{Sub-domain:} scientific journal \\
\vspace{0.2cm}
\rule{\textwidth}{0.4pt}
\vspace{0.2cm}

\textbf{Query:} 
\begin{lstlisting}[
    basicstyle=\ttfamily\scriptsize,
    breaklines=true,
    numbers=none,
    frame=none,
    aboveskip=0pt,
    belowskip=0pt
]
Curious about scientific journals published by Wiley that focus on geography and either use English as their work language or have an English name.

Please find scientific journal that meet these conditions, and provide the following attributes for these scientific journal: scientific journal, inception, country of origin, indexed in bibliographic review, Danish Bibliometric Research Indicator level

**Column Specifications:**
1. **scientific journal**
   * Precise Explanation: A periodical journal that publishes peer-reviewed scientific research focused on geography
   * Format Specification: Full official title of the journal in standard title case
   * Example: "Journal of Coastal Geographic Studies"

(15 lines skipped)

5. **Danish Bibliometric Research Indicator level**
   * Precise Explanation: The official ranking level assigned to the journal under the Danish Bibliometric Research Indicator system for academic publication channels
   * Format Specification: Single integer value (either 1 or 2)
   * Example: "1"

Try to provide all information. Only use '/' when you believe the information is truly unavailable or cannot be obtained.

Please organize the results in one Markdown table with the following columns: scientific journal, inception, country of origin, indexed in bibliographic review, Danish Bibliometric Research Indicator level.

Don't ask me any questions, just output the results according to the columns without omitting cells arbitrarily. The output format is
```markdown
{data_content}
```.
\end{lstlisting}
\vspace{0.2cm}
\rule{\textwidth}{0.4pt}
\vspace{0.2cm}

\textbf{SPARQL Query:}

\begin{lstlisting}[
    basicstyle=\ttfamily\scriptsize,
    breaklines=true,
    numbers=none,
    frame=none,
    aboveskip=0pt,
    belowskip=0pt
]
PREFIX wd: <http://www.wikidata.org/entity/>
PREFIX wdt: <http://www.wikidata.org/prop/direct/>

SELECT DISTINCT ?item WHERE {
  ?item wdt:P31/(wdt:P279|wdt:P1647)* wd:Q5633421 .
  ?item wdt:P123/(wdt:P279|wdt:P1647)* wd:Q1479654 .
  ?item wdt:P407/(wdt:P279|wdt:P1647)* wd:Q1860 .
  ?item wdt:P921/(wdt:P279|wdt:P1647)* wd:Q1071 .
}
\end{lstlisting}
\vspace{0.2cm}
\rule{\textwidth}{0.4pt}
\vspace{0.2cm}

\textbf{Table:}

\vspace{0.1cm}
\scriptsize
\setlength{\arrayrulewidth}{0.3pt}
\setlength{\tabcolsep}{3pt}
\begin{tabularx}{\textwidth}{|X|X|X|X|X|}
\hline
scientific journal & inception & country of origin & indexed in bibliographic review & Danish Bibliometric Research Indicator level \\
\hline
The Geographical Journal & 1831-01-01T00:00:00Z & United Kingdom & Social Sciences Citation Index; Scopus & 1 \\[0.3pt]
\hline
The Canadian Geographer & 1950-01-01T00:00:00Z & Canada & Social Sciences Citation Index; Scopus & 1 \\[0.3pt]
\hline
Geographical Analysis & 1969-01-01T00:00:00Z & United States & Social Sciences Citation Index; Scopus & 1 \\[0.3pt]
\hline
Journal of Biogeography & 1974-01-01T00:00:00Z & United Kingdom & Science Citation Index Expanded; Scopus & 2 \\[0.3pt]
\hline
Transactions of the Institute of British Geographers & 1935-01-01T00:00:00Z & United Kingdom & Scopus & 2 \\[0.3pt]
\hline
Antipode & 1969-01-01T00:00:00Z & United Kingdom & Social Sciences Citation Index; Scopus & 2 \\[0.3pt]
\hline
Area & 1969-01-01T00:00:00Z & United Kingdom & Social Sciences Citation Index; Scopus & 1 \\[0.3pt]
\hline
International Journal of Urban and Regional Research & 1977-01-01T00:00:00Z & United Kingdom & Social Sciences Citation Index; Scopus & 2 \\[0.3pt]
\hline
International Migration & 1961-01-01T00:00:00Z & United Kingdom & Social Sciences Citation Index; Scopus & 1 \\[0.3pt]
\hline
Natural Resources Forum & 1976-01-01T00:00:00Z & Kingdom of the Netherlands & Science Citation Index Expanded; Scopus & 1 \\[0.3pt]
\hline
\multicolumn{5}{|c|}{\textit{(showing first 10 rows, total: 11 rows)}} \\
\hline
\end{tabularx}
\normalsize
\vspace{0.2cm}

\rule{\textwidth}{0.4pt}
\vspace{0.2cm}

\textbf{Rubrics:}

\vspace{0.1cm}
\small
\setlength{\arrayrulewidth}{0.3pt}
\setlength{\tabcolsep}{3pt}
\begin{tabularx}{\textwidth}{|p{2.5cm}|p{2cm}|X|}
\hline
\textbf{Column} & \textbf{Metric} & \textbf{Criterion} \\
\hline
scientific journal & llm\_judge & Accept exact journal names, common recognized abbreviations, and minor title variations (such as omitting leading articles like 'The' or adjusting capitalization) as long as they refer to the same published journal. Reject completely distinct journal titles that do not correspond to the same publication. \\
\hline
inception & llm\_judge & Accept dates that represent the same year, month, and day regardless of string format, including ignoring timestamp suffixes like T00:00:00Z which do not alter the core date value (e.g., '1831-01-01T00:00:00Z' matches 'January 1, 1831' or '1831'). Reject dates referring to a different year, month, or day. \\
\hline
country of origin & llm\_judge & Accept the exact official country name, common colloquial aliases, and official alternative names (e.g., 'United Kingdom' matches 'UK', 'Kingdom of the Netherlands' matches 'Netherlands' or 'Holland'). Reject country names that refer to an entirely different sovereign nation. \\
\hline
indexed in bibliographic review & llm\_judge & Accept lists that contain the same set of indexing services regardless of order, punctuation separators (semicolon vs comma), or minor spacing differences around separators. For example, 'Social Sciences Citation Index; Scopus' matches 'Scopus, Social Sciences Citation Index'. Reject lists that have missing or additional indexing services not present in the reference value. \\
\hline
Danish Bibliometric Research Indicator level & llm\_judge & Accept values that are numerically equivalent to the reference level, regardless of formatting (e.g., the string '2' matches the number 2 or the phrase 'Level 2') as long as they refer to the correct indicator level. Reject values that correspond to a different integer level entirely. \\
\hline
\end{tabularx}
\normalsize
\end{tcolorbox}

\vspace{0.5cm}

\begin{tcolorbox}[
    colback=blue!5!white,
    colframe=blue!75!black,
    title={Example 2: \scalebox{1.3}{\textbf{\texttt{AND\_NOT}}}},
    breakable,
    fonttitle=\bfseries,
    boxsep=3pt,
    left=5pt, right=5pt, top=5pt, bottom=5pt
]

\textbf{Domain:} Natural Geography \\
\textbf{Sub-domain:} lake \\
\vspace{0.2cm}
\rule{\textwidth}{0.4pt}
\vspace{0.2cm}

\textbf{Query:} 
\begin{lstlisting}[
    basicstyle=\ttfamily\scriptsize,
    breaklines=true,
    numbers=none,
    frame=none,
    aboveskip=0pt,
    belowskip=0pt
]
What are all the lakes located in the Northwest Territories, with a basin country of Canada, excluding those that outflow into the Buffalo River?
Please find lake that meet these conditions, and provide the following attributes for these lake: lake, elevation above sea level, coordinate location, country

**Column Specifications:**
1. **lake**
   - Precise Explanation: The formal, titled name of a contained body of still water situated within a natural basin
   - Format Specification: Proper noun written in title case, may include hyphens for connected lake systems
   - Example: "Silvermoss Lake"

(10 lines skipped)

4. **country**
   - Precise Explanation: The sovereign state or recognized territorial entity containing the lake; multiple entries separated by semicolons for cross-border lakes
   - Format Specification: Proper noun country or territorial entity names, separated by semicolons when applicable
   - Example: "Canada; Deninu First Nation"

Try to provide all information. Only use '/' when you believe the information is truly unavailable or cannot be obtained.

Please organize the results in one Markdown table with the following columns: lake, elevation above sea level, coordinate location, country.

Don't ask me any questions, just output the results according to the columns without omitting cells arbitrarily. The output format is
```markdown
{data_content}
```.
\end{lstlisting}
\vspace{0.2cm}
\rule{\textwidth}{0.4pt}
\vspace{0.2cm}

\textbf{SPARQL Query:}

\begin{lstlisting}[
    basicstyle=\ttfamily\scriptsize,
    breaklines=true,
    numbers=none,
    frame=none,
    aboveskip=0pt,
    belowskip=0pt
]
PREFIX wd: <http://www.wikidata.org/entity/>
PREFIX wdt: <http://www.wikidata.org/prop/direct/>

SELECT DISTINCT ?item WHERE {
  ?item wdt:P31/(wdt:P279|wdt:P1647)* wd:Q23397 .
  ?item wdt:P131/(wdt:P279|wdt:P1647)* wd:Q2007 .
  ?item wdt:P205/(wdt:P279|wdt:P1647)* wd:Q16 .
  FILTER NOT EXISTS { ?item wdt:P201/(wdt:P279|wdt:P1647)* wd:Q22605327 . }
}
\end{lstlisting}
\vspace{0.2cm}
\rule{\textwidth}{0.4pt}
\vspace{0.2cm}

\textbf{Table:}

\vspace{0.1cm}
\scriptsize
\setlength{\arrayrulewidth}{0.3pt}
\setlength{\tabcolsep}{3pt}
\begin{tabularx}{\textwidth}{|X|X|X|X|}
\hline
lake & elevation above sea level & coordinate location & country \\
\hline
Clinton-Colden Lake & 375.0 & POINT(-107.466667 63.966667) & Canada \\[0.3pt]
\hline
Colville Lake & 244.0 & POINT(-126.000000 67.166667) & Canada \\[0.3pt]
\hline
Point Lake & 375.0 & POINT(-113.067962 65.250299) & Canada \\[0.3pt]
\hline
Dubawnt Lake & 236.0 & POINT(-102.096400 63.062790); POINT(-101.533333 63.100000) & Canada \\[0.3pt]
\hline
Keller Lake & 255.0 & POINT(-121.584942 63.950142) & Canada \\[0.3pt]
\hline
Artillery Lake & 365.0 & POINT(-107.866667 63.166667) & Canada \\[0.3pt]
\hline
Faber Lake & 213.0 & POINT(-117.466667 63.883333) & Canada \\[0.3pt]
\hline
Lac La Martre & 265.0 & POINT(-117.950000 63.316667) & Tlicho Nation; Canada \\[0.3pt]
\hline
Wholdaia Lake & 364.0 & POINT(-104.167297 60.716880) & Canada \\[0.3pt]
\hline
MacKay Lake & 431.0 & POINT(-111.033333 63.916667) & Canada \\[0.3pt]
\hline
\multicolumn{4}{|c|}{\textit{(showing first 10 rows, total: 34 rows)}} \\
\hline
\end{tabularx}
\normalsize
\vspace{0.2cm}

\rule{\textwidth}{0.4pt}
\vspace{0.2cm}

\textbf{Rubrics:}

\vspace{0.1cm}
\small
\setlength{\arrayrulewidth}{0.3pt}
\setlength{\tabcolsep}{3pt}
\begin{tabularx}{\textwidth}{|p{2.5cm}|p{2cm}|X|}
\hline
\textbf{Column} & \textbf{Metric} & \textbf{Criterion} \\
\hline
lake & llm\_judge & Accept the exact lake name, minor formatting variations like hyphen or spacing changes, and established common aliases for the lake. Reject names referring to an entirely different lake. \\
\hline
elevation above sea level & llm\_judge & Accept elevation values that are numerically equivalent or very close, allowing for reasonable rounding differences or minor measurement variations within ±1 meter or 1\% difference. Ignore formatting differences like trailing .0 or string versus numeric representation. \\
\hline
coordinate location & llm\_judge & Accept coordinate values that represent the same general locations, allowing minor positional deviations within approximately 1km. Ignore formatting differences like POINT() wrapping or separator type (comma vs semicolon), and accept sets of coordinates in any order as long as they refer to the same key locations for the lake. \\
\hline
country & llm\_judge & Accept country or recognized indigenous territorial entity names that correctly represent the lake's location, including combined entries separated by any separator in any order. Allow common aliases for countries or associated indigenous nations, and reject entirely unrelated country or territorial names. \\
\hline
\end{tabularx}
\normalsize
\end{tcolorbox}

\vspace{0.5cm}

\begin{tcolorbox}[
    colback=blue!5!white,
    colframe=blue!75!black,
    title={Example 3: \scalebox{1.3}{\textbf{\texttt{AND\_OR}}}},
    breakable,
    fonttitle=\bfseries,
    boxsep=3pt,
    left=5pt, right=5pt, top=5pt, bottom=5pt
]

\textbf{Domain:} Screen \& Print Media \\
\textbf{Sub-domain:} television series \\
\vspace{0.2cm}
\rule{\textwidth}{0.4pt}
\vspace{0.2cm}

\textbf{Query:} 
\begin{lstlisting}[
    basicstyle=\ttfamily\scriptsize,
    breaklines=true,
    numbers=none,
    frame=none,
    aboveskip=0pt,
    belowskip=0pt
]
What are all the television series that are either science fiction programs set on an island, or television series that are followed by Ashes to Ashes and have Ashes to Ashes as their spin-off?
Please find television series that meet these conditions, and provide the following attributes for these television series: television series, original broadcaster, start time, end time, country of origin, number of seasons, number of episodes

**Column Specifications:**
1.  **television series**
    *   Precise Explanation: A connected collection of episodes released under one unified official title
    *   Format Specification: Full official title of the television series
    *   Example: "Island of Echoes"

(20 lines skipped)

7.  **number of episodes**
    *   Precise Explanation: Total count of episodes that officially aired as part of the television series
    *   Format Specification: Decimal numeric value
    *   Example: "40.0"

Try to provide all information. Only use '/' when you believe the information is truly unavailable or cannot be obtained.

Please organize the results in one Markdown table with the following columns: television series, original broadcaster, start time, end time, country of origin, number of seasons, number of episodes.

Don't ask me any questions, just output the results according to the columns without omitting cells arbitrarily. The output format is
```markdown
{data_content}
```.
\end{lstlisting}
\vspace{0.2cm}
\rule{\textwidth}{0.4pt}
\vspace{0.2cm}

\textbf{SPARQL Query:}

\begin{lstlisting}[
    basicstyle=\ttfamily\scriptsize,
    breaklines=true,
    numbers=none,
    frame=none,
    aboveskip=0pt,
    belowskip=0pt
]
PREFIX wd: <http://www.wikidata.org/entity/>
PREFIX wdt: <http://www.wikidata.org/prop/direct/>

SELECT DISTINCT ?item WHERE {
  ?item wdt:P31/(wdt:P279|wdt:P1647)* wd:Q5398426 .
  {
  {
    ?item wdt:P136/(wdt:P279|wdt:P1647)* wd:Q336059 .
    ?item wdt:P8411/(wdt:P279|wdt:P1647)* wd:Q23442 .
  }
  UNION
  {
    ?item wdt:P156/(wdt:P279|wdt:P1647)* wd:Q725195 .
    ?item wdt:P2512/(wdt:P279|wdt:P1647)* wd:Q725195 .
  }
  }
}
\end{lstlisting}
\vspace{0.2cm}
\rule{\textwidth}{0.4pt}
\vspace{0.2cm}

\textbf{Table:}

\vspace{0.1cm}
\scriptsize
\setlength{\arrayrulewidth}{0.3pt}
\setlength{\tabcolsep}{3pt}
\begin{tabularx}{\textwidth}{|X|X|X|X|}
\hline
television series & start time & country of origin & number of episodes \\
\hline
Intergalactic & 2021-04-30T00:00:00Z & United Kingdom & 8.0 \\[0.3pt]
\hline
Life on Mars & 2006-01-09T00:00:00Z & United Kingdom & 16.0 \\[0.3pt]
\hline
Department S & 1969-03-09T00:00:00Z & United Kingdom & 28.0 \\[0.3pt]
\hline
The Hoobs & 2001-01-15T00:00:00Z; 2009-10-05T00:00:00Z & United Kingdom; United States & / \\[0.3pt]
\hline
Survivors & 1975-04-16T00:00:00Z & United Kingdom & 38.0 \\[0.3pt]
\hline
Summer in Transylvania & 2010-10-25T00:00:00Z & United Kingdom & 20.0 \\[0.3pt]
\hline
Chocky & 1984-01-01T00:00:00Z & United Kingdom & 18.0 \\[0.3pt]
\hline
Agent Carter & 2015-01-06T00:00:00Z & United States & 18.0 \\[0.3pt]
\hline
Sense8 & 2015-06-05T00:00:00Z & United States & 24.0 \\[0.3pt]
\hline
Alice & 2009-12-06T00:00:00Z & United Kingdom; Canada & 2.0 \\[0.3pt]
\hline
\multicolumn{4}{|c|}{\textit{(showing 4 of 7 columns), (showing first 10 rows, total: 41 rows)}} \\
\hline
\end{tabularx}
\normalsize
\vspace{0.2cm}

\rule{\textwidth}{0.4pt}
\vspace{0.2cm}

\textbf{Rubrics:}

\vspace{0.1cm}
\small
\setlength{\arrayrulewidth}{0.3pt}
\setlength{\tabcolsep}{3pt}
\begin{tabularx}{\textwidth}{|p{2.5cm}|p{2cm}|X|}
\hline
\textbf{Column} & \textbf{Metric} & \textbf{Criterion} \\
\hline
television series & llm\_judge & Accept the exact show title, widely recognized alternate spellings, official abbreviations, and titles with minor qualifying additions (like "(TV series)") that clearly refer to the same show. Reject completely unrelated television series names. \\
\hline
original broadcaster & llm\_judge & Accept exact broadcaster names, common official aliases and abbreviations (e.g., "ABC" = "American Broadcasting Company"), lists of broadcasters in any order regardless of separator punctuation, and ignore duplicate entries in the list. Confirm all key original broadcasters are present in the prediction; reject predictions missing major broadcasters or listing entirely unrelated ones. \\
\hline
end time & llm\_judge & Accept dates representing the exact same day, month, and year regardless of string format (e.g., ISO 8601 format like "2007-04-10T00:00:00Z" matches "April 10, 2007"). For entries with multiple end times, accept lists in any order regardless of separator punctuation, and confirm all listed end dates are present. Ignore time zone suffixes if the core date is accurate. Reject predictions with significantly mismatched end dates. \\
\hline
number of seasons & llm\_judge & Accept numeric values that are equivalent to the reference count, whether formatted as integer or decimal (e.g., 1 == 1.0). Allow for minor discrepancies only if they correspond to widely cited alternative season counts for the specific series; otherwise, reject values that differ by more than one season from the reference. \\
\hline
number of episodes & llm\_judge & Accept numeric values that are equivalent to the reference count, whether formatted as integer or decimal (e.g.,8 ==8.0). Allow for minor discrepancies only if they correspond to widely cited alternative episode counts for the specific series; otherwise, reject values that differ by more than five episodes from the reference. \\
\hline
\multicolumn{3}{|c|}{\textit{(showing 5 of 7 columns)}} \\
\hline
\end{tabularx}
\normalsize
\end{tcolorbox}

\vspace{0.5cm}

\begin{tcolorbox}[
    colback=blue!5!white,
    colframe=blue!75!black,
    title={Example 4: \scalebox{1.3}{\textbf{\texttt{AND\_OR\_NOT}}}},
    breakable,
    fonttitle=\bfseries,
    boxsep=3pt,
    left=5pt, right=5pt, top=5pt, bottom=5pt
]

\textbf{Domain:} Business \& Economy \\
\textbf{Sub-domain:} public company \\
\vspace{0.2cm}
\rule{\textwidth}{0.4pt}
\vspace{0.2cm}

\textbf{Query:} 
\begin{lstlisting}[
    basicstyle=\ttfamily\scriptsize,
    breaklines=true,
    numbers=none,
    frame=none,
    aboveskip=0pt,
    belowskip=0pt
]
Curious about all public companies that either operate in the retail industry with Derica W. Rice serving as a board member, or are based in the United States and are members of the Linux Foundation, excluding any companies that had the Target Corporation 2013 data breach as a significant event or have Bullseye as their mascot.

Please find public company that meet these conditions, and provide the following attributes for these public company: public company, headquarters location, stock exchange, legal form, inception

**Column Specifications:**
1.  **public company**
    *   Precise Explanation: A company that offers its securities for sale to the general public, allowing members of the public to invest and own shares in the business
    *   Format Specification: Properly formatted official company name, following standard business naming conventions
    *   Example: Northwood Retail Group

(15 lines skipped)

5.  **inception**
    *   Precise Explanation: The official date when the public company first came into existence, distinct from official opening or launch dates
    *   Format Specification: ISO 8601 date format (yyyy-mm-ddTHH:MM:SSZ), using semicolons to separate multiple inception dates if applicable
    *   Example: 2001-05-15T00:00:00Z; 2005-09-02T00:00:00Z

Try to provide all information. Only use '/' when you believe the information is truly unavailable or cannot be obtained.

Please organize the results in one Markdown table with the following columns: public company, headquarters location, stock exchange, legal form, inception.

Don't ask me any questions, just output the results according to the columns without omitting cells arbitrarily. The output format is
```markdown
{data_content}
```.
\end{lstlisting}
\vspace{0.2cm}
\rule{\textwidth}{0.4pt}
\vspace{0.2cm}

\textbf{SPARQL Query:}

\begin{lstlisting}[
    basicstyle=\ttfamily\scriptsize,
    breaklines=true,
    numbers=none,
    frame=none,
    aboveskip=0pt,
    belowskip=0pt
]
PREFIX wd: <http://www.wikidata.org/entity/>
PREFIX wdt: <http://www.wikidata.org/prop/direct/>

SELECT DISTINCT ?item WHERE {
  ?item wdt:P31/(wdt:P279|wdt:P1647)* wd:Q891723 .
  {
  {
    ?item wdt:P452/(wdt:P279|wdt:P1647)* wd:Q126793 .
    ?item wdt:P3320/(wdt:P279|wdt:P1647)* wd:Q109743489 .
  }
  UNION
  {
    ?item wdt:P17/(wdt:P279|wdt:P1647)* wd:Q30 .
    ?item wdt:P463/(wdt:P279|wdt:P1647)* wd:Q858851 .
  }
  }
  FILTER NOT EXISTS { { ?item wdt:P793/(wdt:P279|wdt:P1647)* wd:Q76313694 . } UNION { ?item wdt:P822/(wdt:P279|wdt:P1647)* wd:Q4997048 . } }
}
\end{lstlisting}
\vspace{0.2cm}
\rule{\textwidth}{0.4pt}
\vspace{0.2cm}

\textbf{Table:}

\vspace{0.1cm}
\scriptsize
\setlength{\arrayrulewidth}{0.3pt}
\setlength{\tabcolsep}{3pt}
\begin{tabularx}{\textwidth}{|X|X|X|X|X|}
\hline
public company & headquarters location & stock exchange & legal form & inception \\
\hline
CME Group & Chicago; London & Nasdaq & Delaware corporation & 2007-07-12T00:00:00Z \\[0.3pt]
\hline
Capital One & McLean & New York Stock Exchange & private limited company; Commercial Bank; Delaware corporation & 1994-01-01T00:00:00Z; 1994-07-21T00:00:00Z \\[0.3pt]
\hline
Juniper Networks & Sunnyvale & New York Stock Exchange & Delaware corporation & 1996-02-06T00:00:00Z \\[0.3pt]
\hline
Ciena & Hanover & New York Stock Exchange; Nasdaq & Delaware corporation & 1999-01-01T00:00:00Z \\[0.3pt]
\hline
Nuance Communications & Burlington & Nasdaq & Delaware corporation & 1992-01-01T00:00:00Z \\[0.3pt]
\hline
VMware & Palo Alto & New York Stock Exchange & corporation & 1998-01-01T00:00:00Z \\[0.3pt]
\hline
Adobe & San Jose & London Stock Exchange; Nasdaq & Delaware corporation & 1982-02-28T00:00:00Z \\[0.3pt]
\hline
Netflix, Inc. & Los Gatos & Nasdaq & Delaware corporation; public company & 1997-08-29T00:00:00Z \\[0.3pt]
\hline
Micron Technology & Boise & Nasdaq & Delaware corporation & 1978-10-01T00:00:00Z \\[0.3pt]
\hline
EPAM Systems & Newtown; Pennsylvania & New York Stock Exchange & Delaware corporation & 1993-01-01T00:00:00Z \\[0.3pt]
\hline
\multicolumn{5}{|c|}{\textit{(showing first 10 rows, total: 63 rows)}} \\
\hline
\end{tabularx}
\normalsize
\vspace{0.2cm}

\rule{\textwidth}{0.4pt}
\vspace{0.2cm}

\textbf{Rubrics:}

\vspace{0.1cm}
\small
\setlength{\arrayrulewidth}{0.3pt}
\setlength{\tabcolsep}{3pt}
\begin{tabularx}{\textwidth}{|p{2.5cm}|p{2cm}|X|}
\hline
\textbf{Column} & \textbf{Metric} & \textbf{Criterion} \\
\hline
public company & llm\_judge & Accept the exact company name, common aliases, standard suffix variations (e.g., 'Netflix' matches 'Netflix, Inc.'), and minor punctuation differences. Reject any entry referring to a completely unrelated public company. \\
\hline
headquarters location & llm\_judge & Accept if the set of headquarters locations matches, regardless of order or separator (comma vs semicolon). Allow common location aliases (e.g., 'PA' for 'Pennsylvania') and accept entries referring to the same general metropolitan or geographic area. Reject entries that list entirely unrelated locations. \\
\hline
stock exchange & llm\_judge & Accept if the set of stock exchanges matches, regardless of order or separator (comma vs semicolon). Allow standard abbreviations (e.g., 'NYSE' matches 'New York Stock Exchange', 'LSE' matches 'London Stock Exchange'). Reject entries that list entirely different stock exchanges. \\
\hline
legal form & llm\_judge & Accept semantic equivalents of the legal form(s), regardless of order or separator (comma vs semicolon). Allow abbreviations (e.g., 'Delaware corp' matches 'Delaware corporation') and equivalent terms (e.g., 'corporation' matches 'Delaware corporation' where the jurisdiction is implied). Reject entries that describe an entirely different legal structure unrelated to the reference. \\
\hline
inception & llm\_judge & Accept dates representing the same day, month, and year, regardless of string format (e.g., '2007-07-12' matches 'July 12, 2007'). Ignore trailing timestamp suffixes like 'T00:00:00Z' as non-essential. If multiple inception dates are listed, accept any matching equivalent date from the reference set. Reject entries with completely different inception dates. \\
\hline
\end{tabularx}
\normalsize
\end{tcolorbox}

\vspace{0.5cm}

\begin{tcolorbox}[
    colback=blue!5!white,
    colframe=blue!75!black,
    title={Example 5: \scalebox{1.3}{\textbf{\texttt{NOT}}}},
    breakable,
    fonttitle=\bfseries,
    boxsep=3pt,
    left=5pt, right=5pt, top=5pt, bottom=5pt
]

\textbf{Domain:} Gaming \\
\textbf{Sub-domain:} video game \\
\vspace{0.2cm}
\rule{\textwidth}{0.4pt}
\vspace{0.2cm}

\textbf{Query:} 
\begin{lstlisting}[
    basicstyle=\ttfamily\scriptsize,
    breaklines=true,
    numbers=none,
    frame=none,
    aboveskip=0pt,
    belowskip=0pt
]
Curious about video games that have been nominated for The Game Awards - Most Anticipated Game, excluding any that have a GRAC rating of 15+

Please find video games that meet these conditions, and provide the following attributes for these video games: video game, review score, input device, platform, distributed by, publication date, distribution format, country of origin, developer, publisher, language of work or name, genre, announcement date, game mode

**Column Specifications:**
1. **video game**
   - Precise Explanation: The official title of an electronic game that features a user interface and provides visual feedback to players during gameplay
   - Format Specification: Properly capitalized official game title matching standard marketing punctuation and spelling
   - Example: "Lunar Frontier: Operation Genesis"

(60 lines skipped)

14. **game mode**
    - Precise Explanation: The official playing modes available for the video game, including single-player and multiplayer options
    - Format Specification: Standard industry terms for video game playing modes
    - Example: "single-player video game; competitive multiplayer"

Try to provide all information. Only use '/' when you believe the information is truly unavailable or cannot be obtained.

Please organize the results in one Markdown table with the following columns: video game, review score, input device, platform, distributed by, publication date, distribution format, country of origin, developer, publisher, language of work or name, genre, announcement date, game mode.

Don't ask me any questions, just output the results according to the columns without omitting cells arbitrarily. The output format is
```markdown
{data_content}
```.
\end{lstlisting}
\vspace{0.2cm}
\rule{\textwidth}{0.4pt}
\vspace{0.2cm}

\textbf{SPARQL Query:}

\begin{lstlisting}[
    basicstyle=\ttfamily\scriptsize,
    breaklines=true,
    numbers=none,
    frame=none,
    aboveskip=0pt,
    belowskip=0pt
]
PREFIX wd: <http://www.wikidata.org/entity/>
PREFIX wdt: <http://www.wikidata.org/prop/direct/>

SELECT DISTINCT ?item WHERE {
  ?item wdt:P31/(wdt:P279|wdt:P1647)* wd:Q7889 .
  ?item wdt:P1411/(wdt:P279|wdt:P1647)* wd:Q68094302 .
  FILTER NOT EXISTS { ?item wdt:P9866/(wdt:P279|wdt:P1647)* wd:Q23005410 . }
}
\end{lstlisting}
\vspace{0.2cm}
\rule{\textwidth}{0.4pt}
\vspace{0.2cm}

\textbf{Table:}

\vspace{0.1cm}
\scriptsize
\setlength{\arrayrulewidth}{0.3pt}
\setlength{\tabcolsep}{3pt}
\begin{tabularx}{\textwidth}{|X|X|X|X|}
\hline
video game & distributed by & publisher & game mode \\
\hline
Resident Evil 4 & PlayStation Store; Microsoft Store; Steam; ... & Capcom & single-player video game \\[0.3pt]
\hline
Final Fantasy VII Rebirth & PlayStation Store; Steam & Square Enix & single-player video game \\[0.3pt]
\hline
Hades II & Nintendo eShop; Steam; Epic Games Store & Supergiant Games & single-player video game \\[0.3pt]
\hline
Death Stranding 2: On the Beach & PlayStation Store & Sony Interactive Entertainment & single-player video game \\[0.3pt]
\hline
Like a Dragon: Infinite Wealth & PlayStation Store; Microsoft Store; Steam & Sega & single-player video game \\[0.3pt]
\hline
Star Wars Outlaws & PlayStation Store; Microsoft Store; Steam; ... & Ubisoft & single-player video game \\[0.3pt]
\hline
Monster Hunter Wilds & PlayStation Store; Microsoft Store; Steam & Capcom & single-player video game \\[0.3pt]
\hline
Ghost of Yōtei & PlayStation Store & Sony Interactive Entertainment & single-player video game \\[0.3pt]
\hline
Quantum Break & Microsoft Store; Steam; Humble Store & Xbox Game Studios & single-player video game \\[0.3pt]
\hline
No Man's Sky & PlayStation Store; Microsoft Store; GOG.com; ... & Sony Interactive Entertainment; Hello Games & co-op mode; single-player video game; multiplayer video game \\[0.3pt]
\hline
\multicolumn{4}{|c|}{\textit{(showing 4 of 14 columns), (showing first 10 rows, total: 34 rows)}} \\
\hline
\end{tabularx}
\normalsize
\vspace{0.2cm}

\rule{\textwidth}{0.4pt}
\vspace{0.2cm}

\textbf{Rubrics:}

\vspace{0.1cm}
\small
\setlength{\arrayrulewidth}{0.3pt}
\setlength{\tabcolsep}{3pt}
\begin{tabularx}{\textwidth}{|p{2.5cm}|p{2cm}|X|}
\hline
\textbf{Column} & \textbf{Metric} & \textbf{Criterion} \\
\hline
video game & llm\_judge & Accept exact official game titles and widely recognized common aliases (e.g., 'RE4' matches 'Resident Evil 4'). Reject entirely unrelated game titles that do not reference the same core game. \\
\hline
review score & llm\_judge & Accept lists of scores that contain the same core numerical values regardless of formatting (percentages vs integer scores, semicolon vs comma separators). Allow minor rounding differences and ignore trivial punctuation variations. Reject scores that are significantly different in magnitude from the reference. \\
\hline
country of origin & llm\_judge & Accept exact country names and standard common aliases (e.g., 'US' matches 'United States', 'UK' matches 'United Kingdom'). Reject countries that are not the official origin of the game's development. \\
\hline
announcement date & llm\_judge & Accept dates that represent the exact same day, month, and year regardless of string format (e.g., '2022-06-02' matches 'June 2, 2022'). Ignore list order and separator punctuation. Reject dates that refer to a different official announcement day. \\
\hline
game mode & llm\_judge & Accept game mode names and common shorthand equivalents (e.g., 'co-op' matches 'co-op mode', 'multiplayer' matches 'multiplayer video game'). Ignore list order and separator punctuation. Reject game modes that are not officially supported by the game. \\
\hline
\multicolumn{3}{|c|}{\textit{(showing 5 of 14 columns)}} \\
\hline
\end{tabularx}
\normalsize
\end{tcolorbox}

\vspace{0.5cm}

\begin{tcolorbox}[
    colback=blue!5!white,
    colframe=blue!75!black,
    title={Example 6: \scalebox{1.3}{\textbf{\texttt{OR}}}},
    breakable,
    fonttitle=\bfseries,
    boxsep=3pt,
    left=5pt, right=5pt, top=5pt, bottom=5pt
]

\textbf{Domain:} Space \\
\textbf{Sub-domain:} airport \\
\vspace{0.2cm}
\rule{\textwidth}{0.4pt}
\vspace{0.2cm}

\textbf{Query:} 
\begin{lstlisting}[
    basicstyle=\ttfamily\scriptsize,
    breaklines=true,
    numbers=none,
    frame=none,
    aboveskip=0pt,
    belowskip=0pt
]
Please find all airports that have Paris-Charles de Gaulle Airport as a scheduled service destination, or are located on Crete, or are owned by Greece, or are located in Heraklion.
Please find airport that meet these conditions, and provide the following attributes for these airport: airport, IATA airport code, elevation above sea level, coordinate location, place served by transport hub, ICAO airport code, aerodrome reference point, date of official opening, named after, patronage, runway, country, located in time zone

**Column Specifications:**
1.  **airport**
    * Precise Explanation: A location where aircraft take off and land with extended support facilities, mostly for commercial air transport
    * Format Specification: Full official airport name including any honorific or nickname enclosed in quotes
    * Example: "Thessaloniki Airport \"Makedonia\""

(55 lines skipped)

13. **located in time zone**
    * Precise Explanation: Official time zone the airport operates within per the tz database
    * Format Specification: Tz database time zone identifier string
    * Example: "Europe/Athens"

Try to provide all information. Only use '/' when you believe the information is truly unavailable or cannot be obtained.

Please organize the results in one Markdown table with the following columns: airport, IATA airport code, elevation above sea level, coordinate location, place served by transport hub, ICAO airport code, aerodrome reference point, date of official opening, named after, patronage, runway, country, located in time zone.

Don't ask me any questions, just output the results according to the columns without omitting cells arbitrarily. The output format is
```markdown
{data_content}
```.
\end{lstlisting}
\vspace{0.2cm}
\rule{\textwidth}{0.4pt}
\vspace{0.2cm}

\textbf{SPARQL Query:}

\begin{lstlisting}[
    basicstyle=\ttfamily\scriptsize,
    breaklines=true,
    numbers=none,
    frame=none,
    aboveskip=0pt,
    belowskip=0pt
]
PREFIX wd: <http://www.wikidata.org/entity/>
PREFIX wdt: <http://www.wikidata.org/prop/direct/>

SELECT DISTINCT ?item WHERE {
  ?item wdt:P31/(wdt:P279|wdt:P1647)* wd:Q1248784 .
  {   { ?item wdt:P521/(wdt:P279|wdt:P1647)* wd:Q46280 . }
    UNION
  { ?item wdt:P706/(wdt:P279|wdt:P1647)* wd:Q34374 . }
    UNION
  { ?item wdt:P127/(wdt:P279|wdt:P1647)* wd:Q41 . }
    UNION
  { ?item wdt:P131/(wdt:P279|wdt:P1647)* wd:Q160544 . }
  }
}
\end{lstlisting}
\vspace{0.2cm}
\rule{\textwidth}{0.4pt}
\vspace{0.2cm}

\textbf{Table:}

\vspace{0.1cm}
\scriptsize
\setlength{\arrayrulewidth}{0.3pt}
\setlength{\tabcolsep}{3pt}
\begin{tabularx}{\textwidth}{|X|X|X|X|}
\hline
airport & place served by transport hub & named after & located in time zone \\
\hline
Heraklion International Airport "Nikos Kazantzakis" & Heraklion & Heraklion; Nikos Kazantzakis & Europe/Athens \\[0.3pt]
\hline
Chania International Airport & Chania & Daskalogiannis; Chania & Europe/Athens \\[0.3pt]
\hline
Thessaloniki Airport "Makedonia" & Thessaloniki & Macedonia; Thessaloniki; Macedonia & Europe/Athens \\[0.3pt]
\hline
Ottawa Macdonald-Cartier International Airport & Ottawa & John A. Macdonald; Ottawa; George-Étienne Cartier & Eastern Time Zone \\[0.3pt]
\hline
Roland Garros Airport & Saint-Denis & Réunion; Roland Garros; Saint-Denis & Indian/Reunion \\[0.3pt]
\hline
Athens International Airport "Eleftherios Venizelos" & Athens & Athens; Eleftherios Venizelos & Europe/Athens \\[0.3pt]
\hline
Toronto Pearson International Airport & Toronto & Lester B. Pearson; Malton; Toronto & Eastern Time Zone \\[0.3pt]
\hline
Vancouver International Airport & Vancouver & Vancouver & America/Vancouver \\[0.3pt]
\hline
Guiyang Longdongbao International Airport & Guiyang & / & Asia/Shanghai \\[0.3pt]
\hline
Warsaw Chopin Airport & Warsaw & Frederic Chopin; Warsaw; Okecie & Europe/Warsaw \\[0.3pt]
\hline
\multicolumn{4}{|c|}{\textit{(showing 4 of 13 columns), (showing first 10 rows, total: 21 rows)}} \\
\hline
\end{tabularx}
\normalsize
\vspace{0.2cm}

\rule{\textwidth}{0.4pt}
\vspace{0.2cm}

\textbf{Rubrics:}

\vspace{0.1cm}
\small
\setlength{\arrayrulewidth}{0.3pt}
\setlength{\tabcolsep}{3pt}
\begin{tabularx}{\textwidth}{|p{2.5cm}|p{2cm}|X|}
\hline
\textbf{Column} & \textbf{Metric} & \textbf{Criterion} \\
\hline
airport & llm\_judge & Accept the exact airport name, shortened variations (e.g., omitting "International"), title case or case-insensitive matches, and common aliases that clearly refer to the same airport. Reject names of entirely different airports. \\
\hline
IATA airport code & llm\_judge & Accept the standard 3-letter IATA code regardless of case (e.g., "HER" matches "her" or "Her"). Reject any non-matching 3-letter code that refers to a different airport. \\
\hline
aerodrome reference point & llm\_judge & Accept reference point coordinates that match the airport's official reference location, allowing minor decimal deviations (within \textasciitilde{}100m), and any valid coordinate format regardless of string structure. Reject points that do not correspond to the airport's reference point. \\
\hline
country & llm\_judge & Accept the exact country name or common aliases (e.g., "China" matches "People's Republic of China" or "PRC"), case variations, and abbreviations that are widely recognized. Reject names of entirely different countries. \\
\hline
located in time zone & llm\_judge & Accept standard time zone names, common abbreviations (e.g., "EST" matches "Eastern Time Zone"), and semantic equivalents that refer to the same time zone. Reject any time zone that does not correspond to the airport's geographic location. \\
\hline
\multicolumn{3}{|c|}{\textit{(showing 5 of 13 columns)}} \\
\hline
\end{tabularx}
\normalsize
\end{tcolorbox}

\vspace{0.5cm}

\begin{tcolorbox}[
    colback=blue!5!white,
    colframe=blue!75!black,
    title={Example 7: \scalebox{1.3}{\textbf{\texttt{OR\_NOT}}}},
    breakable,
    fonttitle=\bfseries,
    boxsep=3pt,
    left=5pt, right=5pt, top=5pt, bottom=5pt
]

\textbf{Domain:} Sports \\
\textbf{Sub-domain:} recurring sporting event edition \\
\vspace{0.2cm}
\rule{\textwidth}{0.4pt}
\vspace{0.2cm}

\textbf{Query:} 
\begin{lstlisting}[
    basicstyle=\ttfamily\scriptsize,
    breaklines=true,
    numbers=none,
    frame=none,
    aboveskip=0pt,
    belowskip=0pt
]
Curious about all recurring sporting event editions that either have Germany at the 2016 Summer Olympics as a participant, are part of the Summer Olympic Games sports season, or include Olympic sporting events as part of their programming, excluding any editions located in Brazil or hosted at Maracana Stadium.

Please find recurring sporting event edition that meet these conditions, and provide the following attributes for these recurring sporting event edition: recurring sporting event edition, number of participants, location, start time, end time, officially opened by

**Column Specifications:**
1.  recurring sporting event edition
    Precise Explanation: A specific yearly or regularly scheduled run of a recurring sporting event, representing one distinct iteration of the competition
    Format Specification: Formal proper noun title including the event name and associated year
    Example: 2032 Winter Paralympic Games

(20 lines skipped)

6.  officially opened by
    Precise Explanation: The public figure or official who performed the formal opening ceremony for the event
    Format Specification: Full formal name of the individual who conducted the opening ceremony
    Example: Charles III of the United Kingdom

Try to provide all information. Only use '/' when you believe the information is truly unavailable or cannot be obtained.

Please organize the results in one Markdown table with the following columns: recurring sporting event edition, number of participants, location, start time, end time, officially opened by.

Don't ask me any questions, just output the results according to the columns without omitting cells arbitrarily. The output format is
```markdown
{data_content}
```.
\end{lstlisting}
\vspace{0.2cm}
\rule{\textwidth}{0.4pt}
\vspace{0.2cm}

\textbf{SPARQL Query:}

\begin{lstlisting}[
    basicstyle=\ttfamily\scriptsize,
    breaklines=true,
    numbers=none,
    frame=none,
    aboveskip=0pt,
    belowskip=0pt
]
PREFIX wd: <http://www.wikidata.org/entity/>
PREFIX wdt: <http://www.wikidata.org/prop/direct/>

SELECT DISTINCT ?item WHERE {
  ?item wdt:P31/(wdt:P279|wdt:P1647)* wd:Q114609228 .
  {   { ?item wdt:P710/(wdt:P279|wdt:P1647)* wd:Q17739057 . }
    UNION
  { ?item wdt:P3450/(wdt:P279|wdt:P1647)* wd:Q159821 . }
    UNION
  { ?item wdt:P2670/(wdt:P279|wdt:P1647)* wd:Q18536594 . }
  }
  FILTER NOT EXISTS { { ?item wdt:P17/(wdt:P279|wdt:P1647)* wd:Q155 . } UNION { ?item wdt:P115/(wdt:P279|wdt:P1647)* wd:Q155174 . } }
}
\end{lstlisting}
\vspace{0.2cm}
\rule{\textwidth}{0.4pt}
\vspace{0.2cm}

\textbf{Table:}

\vspace{0.1cm}
\scriptsize
\setlength{\arrayrulewidth}{0.3pt}
\setlength{\tabcolsep}{3pt}
\begin{tabularx}{\textwidth}{|X|X|X|X|X|X|}
\hline
recurring sporting event edition & number of participants & location & start time & end time & officially opened by \\
\hline
2028 Summer Olympics & / & Los Angeles & 2028-07-14T00:00:00Z & 2028-07-30T00:00:00Z & / \\[0.3pt]
\hline
2020 Summer Olympics & 206.0; 11420.0 & Tokyo & 2021-07-24T00:00:00Z & 2021-08-08T00:00:00Z & Naruhito \\[0.3pt]
\hline
1896 Summer Olympics & 14.0; 241.0 & Athens & 1896-04-06T00:00:00Z & 1896-04-15T00:00:00Z & George I of Greece \\[0.3pt]
\hline
1900 Summer Olympics & 26.0; 1226.0 & Paris & 1900-05-14T00:00:00Z & 1900-10-28T00:00:00Z & / \\[0.3pt]
\hline
1904 Summer Olympics & 12.0; 651.0 & St. Louis & 1904-07-01T00:00:00Z & 1904-11-23T00:00:00Z & David R. Francis \\[0.3pt]
\hline
1908 Summer Olympics & 22.0; 2008.0 & London & 1908-04-27T00:00:00Z & 1908-10-31T00:00:00Z & Edward VII \\[0.3pt]
\hline
1912 Summer Olympics & 28.0; 2407.0 & Stockholm & 1912-05-05T00:00:00Z & 1912-07-27T00:00:00Z & Gustaf V of Sweden \\[0.3pt]
\hline
1916 Summer Olympics & / & Berlin & / & / & / \\[0.3pt]
\hline
1920 Summer Olympics & 29.0; 2622.0 & Antwerp & 1920-04-20T00:00:00Z & 1920-09-12T00:00:00Z & Albert I of Belgium \\[0.3pt]
\hline
1924 Summer Olympics & 44.0; 3088.0 & Paris & 1924-05-04T00:00:00Z & 1924-07-27T00:00:00Z & Gaston Doumergue \\[0.3pt]
\hline
\multicolumn{6}{|c|}{\textit{(showing first 10 rows, total: 27 rows)}} \\
\hline
\end{tabularx}
\normalsize
\vspace{0.2cm}

\rule{\textwidth}{0.4pt}
\vspace{0.2cm}

\textbf{Rubrics:}

\vspace{0.1cm}
\small
\setlength{\arrayrulewidth}{0.3pt}
\setlength{\tabcolsep}{3pt}
\begin{tabularx}{\textwidth}{|p{2.5cm}|p{2cm}|X|}
\hline
\textbf{Column} & \textbf{Metric} & \textbf{Criterion} \\
\hline
recurring sporting event edition & llm\_judge & Accept any reference to the exact same edition of the recurring sporting event, including common aliases and formatting variations (e.g., "1896 Summer Olympics" matches "Athens 1896 Olympics" or "1st Summer Olympic Games"). Reject references to a different year, a different recurring sporting event, or an incorrect edition entirely. \\
\hline
number of participants & llm\_judge & Accept participant counts that match the reference values within a 5\% tolerance for each numerical entry, regardless of separator used (semicolon, comma, space, etc.). Allow omission of trailing decimal zeros (e.g., 206 matches 206.0) and accept the two counts presented in either order as long as both values are correct within tolerance. Reject counts that differ by more than 10\% or are entirely unrelated to the event's participant numbers. \\
\hline
start time & llm\_judge & Accept dates that represent the exact same calendar day, month, and year for the event's start, regardless of string format or time zone notation (e.g., "2028-07-14T00:00:00Z" matches "July 14, 2028" or "14/07/2028"). Reject any date that does not correspond to the correct start date of the event edition. \\
\hline
end time & llm\_judge & Accept dates that represent the exact same calendar day, month, and year for the event's end, regardless of string format or time zone notation (e.g., "2028-07-30T00:00:00Z" matches "July 30, 2028" or "30/07/2028"). Reject any date that does not correspond to the correct end date of the event edition. \\
\hline
officially opened by & llm\_judge & Accept the exact name of the dignitary, standard title variations, common aliases, and omissions of non-essential title details (e.g., "George I of Greece" matches "King George I" or "George I"). Ignore minor variations in name formatting like spacing or punctuation. Reject any person who did not officially open the specified event edition. \\
\hline
\multicolumn{3}{|c|}{\textit{(showing 5 of 6 columns)}} \\
\hline
\end{tabularx}
\normalsize
\end{tcolorbox}

\subsection{Simulated Environment}
\label{app:envir}
To facilitate training and validation, we construct a stable and realistic simulated search engine, utilizing a snapshot of WikiPedia 2025 as the corpus. To guarantee task solvability, we verified that all entities appearing in the answer tables possess corresponding Wikipedia pages and are contained within the utilized dump. We employ Qwen3-0.6B-Embedding\footnote{https://huggingface.co/Qwen/Qwen3-Embedding-0.6B} to extract features from all text data, converting them into corresponding embeddings.This environment exposes two functions:
\begin{itemize}
    \item \texttt{search}: Computes the query embedding on the fly, retrieves the top-k nearest documents from the corpus, and returns their URLs and abstracts.
    \item \texttt{open\_page}: Retrieves the full content of a specific page given its DocID or URL.
\end{itemize}

We show the schema of these tools as below:
\begin{tcolorbox}[
    colback=blue!5!white,
    colframe=blue!75!black,
    title={\textbf{Schema of Tools}},
    breakable,
    fonttitle=\bfseries,
    boxsep=3pt,
    left=5pt, right=5pt, top=5pt, bottom=5pt
]

\begin{lstlisting}[
    basicstyle=\ttfamily\scriptsize,
    breaklines=true,
    numbers=none,
    frame=none,
    aboveskip=0pt,
    belowskip=0pt,
]

SEARCH = {
    'type': 'function',
    'function': {
        "name": "search",
        "description": "Performs a web search: supply a string 'query' and optional 'topk'. The tool retrieves the top 'topk' results (default 10) for the query, returning their docid, url, and document content (may be truncated based on token limits).",
        "parameters": {
            "type": "object",
            "properties": {
                "query": {
                    "type": "string",
                    "description": "The query string for the search."
                },
                "topk": {
                    "type": "integer",
                    "description": "Return the top k pages.",
                }
            },
            "required": [
                "query"
            ]
        }
    }
}


OPEN_PAGE = {
    'type': 'function',
    'function': {
        'name': 'open_page',
        'description': (
            "Open a page by docid or URL and return the complete content. "
            "Provide either 'docid' or 'url'; if both are provided, prefer 'docid'. "
            "The docid or URL must come from prior search tool results."
        ),
        'parameters': {
            'type': 'object',
            'properties': {
                'docid': {
                    'type': 'string',
                    'description': 'Document ID from search results to resolve and fetch.',
                },
                'url': {
                    'type': 'string',
                    'description': 'Absolute URL from search results to fetch.',
                },
            },
            'required': [],
        },
    },
}
\end{lstlisting}

\end{tcolorbox}

\subsection{Evaluation}
\label{app:eval}
To comprehensively assess the quality of the generated tables across different granularities, we employ three evaluation metrics: Success Rate, Row F1, and Item F1. These metrics evaluate the performance at the table, row, and cell levels, respectively.
Specifically, we use the LLM-based judge with column-wise rubrics to evaluate whether each generated cell is aligned with the corresponding ground truth cell. We use the GPT-4.1 as the default judge LLM.
\begin{itemize}
    \item \textbf{Success Rate:} This is the strictest metric, operating at the table level. A sample is considered a success only if the answer table exactly matches the ground truth in terms of both content and structure, without any errors.

    \item \textbf{Row F1:} This metric evaluates the retrieval and generation accuracy at the row level. We calculate the precision and recall of the generated rows against the ground truth rows to compute the F1 score. A predicted row is considered a correct match only if all the cells within that row are perfectly consistent with the corresponding ground truth row. 

    \item \textbf{Item F1:} To provide a fine-grained assessment, Item F1 evaluates performance at the cell level. It calculates the F1 score based on the individual data items (cells) within the table. This metric focuses on the model's ability to extract or generate specific details correctly, regardless of whether the entire row is perfect.
\end{itemize}
\section{Experiments}
\label{app:supplementary_materials}

\subsection{Cold Start}
\label{app:sft}
To bootstrap the unified policy $\pi_\theta$ with the capability to perform complex task decomposition and robust information seeking, we employ a Cold Start phase via Supervised Fine-Tuning (SFT).

\textbf{Trajectory Collection and Filtering.}
We utilize multiple teacher policies (e.g., DeepSeek-V3.2, Kimi-K2) to generate a diverse set of rollout trajectories on the training set $\mathcal{D}_{train}$. For each query $\mathcal{Q}_i$, we collect a set of candidate trajectories $\{\boldsymbol{\mathcal{T}}_{i,m}\}_{m=1}^M$. To ensure the quality of the training signal, we introduce a strict filtering mechanism based on the Item-level F1 score ($F1_{\text{item}}$) against the ground truth table $\mathbf{T}_i^*$. A trajectory is retained for the SFT dataset $\mathcal{D}_{SFT}$ if and only if its performance exceeds a threshold $\eta$:
{
\small 
\begin{equation}
    \mathcal{D}_{SFT} = \left\{ \boldsymbol{\mathcal{T}}_{i,m} \mid \text{Item-F1}(\text{Answer}(\boldsymbol{\mathcal{T}}_{i,m}), \mathbf{T}_i^*) > \eta \right\}
\end{equation}
}

We set the $\eta$ as 0.6.

\textbf{SFT Optimization.}
The policy $\pi_\theta$ is initialized by minimizing the standard negative log-likelihood loss over the filtered high-quality trajectories. Let $\boldsymbol{\mathcal{T}}$ be represented as a sequence of tokens $(x_1, x_2, \dots, x_L)$. The SFT objective is defined as:
{
\small
\begin{equation}
    \mathcal{L}_{SFT}(\theta) = - \mathbb{E}_{\boldsymbol{\mathcal{T}} \sim \mathcal{D}_{SFT}} \left[ \sum_{t=1}^{|\boldsymbol{\mathcal{T}}|} \log \pi_\theta(x_t \mid x_{<t}) \right]
\end{equation}
}
The loss is only computed on the tokens generated by models itself (the thoughts and actions).

\subsection{Training Dynamics}
\label{app:training_dynamics}

\begin{figure}[h]
    \centering
    \includegraphics[width=\linewidth]{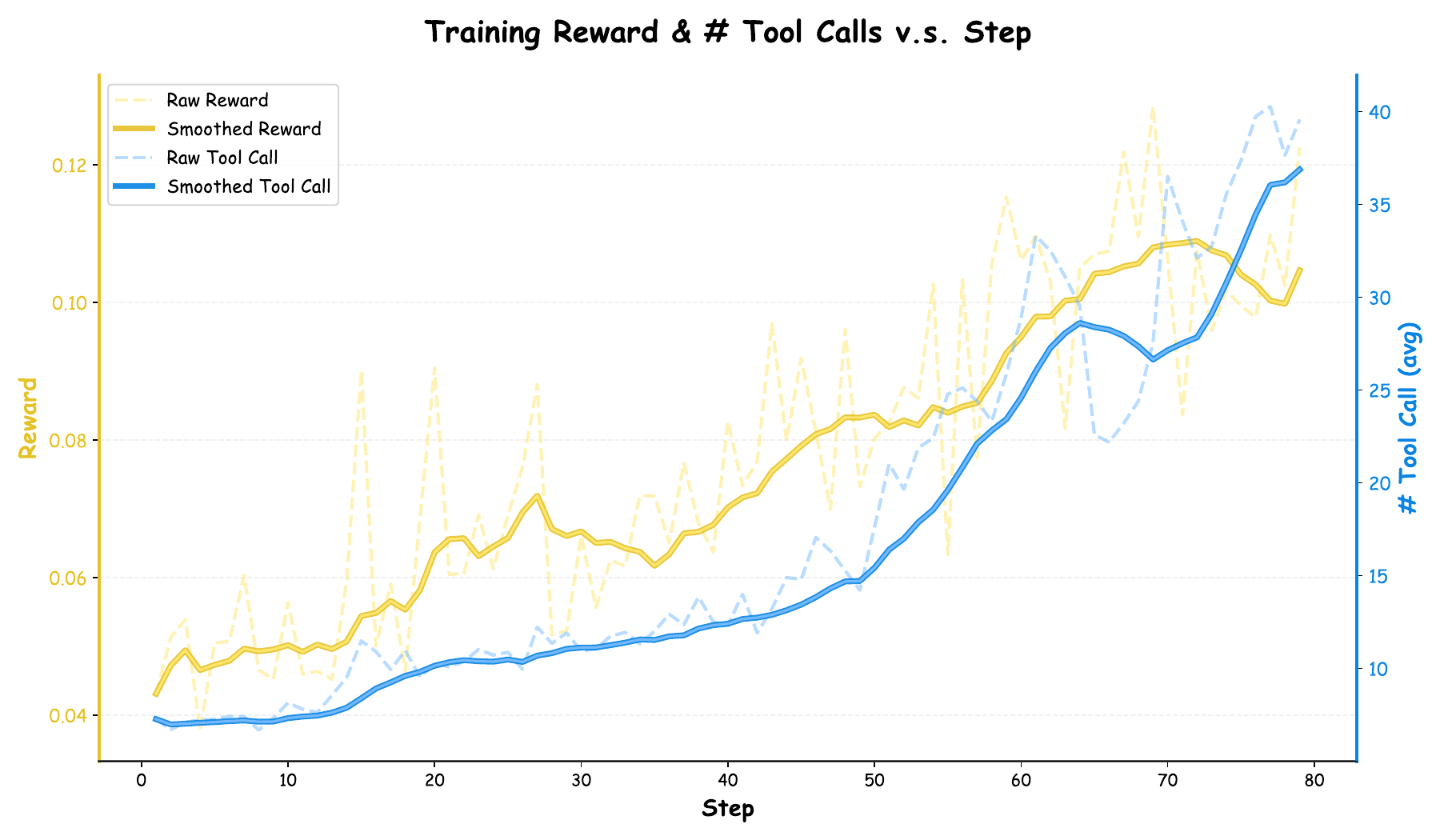}
    \caption{The trianing dynamics of WideSeek-8B-RL. We present the evolution of training rewards and the times of tool calls throughout the entire training process.}
    \label{fig:reward_tool_call}
\end{figure}

\subsection{Setting}
\label{app:setting}
We use VERL \cite{Sheng_2025} and AgentLightning \cite{luo2025agentlightningtrainai} as the RL training framework. We use Qwen3-8B \cite{yang2025qwen3technicalreport} as the base model. The RL hyper parameters are shown in Table \ref{tab:hyper_param}. We use 64 H100 GPUs for RL Training. We use function calling to create sub-agent for main agent. The corresponding tool schema is as below. We use the GPT-4.1 as the default judge LLM.
\begin{table}[h]
    \centering
    \begin{tabular}{cc}
    \toprule
     Hyper Parameter   &  Value \\
     \midrule
     Batch Size    &  64\\
     Num of Rollout & 6 \\
     Max Prompt Length & 32768 \\
     Max Response Length & 8192 \\
     Learning Rate & 1e-6 \\
     Clip High & 0.28 \\ 
     Clip Low & 0.2 \\
     Training Step & 80 \\
     \bottomrule
    \end{tabular}
    \caption{The hyper parameters for RL traning.}
    \label{tab:hyper_param}
\end{table}

\begin{tcolorbox}[
    colback=blue!5!white,
    colframe=blue!75!black,
    title={\textbf{Schema of Tools}},
    breakable,
    fonttitle=\bfseries,
    boxsep=3pt,
    left=5pt, right=5pt, top=5pt, bottom=5pt
]

\begin{lstlisting}[
    basicstyle=\ttfamily\scriptsize,
    breaklines=true,
    numbers=none,
    frame=none,
    aboveskip=0pt,
    belowskip=0pt,
]


CREATE_SUB_AGENT = {
    "type": "function",
    "function": {
        "name": "create_sub_agent",
        "description": "Create more sub-agents to gather information in parallel.",
        "parameters": {
            "type": "object",
            "properties": {
                "tasks": {
                    "type": "array",
                    "description": "A list of tasks, each for a new sub-agent. The number of subtasks created at one time can be large or small.",
                    "items": {
                        "type": "object",
                        "properties": {
                            "agent_id": {"type": "string", "description": "A unique identifier for the sub-agent, e.g., 'agent_001'."},
                            "task": {"type": "string", "description": "The specific task this sub-agent must perform."}
                        },
                        "required": ["agent_id", "task"]
                    }
                }
            },
            "required": ["tasks"]
        }
    }
}

\end{lstlisting}
\end{tcolorbox}

\subsection{Case Study}
\label{app:case_study}
\begin{figure}[htbp]
    \centering
    \includegraphics[width=0.9\textwidth]{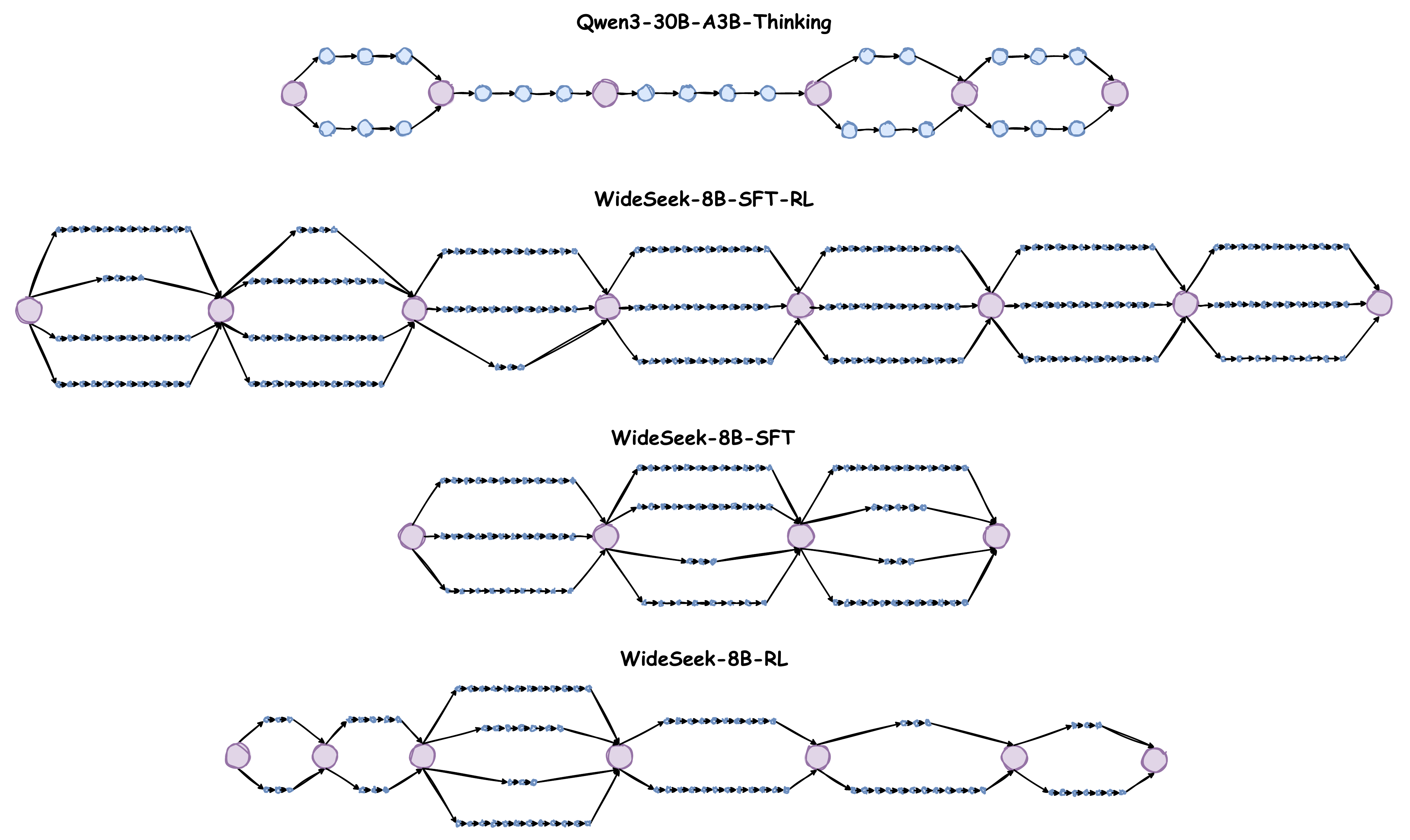}
    \caption{Multi-Agent Trajectory}
    \label{fig:case_study}
\end{figure}

We illustrate the unified trajectory of the same task query produced by 4 models in Figure \ref{fig:case_study}: Qwen3-30B-A3B-Thinking, WideSeek-8B-SFT-RL, WideSeek-8B-SFT, WideSeek-8B-RL. And for better understanding, we show a case trajectory of WideSeek-8B-RL as follows.

\newcommand{\think}[1]{\textcolor{blue}{\texttt{\textbf{<think>}}} #1 \textcolor{blue}{\texttt{\textbf{</think>}}}}
\newcommand{\search}[1]{\textcolor{cyan}{\texttt{\textbf{<tool\_call>}}} #1 \textcolor{cyan}{\texttt{\textbf{</tool\_call>}}}}
\newcommand{\response}[1]{%
  {\color{orange}\ttfamily\bfseries <tool\_response>}%
  #1%
  {\color{orange}\ttfamily\bfseries </tool\_response>}%
}
\newcommand{\answer}[1]{\textcolor{green}{\texttt{\textbf{<answer>}}} #1 \textcolor{green}{\texttt{\textbf{</answer>}}}}
\newcommand{\summary}[1]{\textcolor{red}{\texttt{\textbf{<summary>}}} #1 \textcolor{red}{\texttt{\textbf{</summary>}}}}
\newcommand{\agentresponse}[1]{\textcolor{purple}{\texttt{\textbf{<agent\_response>}}} #1 \textcolor{purple}{\texttt{\textbf{</agent\_response>}}}}

\begin{tcolorbox}[breakable, title={\textbf{Case trajectory on WideSeekBench}}]
\textcolor{black}{\textbf{Question:} Please find all films that were nominated for the Academy Award for Best Sound and received the National Board of Review: Top Ten Films award, excluding any films that feature Constantin Alexandrov as a cast member or have Anna Hamilton Phelan as a screenwriter. \\Please find film that meet these conditions, and provide the following attributes for these film: film, color, country of origin, producer, review score, CNC film rating (France), assessment, narrative location, distributed by, original language of film or TV show, composer, distribution format, box office, director, Kijkwijzer rating, production designer, duration, filming location, film editor, production company, FSK film rating \\ \\**Column Specifications:** \\ \\1.  **film** \\*   Precise Explanation: A sequence of moving images stored on film stock, released as a feature-length motion picture. \\*   Format Specification: Full official title of the film as a single string. \\*   Example: "Midnight Tide" \\ \\2.  **color** \\*   Precise Explanation: The color format used to produce the film. \\*   Format Specification: Single term indicating the color format, typically "color" or "black and white". \\*   Example: "black and white" \\ \\3.  **country of origin** \\*   Precise Explanation: The country or countries where the film was primarily produced and created. \\*   Format Specification: List of country names separated by semicolons if multiple countries apply. \\*   Example: "United Kingdom; Canada" \\ \\4.  **producer** \\*   Precise Explanation: The primary person or persons responsible for overseeing the film's core production work, excluding executive or associate producers. \\*   Format Specification: List of producer full names separated by semicolons if multiple producers are credited. \\*   Example: "Elara Voss; Marcus Hale" \\ \\5.  **review score** \\*   Precise Explanation: Aggregate or individual critical review scores the film received from rating platforms or professional critics. \\*   Format Specification: Multiple score values (percentages, numerical ratings) separated by semicolons. \\*   Example: "88\%; 9.2; 95" \\ \\6.  **CNC film rating (France)** \\*   Precise Explanation: Age restriction or content warning category assigned to the film by France's CNC Board of Film Classification. \\*   Format Specification: Rating category label, with multiple categories separated by semicolons if applicable. \\*   Example: "12 years and over; violence warning" \\ \\7.  **assessment** \\*   Precise Explanation: List of social representation or media diversity tests the film has been evaluated against. \\*   Format Specification: Names of assessment tests separated by semicolons. \\*   Example: "Bechdel test; Mako Mori test" \\ \\8.  **narrative location** \\*   Precise Explanation: The fictional or real-world settings where the film's plot and story take place. \\*   Format Specification: List of location names separated by semicolons if multiple settings are featured. \\*   Example: "Iceland; Reykjavík; Northern Fjords" \\ \\9.  **distributed by** \\*   Precise Explanation: The companies responsible for distributing the film to audiences via theatrical, streaming, or home video channels. \\*   Format Specification: List of distribution company names separated by semicolons if multiple distributors are involved. \\*   Example: "Warner Bros. Pictures; Max" \\ \\10. **original language of film or TV show** \\*   Precise Explanation: The primary language used during the original production and recording of the film's dialogue and core audio. \\*   Format Specification: Single language name or list separated by semicolons if multiple original languages are used. \\*   Example: "Spanish; Catalan" \\ \\11. **composer** \\*   Precise Explanation: The person or persons who created the original musical score for the film, separate from lyricists. \\*   Format Specification: List of composer full names separated by semicolons if multiple composers are credited. \\*   Example: "Lila Marquez; Theo Carter" \\ \\12. **distribution format** \\*   Precise Explanation: The primary method used to distribute the film to viewers. \\*   Format Specification: Single term describing the distribution format, such as theatrical release or video on demand. \\*   Example: "theatrical release" \\ \\13. **box office** \\*   Precise Explanation: The total ticket sales revenue earned by the film, potentially including domestic and international totals. \\*   Format Specification: Numeric box office values separated by semicolons if multiple totals are provided. \\*   Example: "78245100.0; 195320000.0" \\ \\14. **director** \\*   Precise Explanation: The person or persons responsible for overseeing the film's creative direction and artistic vision throughout production. \\*   Format Specification: List of director full names separated by semicolons if multiple directors are credited. \\*   Example: "Jasper Reed" \\ \\15. **Kijkwijzer rating** \\*   Precise Explanation: The age-based content rating assigned to the film under the Dutch Kijkwijzer media rating system. \\*   Format Specification: Numeric rating value, with multiple ratings separated by semicolons if applicable. \\*   Example: "9; 12" \\ \\16. **production designer** \\*   Precise Explanation: The person responsible for creating the film's overall visual aesthetic, including set design, props, and visual conceptualization. \\*   Format Specification: List of production designer full names separated by semicolons if multiple designers are credited. \\*   Example: "Zara Bennett" \\ \\17. **duration** \\*   Precise Explanation: The total runtime of the film measured in minutes. \\*   Format Specification: Numeric value representing the total runtime in minutes. \\*   Example: "122.0" \\ \\18. **filming location** \\*   Precise Explanation: The actual real-world locations where the film was physically shot, distinct from its narrative setting. \\*   Format Specification: List of filming location names separated by semicolons if multiple locations were used. \\*   Example: "Vancouver; British Columbia; Stanley Park" \\ \\19. **film editor** \\*   Precise Explanation: The person or persons responsible for assembling raw footage into the final finished cut of the film. \\*   Format Specification: List of film editor full names separated by semicolons if multiple editors are credited. \\*   Example: "Noah Brooks; Sienna Torres" \\ \\20. **production company** \\*   Precise Explanation: The companies that funded and oversaw the core production process of the film. \\*   Format Specification: List of production company names separated by semicolons if multiple companies were involved. \\*   Example: "A24; Reel Pictures Studio" \\ \\21. **FSK film rating** \\*   Precise Explanation: The age restriction rating assigned to the film under the German FSK film classification system. \\*   Format Specification: Rating in the standard format "FSK [age]". \\*   Example: "FSK 12" \\ \\Try to provide all information. Only use '/' when you believe the information is truly unavailable or cannot be obtained. \\ \\Please organize the results in one Markdown table with the following columns: film, color, country of origin, producer, review score, CNC film rating (France), assessment, narrative location, distributed by, original language of film or TV show, composer, distribution format, box office, director, Kijkwijzer rating, production designer, duration, filming location, film editor, production company, FSK film rating. \\ \\Don't ask me any questions, just output the results according to the columns without omitting cells arbitrarily. The output format is \\```markdown \\\{data\_content\} \\```.}

\colorbox{blue!10}{%
    \parbox{\dimexpr\textwidth-2\fboxsep}{%
    \centering
        \textbf{Round 1 (Main Agent + Sub Agents: agent\_001, agent\_002)}
    }
}

\textbf{\textcolor{blue!50!black}{Main Agent}}\\

\think{
Okay, let's tackle this query step by step. The user wants all films that were nominated for the Academy Award for Best Sound and received the National Board of Review: Top Ten Films award, but exclude those with Constantin Alexandrov as a cast member or Anna Hamilton Phelan as a screenwriter. Then, they need a list of attributes for each film. First, I need to find films that are in both the Academy Award Best Sound nominees and the National Board of Review Top Ten Films. Then, filter out any films that have either of the two excluded people. But the user also wants a lot of specific details about each film, which might not be straightforward to get. Since I don't have access to external databases like Wikidata, I'll need to create sub-agents to search for this information. Let me start by creating sub-agents to get the list of Academy Award Best Sound nominees and the National Board of Review Top Ten Films. Then, I can cross-reference these lists to find overlapping films. Next, check each film in the overlap for the presence of Constantin Alexandrov or Anna Hamilton Phelan, and exclude those that do. Finally, collect all the required attributes for each remaining film. But how to structure this? Let me first split the tasks. First, get the Academy Award nominees for Best Sound. Then, get the National Board of Review Top Ten Films. Then, find the intersection. Then, for each of those, check for the two excluded...
}

\search{
\{"name":"create\_sub\_agent","arguments":\{"tasks":[\{"agent\_id":"agent\_001","task":"List all films nominated for the Academy Award for Best Sound."\},\{"agent\_id":"agent\_002","task":"List all films that received the National Board of Review: Top Ten Films award."\}]\}\}
}
\vspace{0.5em}
\response{
Sub-agents for call chatcmpl-tool-7f2e53c1e9bd41d6bf46824416e0be1f completed. Reports: \\\\ --- Report from agent\_001 --- \\ Here is a list of films nominated for the Academy Award for Best Sound, based on the historical data available in the provided Wikipedia content (up to the 1970s): \\\\ \#\#\# **1930s** \\ - **1929–1930 (3rd)**: *The Big House*, *The Case of Sergeant Grischa*, *The Love Parade*, *Raffles*, *Song of the Flame* \\ - **1930–1931 (4th)**: *Paramount Publix Studio Sound Department*, *Metro-Goldwyn-Mayer Studio Sound Department*, *RKO Radio Studio Sound Department*, *Samuel Goldwyn-United Artists Studio Sound Department* \\ - **1931–1932 (5th)**: *Paramount Publix Studio Sound Department*, *Metro-Goldwyn-Mayer Studio Sound Department*, *RKO Radio Studio Sound Department*, *Warner Bros.-First National Studio Sound Department* \\ - **1932–1933 (6th)**: *A Farewell to Arms*, *42nd Street*, *Gold Diggers of 1933*, *I Am a Fugitive from a Chain Gang* \\ - **1934 (7th)**: *One Night of Love*, *The Affairs of Cellini*, *Cleopatra*, *Flirtation Walk*, *The Gay Divorcee*, *Imitation of Life*, *Viva Villa!*, *The White Parade* \\ - **1935 (8th)**: *Naughty Marietta*, *\$1,000 a Minute*, *Bride of Frankenstein*, *Captain Blood*, *The Dark Angel*, *I Dream Too Much*, *Lost Horizon*, *Maytime*, *Thanks a Million* \\ - **1936 (9th)**: *San Francisco*, *Banjo on My Knee*, *The Charge of the Light Brigade*, *Dodsworth*, *General Spanky*, *Mr. Deeds Goes to Town*, *The Texas Rangers*, *That Girl from Paris*, *You Can't Take It with You* \\ - **1937 (10th)**: *The Hurricane*, *The Girl Said No*, *Hitting a New High*, *In Old Chicago*, *The Life of Emile Zola*, *Lost Horizon*, *Maytime*, *Topper*, *Wells Fargo* \\\\ \#\#\# **1940s** \\ - **1940 (13th)**: *Strike Up the...
}

\begin{tcolorbox}[colback=green!10!white, colframe=green!70!black, boxrule=1.5pt, arc=3mm, left=3mm, right=3mm, boxsep=2mm, breakable, title={\textbf{\textcolor{green!90!black}{Sub Agent: agent\_001}}}]

\textbf{Task: }
List all films nominated for the Academy Award for Best Sound.

\hrule
\vspace{0.5em}

\textbf{\textcolor{green!80!black}{Tool Call 1:}}\\
\think{
Okay, I need to list all films that were nominated for the Academy Award for Best Sound. Let me think about how to approach this. First, the user is asking...
}

\search{
\{"name":"search","arguments":\{"query":"Academy Award for Best Sound nominations","topk":10\}\}
}

\response{
...
}

\hrule
\vspace{0.5em}

\textbf{\textcolor{green!80!black}{Tool Call 2:}}\\
\think{
Okay, so the user is asking for a list of all films nominated for the Academy Award for Best Sound. Let me start by understanding the search results they provided....
}

\search{
\{"name":"open\_page","arguments":\{"docid":"enwiki/99028"\}\}
}

\response{
...
}

\hrule
\vspace{0.5em}

\textbf{\textcolor{green!80!black}{Response}}\\
\think{
Okay, so the user wants a list of all films nominated for the Academy Award for Best Sound. Let me start by understanding the information I have from the search...
}

\agentresponse{
Here is a list of films nominated for the Academy Award for Best Sound, based on the historical data available in the provided Wikipedia content (up to the 1970s): \\ \\\#\#\#...
}

\end{tcolorbox}

\begin{tcolorbox}[colback=green!10!white, colframe=green!70!black, boxrule=1.5pt, arc=3mm, left=3mm, right=3mm, boxsep=2mm, breakable, title={\textbf{\textcolor{green!90!black}{Sub Agent: agent\_002}}}]

\textbf{Task: }
List all films that received the National Board of Review: Top Ten Films award.

\hrule
\vspace{0.5em}

\textbf{\textcolor{green!80!black}{Tool Call 1:}}\\
\think{
Okay, the user wants a list of all films that have won the National Board of Review's Top Ten Films award. First, I need to figure out how to get...
}

\search{
\{"name":"search","arguments":\{"query":"National Board of Review Top Ten Films award winners","topk":20\}\}
}

\response{
...
}

\hrule
\vspace{0.5em}

\textbf{\textcolor{green!80!black}{Tool Call 2:}}\\
\think{
Okay, so the user wants a list of all films that received the National Board of Review: Top Ten Films award. Let me check the search results they provided. Looking...
}

\search{
\{"name":"open\_page","arguments":\{"docid":"enwiki/8644529"\}\}
}

\response{
...
}

\hrule
\vspace{0.5em}

\textbf{\textcolor{green!80!black}{Response}}\\
\think{
Okay, the user is asking for a list of all films that received the National Board of Review: Top Ten Films award. Let me look through the information I have....
}

\agentresponse{
Here is a list of all films that received the **National Board of Review: Top Ten Films** award, organized by year (1929–2024): \\ \\\#\#\# **1920s** \\- **1929**: *Applause*, *Broadway*, *Bulldog Drummond*,...
}

\end{tcolorbox}

\colorbox{blue!10}{%
    \parbox{\dimexpr\textwidth-2\fboxsep}{%
    \centering
        \textbf{Round 2 (Main Agent + Sub Agents: agent\_003, agent\_004)}
    }
}

\textbf{\textcolor{blue!50!black}{Main Agent}}\\

\think{
Okay, so the user wants a list of films that were nominated for the Academy Award for Best Sound and also received the National Board of Review: Top Ten Films award. But there's a catch: we need to exclude any films that have Constantin Alexandrov as a cast member or Anna Hamilton Phelan as a screenwriter. Then, for each qualifying film, we need to gather a bunch of attributes like color, country of origin, producer, etc. First, I need to find the intersection of the two lists. The first sub-agent gave me the list of Best Sound nominees from the 1930s to 1970s, and the second sub-agent provided the Top Ten Films from the National Board of Review from 1929 to 2024. So, I need to compare these two lists to find films that are present in both. But looking at the reports, the Best Sound list from the first agent has some entries that are studio sound departments, like "Paramount Publix Studio Sound Department." Those might not be actual films, so I should probably ignore those. The user mentioned "films," so maybe those entries are errors or not applicable. I should check each entry to see if they are actual films. Next, I need to cross-reference the two lists. Let me start by looking for common films in both lists. For example, in the 1930s, the Best Sound list includes "The Love Parade" and "Viva Villa!" which also appear in the National Board of Review list. But I need...
}

\search{
\{"name":"create\_sub\_agent","arguments":\{"tasks":[\{"agent\_id":"agent\_003","task":"Identify films that are common between the Academy Award for Best Sound nominees and the National Board of Review: Top Ten Films award list, from 1930s to 1970s."\},\{"agent\_id":"agent\_004","task":"Check if any of the common films listed in agent\_003's report include Constantin Alexandrov as a cast member or Anna Hamilton Phelan as a screenwriter."\}]\}\}
}
\vspace{0.5em}
\response{
Sub-agents for call chatcmpl-tool-a9815765f0ec4795b345b11a27db9fbd completed. Reports: \\\\ --- Report from agent\_003 --- \\ The films common between the Academy Award for Best Sound nominees and the National Board of Review's Top Ten Films from the 1930s to 1970s are as follows: \\\\ \#\#\# **1930s** \\ - **1930**: \\ - *The Love Parade* (National Board) – Academy Award nominee (1930). \\ - **1931**: \\ - *City Lights* (National Board) – Academy Award nominee (1931). \\ - **1932**: \\ - *I Am a Fugitive from a Chain Gang* (National Board) – Academy Award nominee (1932). \\ - **1933**: \\ - *Topaze* (National Board) – Academy Award nominee (1933). \\ - **1934**: \\ - *It Happened One Night* (National Board) – Academy Award nominee (1934). \\ - **1935**: \\ - *The Informer* (National Board) – Academy Award nominee (1935). \\ - **1936**: \\ - *Mr. Deeds Goes to Town* (National Board) – Academy Award nominee (1936). \\ - **1937**: \\ - *A Star Is Born* (National Board) – Academy Award nominee (1937). \\ - **1938**: \\ - *Snow White and the Seven Dwarfs* (National Board) – Academy Award nominee (1939). \\ - **1939**: \\ - *Stagecoach* (National Board) – Academy Award nominee (1939). \\\\ \#\#\# **1940s** \\ - **1940**: \\ - *The Grapes of Wrath* (National Board) – Academy Award nominee (1940). \\ - **1941**: \\ - *Citizen Kane* (National Board) – Academy Award nominee (1941). \\ - **1942**: \\ - *Casablanca* (National Board) – Academy Award nominee (1942). \\ - **1943**: \\ - *Casablanca* (National Board) – Academy Award nominee (1943). \\ - **1944**: \\ - *The Miracle of Morgan's Creek* (National Board) – Academy Award nominee (1944). \\ - **1945**: \\ - *The Lost Weekend* (National Board)...
}

\begin{tcolorbox}[colback=green!10!white, colframe=green!70!black, boxrule=1.5pt, arc=3mm, left=3mm, right=3mm, boxsep=2mm, breakable, title={\textbf{\textcolor{green!90!black}{Sub Agent: agent\_003}}}]

\textbf{Task: }
Identify films that are common between the Academy Award for Best Sound nominees and the National Board of Review: Top Ten Films award list, from 1930s to 1970s.

\hrule
\vspace{0.5em}

\textbf{\textcolor{green!80!black}{Tool Call 1:}}\\
\think{
Okay, so I need to find films that are common between the Academy Award for Best Sound nominees and the National Board of Review's Top Ten Films list from the...
}

\search{
\{"name":"search","arguments":\{"query":"National Board of Review Top Ten Films 1930s to 1970s","topk":10\}\}
}

\response{
...
}

\hrule
\vspace{0.5em}

\textbf{\textcolor{green!80!black}{Tool Call 2:}}\\
\think{
Okay, so the user wants to find films that are common between the Academy Award for Best Sound nominees and the National Board of Review's Top Ten Films list from...
}

\search{
\{"name":"open\_page","arguments":\{"docid":"enwiki/99028"\}\}
}

\response{
...
}

\hrule
\vspace{0.5em}

\textbf{\textcolor{green!80!black}{Response}}\\
\think{
Okay, so the user wants to find films that are common between the Academy Award for Best Sound nominees and the National Board of Review's Top Ten Films list from...
}

\agentresponse{
The films common between the Academy Award for Best Sound nominees and the National Board of Review's Top Ten Films from the 1930s to 1970s are as follows: \\ \\\#\#\# **1930s**
}

\end{tcolorbox}

\begin{tcolorbox}[colback=green!10!white, colframe=green!70!black, boxrule=1.5pt, arc=3mm, left=3mm, right=3mm, boxsep=2mm, breakable, title={\textbf{\textcolor{green!90!black}{Sub Agent: agent\_004}}}]

\textbf{Task: }
Check if any of the common films listed in agent\_003's report include Constantin Alexandrov as a cast member or Anna Hamilton Phelan as a screenwriter.

\hrule
\vspace{0.5em}

\textbf{\textcolor{green!80!black}{Tool Call 1:}}\\
\think{
Okay, I need to figure out if Constantin Alexandrov is a cast member in any of the common films listed in agent\_003's report or if Anna Hamilton Phelan is a...
}

\search{
\{"name":"search","arguments":\{"query":"agent\_003's report common films","topk":10\}\}
}

\response{
...
}

\hrule
\vspace{0.5em}

\textbf{\textcolor{green!80!black}{Tool Call 2:}}\\
\think{
Okay, so the user wants to check if any of the common films listed in agent\_003's report include Constantin Alexandrov as a cast member or Anna Hamilton Phelan as a...
}

\search{
\{"name":"search","arguments":\{"query":"Constantin Alexandrov filmography Anna Hamilton Phelan screenwriting credits","topk":5\}\}
}

\response{
...
}

\hrule
\vspace{0.5em}

\textbf{\textcolor{green!80!black}{Tool Call 3:}}\\
\think{
Okay, let's see. The user wants to check if any of the common films listed in agent\_003's report include Constantin Alexandrov as a cast member or Anna Hamilton Phelan as...
}

\search{
\{"name":"search","arguments":\{"query":"Constantin Alexandrov filmography","topk":5\}\}
}

\response{
...
}

\hrule
\vspace{0.5em}

\textbf{\textcolor{green!80!black}{Tool Call 4:}}\\
\think{
Okay, so the task is to check if any of the common films listed in agent\_003's report include Constantin Alexandrov as a cast member or Anna Hamilton Phelan as a...
}

\search{
\{"name":"open\_page","arguments":\{"docid":"enwiki/18716679"\}\}
}

\response{
...
}

\hrule
\vspace{0.5em}

\textbf{\textcolor{green!80!black}{Response}}\\
\think{
Okay, so the task is to check if any of the common films listed in agent\_003's report include Constantin Alexandrov as a cast member or Anna Hamilton Phelan as a...
}

\agentresponse{
After reviewing the search results and full content of the films listed in agent\_003's report, there is no evidence that **Constantin Alexandrov** appears as a cast member or **Anna Hamilton...
}

\end{tcolorbox}

\colorbox{blue!10}{%
    \parbox{\dimexpr\textwidth-2\fboxsep}{%
    \centering
        \textbf{Round 3 (Main Agent + Sub Agents: agent\_005, agent\_006, agent\_007, agent\_008)}
    }
}

\textbf{\textcolor{blue!50!black}{Main Agent}}\\

\think{
Okay, so the user wants all films that were nominated for the Academy Award for Best Sound and also received the National Board of Review: Top Ten Films award, but exclude any that have Constantin Alexandrov as a cast member or Anna Hamilton Phelan as a screenwriter. Then, they need a bunch of attributes for each film. First, I need to find the intersection of the two lists. From the first sub-agent, agent\_001 gave the Best Sound nominees, and agent\_002 gave the National Board Top Ten Films. Then agent\_003 found the common films between those two lists. The result was a list of films from 1930s to 1970s. But then agent\_004 checked if any of those films include Constantin Alexandrov or Anna Hamilton Phelan. The report said no, so those films are safe. But I should make sure that agent\_004's conclusion is accurate. Maybe double-check some of the films mentioned. For example, *Casablanca* is in both lists. Let me verify if Constantin Alexandrov was in *Casablanca*. Wait, the report says agent\_004 checked and found no mention of him. Similarly, Anna Hamilton Phelan isn't listed as a screenwriter for any of them. Now, the next step is to gather all the required attributes for each of these films. The columns are film, color, country of origin, producer, review score, CNC film rating (France), assessment, narrative location, distributed by, original language, composer, distribution format, box office, director, Kijkwijzer rating, production designer, duration, filming location, film editor, production company, FSK film rating. But...
}

\search{
\{"name":"create\_sub\_agent","arguments":\{"tasks":[\{"agent\_id":"agent\_005","task":"Retrieve detailed attributes for *It Happened One Night* (1934): film, color, country of origin, producer, review score, CNC film rating (France), assessment, narrative location, distributed by, original language, composer, distribution format, box office, director, Kijkwijzer rating, production designer, duration, filming location, film editor, production company, FSK film rating."\},\{"agent\_id":"agent\_006","task":"Retrieve detailed attributes for *Casablanca* (1942): film, color, country of origin, producer, review score, CNC film rating (France), assessment, narrative location, distributed by, original language, composer, distribution format, box office, director, Kijkwijzer rating, production designer, duration, filming location, film editor, production company, FSK film rating."\},\{"agent\_id":"agent\_007","task":"Retrieve detailed attributes for *The Grapes of Wrath* (1940): film, color, country of origin, producer, review score, CNC film rating (France), assessment, narrative location, distributed by, original language, composer, distribution format, box office, director, Kijkwijzer rating, production designer, duration, filming location, film editor, production company, FSK film rating."\},\{"agent\_id":"agent\_008","task":"Retrieve detailed attributes for *Citizen Kane* (1941): film, color, country of origin, producer, review score, CNC film rating (France), assessment, narrative location, distributed by, original language, composer, distribution format, box office, director, Kijkwijzer rating, production designer, duration, filming location, film editor, production company, FSK film rating."\}]\}\}
}
\vspace{0.5em}
\response{
Sub-agents for call chatcmpl-tool-b83f2882255149a29a02640892fa7924 completed. Reports: \\\\ --- Report from agent\_005 --- \\ The detailed attributes for *It Happened One Night* (1934) are as follows: \\\\ \#\#\# **Film Attributes** \\ - **Title**: *It Happened One Night* \\ - **Color**: Black-and-white (typical for 1934 films). \\ - **Country of Origin**: United States. \\ - **Producer**: Frank Capra and Harry Cohn. \\ - **Review Score**: \\ - **Rotten Tomatoes**: 98\% score, 9.1/10 average. \\ - **Metacritic**: 87/100. \\ - **CNC Film Rating (France)**: Not explicitly mentioned in available sources. France’s CNC (Commission Nationale du Cinema) ratings may require checking French-specific databases or archives. \\ - **Assessment**: Widely regarded as one of the greatest films ever made, winning all five major Academy Awards (Best Picture, Best Director, Best Actor, Best Actress, Best Adapted Screenplay). \\ - **Narrative Location**: Florida (Ellie’s escape), New York City, and cross-country bus journey. \\ - **Distributed By**: Columbia Pictures. \\ - **Original Language**: English. \\ - **Composer**: Howard Jackson and Louis Silvers. \\ - **Distribution Format**: Theatrical release (standard for 1934). \\ - **Box Office**: \$2.5 million (worldwide rentals). \\ - **Director**: Frank Capra. \\ - **Kijkwijzer Rating (Netherlands)**: Not found in available sources. Dutch ratings would need a specialized database. \\ - **Production Designer**: Not explicitly listed in the Wikipedia infobox. Research into production design credits or industry archives may reveal this. \\ - **Duration**: 105 minutes. \\ - **Filming Location**: Part of the film was shot on Thousand Oaks Boulevard in Thousand Oaks, California. \\ - **Film Editor**: Gene Havlick. \\ - **Production Company**: Columbia Pictures. \\ - **FSK Film Rating (Germany)**: Not specified in sources. Germany’s FSK ratings would require checking...
}

\begin{tcolorbox}[colback=green!10!white, colframe=green!70!black, boxrule=1.5pt, arc=3mm, left=3mm, right=3mm, boxsep=2mm, breakable, title={\textbf{\textcolor{green!90!black}{Sub Agent: agent\_005}}}]

\textbf{Task: }
Retrieve detailed attributes for *It Happened One Night* (1934): film, color, country of origin, producer, review score, CNC film rating (France), assessment, narrative location, distributed by, original language, composer, distribution format, box office, director, Kijkwijzer rating, production designer, duration, filming location, film editor, production company, FSK film rating.

\hrule
\vspace{0.5em}

\textbf{\textcolor{green!80!black}{Tool Call 1:}}\\
\think{
Okay, so the user wants detailed attributes for the 1934 film "It Happened One Night." Let me start by figuring out what information I need to retrieve. The task includes...
}

\search{
\{"name":"search","arguments":\{"query":"It Happened One Night 1934 detailed attributes","topk":10\}\}
}

\response{
...
}

\hrule
\vspace{0.5em}

\textbf{\textcolor{green!80!black}{Tool Call 2:}}\\
\think{
Okay, so the user wants detailed attributes for the 1934 film "It Happened One Night." Let me look through the search results they provided. The first result is the Wikipedia...
}

\search{
\{"name":"open\_page","arguments":\{"docid":"enwiki/44696"\}\}
}

\response{
...
}

\hrule
\vspace{0.5em}

\textbf{\textcolor{green!80!black}{Tool Call 3:}}\\
\think{
Okay, let me start by looking at the user's request. They want detailed attributes for the 1934 film "It Happened One Night." The attributes listed include film, color, country of...
}

\search{
\{"name":"search","arguments":\{"query":"It Happened One Night 1934 FSK rating","topk":1\}\}
}

\response{
...
}

\hrule
\vspace{0.5em}

\textbf{\textcolor{green!80!black}{Tool Call 4:}}\\
\think{
Okay, let me figure out how to approach this. The user wants detailed attributes for the 1934 film "It Happened One Night." So far, I've used the search tool to...
}

\search{
\{"name":"search","arguments":\{"query":"It Happened One Night 1934 production designer","topk":1\}\}
}

\response{
...
}

\hrule
\vspace{0.5em}

\textbf{\textcolor{green!80!black}{Tool Call 5:}}\\
\think{
Okay, so the user is asking for detailed attributes of the 1934 film "It Happened One Night." Let me go through what they need. The attributes listed include film, color,...
}

\search{
\{"name":"search","arguments":\{"query":"It Happened One Night 1934 production designer","topk":1\}\}
}

\response{
...
}

\hrule
\vspace{0.5em}

\textbf{\textcolor{green!80!black}{Tool Call 6:}}\\
\think{
Okay, let's see. The user is asking for detailed attributes of the 1934 film "It Happened One Night." They want info on film, color, country of origin, producer, review score,...
}

\search{
\{"name":"open\_page","arguments":\{"docid":"enwiki/44696"\}\}
}

\response{
...
}

\hrule
\vspace{0.5em}

\textbf{\textcolor{green!80!black}{Response}}\\
\think{
Okay, let me try to figure out how to get all the detailed attributes for "It Happened One Night" (1934). The user is asking for a bunch of specific info:...
}

\agentresponse{
The detailed attributes for *It Happened One Night* (1934) are as follows: \\ \\\#\#\# **Film Attributes** \\- **Title**: *It Happened One Night* \\- **Color**: Black-and-white (typical for 1934 films). \\- **Country...
}

\end{tcolorbox}

\begin{tcolorbox}[colback=green!10!white, colframe=green!70!black, boxrule=1.5pt, arc=3mm, left=3mm, right=3mm, boxsep=2mm, breakable, title={\textbf{\textcolor{green!90!black}{Sub Agent: agent\_006}}}]

\textbf{Task: }
You have reached the token limit. Please summarize your relevant findings immediately based on existing information.

\hrule
\vspace{0.5em}

\textbf{\textcolor{green!80!black}{Tool Call 1:}}\\
\think{
Okay, I need to retrieve detailed attributes for the 1942 film Casablanca. Let me start by breaking down the user's request. They want information on various aspects like film, color,...
}

\search{
\{"name":"search","arguments":\{"query":"Casablanca 1942 detailed attributes","topk":10\}\}
}

\response{
...
}

\hrule
\vspace{0.5em}

\textbf{\textcolor{green!80!black}{Tool Call 2:}}\\
\think{
Okay, so the user wants detailed attributes for the 1942 film "Casablanca." Let me look through the search results they provided. The first result is about 1942 in film, which...
}

\search{
\{"name":"open\_page","arguments":\{"docid":"enwiki/57585"\}\}
}

\response{
...
}

\hrule
\vspace{0.5em}

\textbf{\textcolor{green!80!black}{Tool Call 3:}}\\
\think{
Okay, let's see. The user wants detailed attributes for the 1942 film Casablanca. The search results gave me a few documents, but the most relevant one is the Wikipedia page...
}

\search{
\{"name":"search","arguments":\{"query":"Casablanca 1942 CNC rating France","topk":1\}\}
}

\response{
...
}

\hrule
\vspace{0.5em}

\textbf{\textcolor{green!80!black}{Tool Call 4:}}\\
\think{
Okay, let's tackle this task step by step. The user wants detailed attributes for the 1942 film Casablanca, including things like color, country of origin, producer, review score, CNC film...
}

\search{
\{"name":"search","arguments":\{"query":"Casablanca 1942 Kijkwijzer rating","topk":1\}\}
}

\response{
...
}

\hrule
\vspace{0.5em}

\textbf{\textcolor{green!80!black}{Tool Call 5:}}\\
\think{
Okay, let's tackle this task step by step. The user wants detailed attributes for the 1942 film "Casablanca." So far, I've used the search and open\_page tools to get information...
}

\search{
\{"name":"search","arguments":\{"query":"Casablanca 1942 CNC film rating France","topk":1\}\}
}

\response{
...
}

\hrule
\vspace{0.5em}

\textbf{\textcolor{green!80!black}{Tool Call 6:}}\\
\think{
Okay, let's see. The user is asking for detailed attributes of the 1942 film Casablanca. They want information on film, color, country of origin, producer, review score, CNC film rating...
}

\search{
\{"name":"search","arguments":\{"query":"Casablanca 1942 FSK film rating","topk":1\}\}
}

\response{
...
}

\hrule
\vspace{0.5em}

\textbf{\textcolor{green!80!black}{Tool Call 7:}}\\
\think{
Okay, let's see. The user wants detailed attributes for the 1942 film Casablanca. They listed a bunch of specific fields: film, color, country of origin, producer, review score, CNC film...
}

\search{
\{"name":"search","arguments":\{"query":"Casablanca 1942 CNC rating France","topk":1\}\}
}

\response{
...
}

\end{tcolorbox}

\begin{tcolorbox}[colback=green!10!white, colframe=green!70!black, boxrule=1.5pt, arc=3mm, left=3mm, right=3mm, boxsep=2mm, breakable, title={\textbf{\textcolor{green!90!black}{Sub Agent: agent\_006 (continued)}}}]

\hrule
\vspace{0.5em}

\textbf{\textcolor{green!80!black}{Tool Call 8:}}\\
\think{
Okay, let's see. The user wants detailed attributes for the 1942 film Casablanca. They listed a bunch of specific pieces of information: film, color, country of origin, producer, review score,...
}

\search{
\{"name":"search","arguments":\{"query":"Casablanca 1942 production designer","topk":1\}\}
}

\response{
...
}

\hrule
\vspace{0.5em}

\textbf{\textcolor{green!80!black}{Tool Call 9:}}\\
\think{
Okay, let's tackle this task step by step. The user is asking for detailed attributes of the 1942 film "Casablanca." They want information on film, color, country of origin, producer,...
}

\search{
\{"name":"open\_page","arguments":\{"docid":"enwiki/57585"\}\}
}

\response{
...
}

\hrule
\vspace{0.5em}

\textbf{\textcolor{green!80!black}{Tool Call 10:}}\\
\think{
Okay, let me tackle this task. The user wants detailed attributes for the 1942 film Casablanca. They listed a bunch of categories like film, color, country of origin, producer, review...
}

\search{
\{"name":"search","arguments":\{"query":"Casablanca 1942 production designer","topk":1\}\}
}

\response{
...
}

\hrule
\vspace{0.5em}

\textbf{\textcolor{green!80!black}{Tool Call 11:}}\\
\think{
Okay, let's see. The user wants detailed attributes for the 1942 film Casablanca. They listed a bunch of categories: film, color, country of origin, producer, review score, CNC film rating...
}

\search{
\{"name":"search","arguments":\{"query":"Casablanca 1942 production designer","topk":1\}\}
}

\response{
reached the token limit
}

\hrule
\vspace{0.5em}

\textbf{\textcolor{green!80!black}{Response}}\\
\think{
Okay, let's tackle this. The user wants detailed attributes for the 1942 film Casablanca. They listed a bunch of categories like film, color, country of origin, producer, etc. First, I...
}

\agentresponse{
Based on the available information from the Wikipedia page for *Casablanca* (1942), here is a comprehensive summary of its detailed attributes: \\ \\\#\#\# **Film Details** \\- **Title**: Casablanca \\- **Type**: American...
}

\end{tcolorbox}

\begin{tcolorbox}[colback=green!10!white, colframe=green!70!black, boxrule=1.5pt, arc=3mm, left=3mm, right=3mm, boxsep=2mm, breakable, title={\textbf{\textcolor{green!90!black}{Sub Agent: agent\_007}}}]

\textbf{Task: }
You have reached the token limit. Please summarize your relevant findings immediately based on existing information.

\hrule
\vspace{0.5em}

\textbf{\textcolor{green!80!black}{Tool Call 1:}}\\
\think{
Okay, I need to find detailed attributes for the 1940 film "The Grapes of Wrath." Let me start by breaking down the task. The user is asking for a lot...
}

\search{
\{"name":"search","arguments":\{"query":"The Grapes of Wrath 1940","topk":10\}\}
}

\response{
...
}

\hrule
\vspace{0.5em}

\textbf{\textcolor{green!80!black}{Tool Call 2:}}\\
\think{
Okay, let's tackle this task. The user wants detailed attributes for the 1940 film "The Grapes of Wrath." First, I need to figure out which of the search results are...
}

\search{
\{"name":"open\_page","arguments":\{"docid":"enwiki/4350811"\}\}
}

\response{
...
}

\hrule
\vspace{0.5em}

\textbf{\textcolor{green!80!black}{Tool Call 3:}}\\
\think{
Okay, let's see. The user wants detailed attributes for the 1940 film "The Grapes of Wrath." They listed a bunch of specific fields like film, color, country of origin, producer,...
}

\search{
\{"name":"search","arguments":\{"query":"The Grapes of Wrath 1940 FSK rating","topk":1\}\}
}

\response{
...
}

\hrule
\vspace{0.5em}

\textbf{\textcolor{green!80!black}{Tool Call 4:}}\\
\think{
Okay, so the user wants detailed attributes for the 1940 film "The Grapes of Wrath." Let me start by going through the information I have. First, from the initial search...
}

\search{
\{"name":"search","arguments":\{"query":"The Grapes of Wrath 1940 production designer","topk":1\}\}
}

\response{
...
}

\hrule
\vspace{0.5em}

\textbf{\textcolor{green!80!black}{Tool Call 5:}}\\
\think{
Okay, so I need to retrieve detailed attributes for the 1940 film "The Grapes of Wrath." Let me start by recalling the information I already have from the previous searches....
}

\search{
\{"name":"search","arguments":\{"query":"The Grapes of Wrath 1940 CNC rating","topk":1\}\}
}

\response{
...
}

\hrule
\vspace{0.5em}

\textbf{\textcolor{green!80!black}{Tool Call 6:}}\\
\think{
Okay, so the user is asking for detailed attributes of the 1940 film "The Grapes of Wrath." Let me see what information I have so far. From the previous search...
}

\search{
\{"name":"search","arguments":\{"query":"The Grapes of Wrath 1940 production designer","topk":1\}\}
}

\response{
...
}

\hrule
\vspace{0.5em}

\textbf{\textcolor{green!80!black}{Tool Call 7:}}\\
\think{
Okay, so the user is asking for detailed attributes of the 1940 film "The Grapes of Wrath." Let me start by going through the information I have. First, from the...
}

\search{
\{"name":"search","arguments":\{"query":"The Grapes of Wrath 1940 production designer","topk":1\}\}
}

\response{
...
}

\end{tcolorbox}

\begin{tcolorbox}[colback=green!10!white, colframe=green!70!black, boxrule=1.5pt, arc=3mm, left=3mm, right=3mm, boxsep=2mm, breakable, title={\textbf{\textcolor{green!90!black}{Sub Agent: agent\_007 (continued)}}}]

\hrule
\vspace{0.5em}

\textbf{\textcolor{green!80!black}{Tool Call 8:}}\\
\think{
Okay, let's see. The user wants detailed attributes for the 1940 film "The Grapes of Wrath." They've already done a search and opened the Wikipedia page for the film. The...
}

\search{
\{"name":"search","arguments":\{"query":"The Grapes of Wrath 1940 production designer","topk":1\}\}
}

\response{
...
}

\hrule
\vspace{0.5em}

\textbf{\textcolor{green!80!black}{Tool Call 9:}}\\
\think{
Okay, let's tackle this task step by step. The user wants detailed attributes for "The Grapes of Wrath" (1940), including film, color, country of origin, producer, review score, CNC film...
}

\search{
\{"name":"open\_page","arguments":\{"docid":"enwiki/4350811"\}\}
}

\response{
...
}

\hrule
\vspace{0.5em}

\textbf{\textcolor{green!80!black}{Tool Call 10:}}\\
\think{
Okay, let me see. The user is asking for detailed attributes of the 1940 film "The Grapes of Wrath." They listed a bunch of categories like film, color, country of...
}

\search{
\{"name":"search","arguments":\{"query":"The Grapes of Wrath 1940 production designer","topk":1\}\}
}

\response{
...
}

\hrule
\vspace{0.5em}

\textbf{\textcolor{green!80!black}{Tool Call 11:}}\\
\think{
Okay, so I need to retrieve detailed attributes for "The Grapes of Wrath" (1940) film. Let me start by recalling the user's query. They want information on film, color, country...
}

\search{
\{"name":"search","arguments":\{"query":"The Grapes of Wrath 1940 production designer","topk":1\}\}
}

\response{
reached the token limit
}

\hrule
\vspace{0.5em}

\textbf{\textcolor{green!80!black}{Response}}\\
\think{
Okay, so the user wants detailed attributes for the 1940 film "The Grapes of Wrath." Let me start by recalling the information I have from the previous searches and the...
}

\agentresponse{
\#\#\# Detailed Attributes for *The Grapes of Wrath* (1940) \\1. **Film**: \\- **Title**: *The Grapes of Wrath* (1940) \\- **Type**: Drama / Adaptation of John Steinbeck’s novel. \\- **Country of...
}

\end{tcolorbox}

\begin{tcolorbox}[colback=green!10!white, colframe=green!70!black, boxrule=1.5pt, arc=3mm, left=3mm, right=3mm, boxsep=2mm, breakable, title={\textbf{\textcolor{green!90!black}{Sub Agent: agent\_008}}}]

\textbf{Task: }
Retrieve detailed attributes for *Citizen Kane* (1941): film, color, country of origin, producer, review score, CNC film rating (France), assessment, narrative location, distributed by, original language, composer, distribution format, box office, director, Kijkwijzer rating, production designer, duration, filming location, film editor, production company, FSK film rating.

\hrule
\vspace{0.5em}

\textbf{\textcolor{green!80!black}{Tool Call 1:}}\\
\think{
Okay, I need to retrieve detailed attributes for the film *Citizen Kane* (1941). The user has listed a bunch of specific pieces of information they want, so I should start...
}

\search{
\{"name":"search","arguments":\{"query":"Citizen Kane 1941","topk":10\}\}
}

\response{
...
}

\hrule
\vspace{0.5em}

\textbf{\textcolor{green!80!black}{Tool Call 2:}}\\
\think{
Okay, let me start by understanding the task. The user wants detailed attributes for the film "Citizen Kane" (1941). They listed a bunch of specific pieces of information: color, country...
}

\search{
\{"name":"open\_page","arguments":\{"docid":"enwiki/5224"\}\}
}

\response{
...
}

\hrule
\vspace{0.5em}

\textbf{\textcolor{green!80!black}{Response}}\\
\think{
Okay, let me start by figuring out what the user needs. They want detailed attributes for the film *Citizen Kane* (1941), including things like color, country of origin, producer, review...
}

\agentresponse{
Here is the detailed information about *Citizen Kane* (1941): \\ \\1. **Film**: *Citizen Kane* \\2. **Color**: Black and white (though the innovative 1940 trailer was in color). \\3. **Country of Origin**:...
}

\end{tcolorbox}

\colorbox{blue!10}{%
    \parbox{\dimexpr\textwidth-2\fboxsep}{%
    \centering
        \textbf{Round 4 (Main Agent + Sub Agents: agent\_009, agent\_010)}
    }
}

\textbf{\textcolor{blue!50!black}{Main Agent}}\\

\think{
Okay, so the user is asking for all films that were nominated for the Academy Award for Best Sound and received the National Board of Review: Top Ten Films award, excluding any films with Constantin Alexandrov as a cast member or Anna Hamilton Phelan as a screenwriter. Then, they want all the specified attributes for those films. First, I need to find the intersection of the two lists: Academy Award Best Sound nominees and National Board of Review Top Ten Films. The previous agents did that for the 1930s to 1970s and found some common films like *It Happened One Night*, *Casablanca*, *The Grapes of Wrath*, and *Citizen Kane*. Agent\_004 checked if any of these have the excluded names and found none, so those four films are still in the running. Now, the user wants detailed attributes for each of these films. Agents 005, 006, 007, and 008 retrieved info for each, but some details are missing, like regional ratings (CNC, Kijkwijzer, FSK) and production designers. The user wants all information, using '/' for unavailable data. I need to compile the data from these reports into the required markdown table. Let me check each film's details: For *It Happened One Night* (1934): - Color: Black-and-white - Country of Origin: United States - Producer: Frank Capra and Harry Cohn - Review Score: 98\% on Rotten Tomatoes, 9.1/10 average - CNC Film Rating: Not found - Assessment: Won 5 Academy Awards - Narrative Location: Florida, New York City, cross-country bus journey -...
}

\search{
\{"name":"create\_sub\_agent","arguments":\{"tasks":[\{"agent\_id":"agent\_009","task":"Retrieve detailed attributes for *The Life of Emile Zola* (1937): film, color, country of origin, producer, review score, CNC film rating (France), assessment, narrative location, distributed by, original language, composer, distribution format, box office, director, Kijkwijzer rating, production designer, duration, filming location, film editor, production company, FSK film rating."\},\{"agent\_id":"agent\_010","task":"Retrieve detailed attributes for *The Best Years of Our Lives* (1946): film, color, country of origin, producer, review score, CNC film rating (France), assessment, narrative location, distributed by, original language, composer, distribution format, box office, director, Kijkwijzer rating, production designer, duration, filming location, film editor, production company, FSK film rating."\}]\}\}
}
\vspace{0.5em}
\response{
Sub-agents for call chatcmpl-tool-82a2e428e1c443c29b5e1a8683cbeb7b completed. Reports: \\\\ --- Report from agent\_009 --- \\ \#\#\# Detailed Attributes for *The Life of Emile Zola* (1937): \\\\ 1. **Film**: Biographical drama based on Émile Zola’s life, focusing on the Dreyfus Affair and his literary career. \\ 2. **Color**: Black-and-white (standard for 1937 films). \\ 3. **Country of Origin**: United States. \\ 4. **Producer**: Henry Blanke (produced by Warner Bros.). \\ 5. **Review Score**: \\ - **Rotten Tomatoes**: 92\% from 72 reviews (average score: 7.5/10). \\ - **Critics Consensus**: Praised for its historical significance, Paul Muni’s performance, and thematic relevance to 1930s political tensions. \\ 6. **CNC Film Rating (France)**: Not explicitly mentioned in available sources. The film’s connection to the Dreyfus Affair may have influenced its reception in France, but no official CNC rating is listed. \\ 7. **Assessment**: Regarded as a landmark biographical film, winning three Oscars (Best Picture, Best Supporting Actor, Best Screenplay). Critics highlighted its moral urgency and Paul Muni’s portrayal of Zola. \\ 8. **Narrative Location**: Set in 19th-century France, spanning Paris, French Guiana (Devil’s Island), and other locations tied to Zola’s life and the Dreyfus Affair. \\ 9. **Distributed By**: Warner Bros. Pictures. \\ 10. **Original Language**: English (though the story is about a French author and historical events). \\ 11. **Composer**: Max Steiner (score awarded to Leo F. Forbstein). \\ 12. **Distribution Format**: Theatrical release (no mention of home media formats). \\ 13. **Box Office**: Specific revenue figures are not provided in available sources. \\ 14. **Director**: William Dieterle. \\ 15. **Kijkwijzer Rating (Netherlands)**: Not mentioned in sources. \\ 16. **Production Designer**: Not explicitly listed in the...
}

\begin{tcolorbox}[colback=green!10!white, colframe=green!70!black, boxrule=1.5pt, arc=3mm, left=3mm, right=3mm, boxsep=2mm, breakable, title={\textbf{\textcolor{green!90!black}{Sub Agent: agent\_009}}}]

\textbf{Task: }
Retrieve detailed attributes for *The Life of Emile Zola* (1937): film, color, country of origin, producer, review score, CNC film rating (France), assessment, narrative location, distributed by, original language, composer, distribution format, box office, director, Kijkwijzer rating, production designer, duration, filming location, film editor, production company, FSK film rating.

\hrule
\vspace{0.5em}

\textbf{\textcolor{green!80!black}{Tool Call 1:}}\\
\think{
Okay, I need to find detailed attributes for the 1937 film "The Life of Emile Zola." Let's start by breaking down the task. The user is asking for a lot...
}

\search{
\{"name":"search","arguments":\{"query":"The Life of Emile Zola 1937 film details","topk":10\}\}
}

\response{
...
}

\hrule
\vspace{0.5em}

\textbf{\textcolor{green!80!black}{Tool Call 2:}}\\
\think{
Okay, so the user is asking for detailed attributes of the 1937 film "The Life of Emile Zola." Let me start by looking at the search results they provided. The...
}

\search{
\{"name":"open\_page","arguments":\{"docid":"enwiki/44210"\}\}
}

\response{
...
}

\hrule
\vspace{0.5em}

\textbf{\textcolor{green!80!black}{Tool Call 3:}}\\
\think{
Okay, let me start by going through the information provided from the open\_page tool. The user wants detailed attributes for the 1937 film "The Life of Emile Zola." The first...
}

\search{
\{"name":"search","arguments":\{"query":"The Life of Emile Zola 1937 CNC rating FSK rating box office","topk":3\}\}
}

\response{
...
}

\hrule
\vspace{0.5em}

\textbf{\textcolor{green!80!black}{Tool Call 4:}}\\
\think{
Okay, so I need to find detailed attributes for the 1937 film "The Life of Emile Zola." The user is asking for a lot of information: film, color, country of...
}

\search{
\{"name":"open\_page","arguments":\{"docid":"enwiki/44210"\}\}
}

\response{
...
}

\hrule
\vspace{0.5em}

\textbf{\textcolor{green!80!black}{Tool Call 5:}}\\
\think{
Okay, let's see. The user is asking for detailed attributes of the 1937 film "The Life of Emile Zola." They want information on film, color, country of origin, producer, review...
}

\search{
\{"name":"search","arguments":\{"query":"The Life of Emile Zola 1937 box office revenue CNC rating FSK rating Kijkwijzer rating","topk":3\}\}
}

\response{
...
}

\hrule
\vspace{0.5em}

\textbf{\textcolor{green!80!black}{Tool Call 6:}}\\
\think{
Okay, let me tackle this task step by step. The user is asking for detailed attributes of the 1937 film "The Life of Emile Zola." They want information on film,...
}

\search{
\{"name":"search","arguments":\{"query":"The Life of Emile Zola 1937 box office revenue FSK rating CNC rating Kijkwijzer rating","topk":3\}\}
}

\response{
...
}

\hrule
\vspace{0.5em}

\textbf{\textcolor{green!80!black}{Tool Call 7:}}\\
\think{
Okay, let's tackle this query. The user wants detailed attributes for the 1937 film "The Life of Emile Zola." They listed a bunch of specific fields: film, color, country, producer,...
}

\search{
\{"name":"search","arguments":\{"query":"The Life of Emile Zola 1937 box office revenue FSK rating CNC rating Kijkwijzer rating","topk":3\}\}
}

\response{
...
}

\end{tcolorbox}

\begin{tcolorbox}[colback=green!10!white, colframe=green!70!black, boxrule=1.5pt, arc=3mm, left=3mm, right=3mm, boxsep=2mm, breakable, title={\textbf{\textcolor{green!90!black}{Sub Agent: agent\_009 (continued)}}}]

\hrule
\vspace{0.5em}

\textbf{\textcolor{green!80!black}{Tool Call 8:}}\\
\think{
Okay, so the user is asking for detailed attributes of the 1937 film "The Life of Emile Zola." They want information on film, color, country of origin, producer, review score,...
}

\search{
\{"name":"search","arguments":\{"query":"The Life of Emile Zola 1937 box office revenue FSK rating CNC rating Kijkwijzer rating","topk":3\}\}
}

\response{
...
}

\hrule
\vspace{0.5em}

\textbf{\textcolor{green!80!black}{Tool Call 9:}}\\
\think{
Okay, so I need to find detailed attributes for the 1937 film "The Life of Emile Zola." The user is asking for a bunch of specific info: film, color, country...
}

\search{
\{"name":"search","arguments":\{"query":"The Life of Emile Zola 1937 box office revenue FSK rating CNC rating Kijkwijzer rating","topk":3\}\}
}

\response{
...
}

\hrule
\vspace{0.5em}

\textbf{\textcolor{green!80!black}{Response}}\\
\think{
Okay, so I need to retrieve detailed attributes for the 1937 film "The Life of Emile Zola." The user is asking for a list of specific information including film, color,...
}

\agentresponse{
\#\#\# Detailed Attributes for *The Life of Emile Zola* (1937): \\ \\1. **Film**: Biographical drama based on Émile Zola’s life, focusing on the Dreyfus Affair and his literary career. \\2. **Color**:...
}

\end{tcolorbox}

\begin{tcolorbox}[colback=green!10!white, colframe=green!70!black, boxrule=1.5pt, arc=3mm, left=3mm, right=3mm, boxsep=2mm, breakable, title={\textbf{\textcolor{green!90!black}{Sub Agent: agent\_010}}}]

\textbf{Task: }
You have reached the token limit. Please summarize your relevant findings immediately based on existing information.

\hrule
\vspace{0.5em}

\textbf{\textcolor{green!80!black}{Tool Call 1:}}\\
\think{
Okay, I need to retrieve detailed attributes for the movie "The Best Years of Our Lives" (1946). Let me start by breaking down the task. The user is asking for...
}

\search{
\{"name":"search","arguments":\{"query":"The Best Years of Our Lives 1946","topk":10\}\}
}

\response{
...
}

\hrule
\vspace{0.5em}

\textbf{\textcolor{green!80!black}{Tool Call 2:}}\\
\think{
Okay, so the user is asking for detailed attributes of the film "The Best Years of Our Lives" from 1946. Let me start by looking at the search results they...
}

\search{
\{"name":"open\_page","arguments":\{"docid":"enwiki/42159"\}\}
}

\response{
...
}

\hrule
\vspace{0.5em}

\textbf{\textcolor{green!80!black}{Tool Call 3:}}\\
\think{
Okay, so the user wants detailed attributes for "The Best Years of Our Lives" (1946). Let me start by going through the information I have from the search results and...
}

\search{
\{"name":"search","arguments":\{"query":"The Best Years of Our Lives 1946 CNC rating","topk":1\}\}
}

\response{
...
}

\hrule
\vspace{0.5em}

\textbf{\textcolor{green!80!black}{Tool Call 4:}}\\
\think{
Okay, let me see. The user wants detailed attributes for "The Best Years of Our Lives" (1946). They listed a bunch of categories like film, color, country of origin, producer,...
}

\search{
\{"name":"search","arguments":\{"query":"The Best Years of Our Lives 1946 CNC rating","topk":1\}\}
}

\response{
...
}

\hrule
\vspace{0.5em}

\textbf{\textcolor{green!80!black}{Tool Call 5:}}\\
\think{
Okay, let's tackle this query step by step. The user wants detailed attributes for "The Best Years of Our Lives" (1946). They listed a bunch of categories like film, color,...
}

\search{
\{"name":"search","arguments":\{"query":"The Best Years of Our Lives 1946 CNC rating","topk":1\}\}
}

\response{
...
}

\hrule
\vspace{0.5em}

\textbf{\textcolor{green!80!black}{Tool Call 6:}}\\
\think{
Okay, so the user is asking for detailed attributes of the 1946 film "The Best Years of Our Lives." Let me start by going through the information I have from...
}

\search{
\{"name":"search","arguments":\{"query":"The Best Years of Our Lives 1946 CNC rating","topk":1\}\}
}

\response{
...
}

\hrule
\vspace{0.5em}

\textbf{\textcolor{green!80!black}{Tool Call 7:}}\\
\think{
Okay, let's tackle this task step by step. The user wants detailed attributes for the film "The Best Years of Our Lives" (1946). They listed a bunch of specific pieces...
}

\search{
\{"name":"search","arguments":\{"query":"The Best Years of Our Lives 1946 CNC rating","topk":1\}\}
}

\response{
...
}

\end{tcolorbox}

\begin{tcolorbox}[colback=green!10!white, colframe=green!70!black, boxrule=1.5pt, arc=3mm, left=3mm, right=3mm, boxsep=2mm, breakable, title={\textbf{\textcolor{green!90!black}{Sub Agent: agent\_010 (continued)}}}]

\hrule
\vspace{0.5em}

\textbf{\textcolor{green!80!black}{Tool Call 8:}}\\
\think{
Okay, so the user is asking for detailed attributes of the film "The Best Years of Our Lives" from 1946. They listed a bunch of specific pieces of information they...
}

\search{
\{"name":"search","arguments":\{"query":"The Best Years of Our Lives 1946 CNC rating","topk":1\}\}
}

\response{
...
}

\hrule
\vspace{0.5em}

\textbf{\textcolor{green!80!black}{Tool Call 9:}}\\
\think{
Okay, let's tackle this query step by step. The user wants detailed attributes for "The Best Years of Our Lives" (1946). From the previous search results, I found the Wikipedia...
}

\search{
\{"name":"search","arguments":\{"query":"The Best Years of Our Lives 1946 CNC rating","topk":1\}\}
}

\response{
...
}

\hrule
\vspace{0.5em}

\textbf{\textcolor{green!80!black}{Tool Call 10:}}\\
\think{
Okay, so I need to retrieve detailed attributes for "The Best Years of Our Lives" (1946). Let's start by looking at the information I have from the previous searches. First,...
}

\search{
\{"name":"search","arguments":\{"query":"The Best Years of Our Lives 1946 CNC rating","topk":1\}\}
}

\response{
...
}

\hrule
\vspace{0.5em}

\textbf{\textcolor{green!80!black}{Tool Call 11:}}\\
\think{
Okay, so I need to retrieve detailed attributes for "The Best Years of Our Lives" (1946). Let me start by going through the information I have so far. First, the...
}

\search{
\{"name":"search","arguments":\{"query":"The Best Years of Our Lives 1946 CNC rating","topk":1\}\}
}

\response{
reached the token limit
}

\hrule
\vspace{0.5em}

\textbf{\textcolor{green!80!black}{Response}}\\
\think{
Okay, let me try to figure this out. The user is asking for detailed attributes of the 1946 film "The Best Years of Our Lives." They listed a bunch of...
}

\agentresponse{
\#\#\# Summary of *The Best Years of Our Lives* (1946) \\**Film**: *The Best Years of Our Lives* (also known as *Glory for Me* and *Home Again*). \\**Color**: Color. \\**Country of...
}

\end{tcolorbox}

\colorbox{blue!10}{%
    \parbox{\dimexpr\textwidth-2\fboxsep}{%
    \centering
        \textbf{Round 5 (Main Agent + Sub Agents: agent\_011, agent\_012)}
    }
}

\textbf{\textcolor{blue!50!black}{Main Agent}}\\

\think{
Okay, let me start by understanding the user's question again. They want all films that were nominated for the Academy Award for Best Sound and received the National Board of Review: Top Ten Films award, excluding any films that feature Constantin Alexandrov as a cast member or have Anna Hamilton Phelan as a screenwriter. Then, they need specific attributes for each of these films. First, I need to find the intersection of the two lists: Academy Award Best Sound nominees and National Board of Review Top Ten Films. From the previous reports, agent\_003 provided a list of common films between these two categories from the 1930s to 1970s. That list includes films like *It Happened One Night*, *Casablanca*, *The Grapes of Wrath*, *Citizen Kane*, and others. Next, I have to check if any of these films include Constantin Alexandrov as a cast member or Anna Hamilton Phelan as a screenwriter. Agent\_004 checked this and found no evidence of either name in the listed films. So, the initial list seems okay, but maybe I should verify a couple of films just to be thorough, especially if there's any chance of a misspelling or alternative names. However, since the report says no matches, I can proceed. Now, the user wants detailed attributes for each film. Agents 005 to 010 have retrieved information for *It Happened One Night*, *Casablanca*, *The Grapes of Wrath*, *Citizen Kane*, and *The Life of Emile Zola* and *The Best Years of Our Lives*. Each of these reports provides...
}

\search{
\{"name":"create\_sub\_agent","arguments":\{"tasks":[\{"agent\_id":"agent\_011","task":"Retrieve detailed attributes for *The Best Years of Our Lives* (1946): film, color, country of origin, producer, review score, CNC film rating (France), assessment, narrative location, distributed by, original language, composer, distribution format, box office, director, Kijkwijzer rating, production designer, duration, filming location, film editor, production company, FSK film rating."\},\{"agent\_id":"agent\_012","task":"Retrieve detailed attributes for *The Life of Emile Zola* (1937): film, color, country of origin, producer, review score, CNC film rating (France), assessment, narrative location, distributed by, original language, composer, distribution format, box office, director, Kijkwijzer rating, production designer, duration, filming location, film editor, production company, FSK film rating."\}]\}\}
}
\vspace{0.5em}
\response{
Sub-agents for call chatcmpl-tool-989f200e99584922b8e6826a3e5e742c completed. Reports: \\\\ --- Report from agent\_011 --- \\ The Best Years of Our Lives (1946) is a 172-minute American drama film directed by William Wyler. Here are the detailed attributes based on available information: \\\\ 1. **Film**: Drama \\ 2. **Color**: Color (as cinematographer Gregg Toland used deep focus and color photography). \\ 3. **Country of Origin**: United States \\ 4. **Producer**: Samuel Goldwyn \\ 5. **Review Score**: 97\% "Fresh" rating on Rotten Tomatoes (8.9/10 average). \\ 6. **CNC Film Rating (France)**: Not explicitly mentioned in available sources. \\ 7. **Assessment**: Critically acclaimed as a masterpiece, praised for its realistic portrayal of WWII veterans' struggles and ensemble casting. Roger Ebert called it "modern, lean, and honest." \\ 8. **Narrative Location**: Boone City (fictional, based on Cincinnati, Ohio). \\ 9. **Distributed By**: RKO Radio Pictures \\ 10. **Original Language**: English \\ 11. **Composer**: Hugo Friedhofer (with Emil Newman as musical director). \\ 12. **Distribution Format**: Theatrical release (initially limited to six shows per day due to length). \\ 13. **Box Office**: \$23.7 million (adjusted for inflation, one of the top 100 grossing films in U.S. history). \\ 14. **Director**: William Wyler \\ 15. **Kijkwijzer Rating (Dutch)**: Not found in available sources. \\ 16. **Production Designer**: Not explicitly listed in the text, but Wyler emphasized realistic sets and life-size constructions. \\ 17. **Duration**: 172 minutes \\ 18. **Filming Locations**: Los Angeles County Arboretum, Ontario International Airport (California), Raleigh Studios, Samuel Goldwyn/Warner Hollywood Studios. \\ 19. **Film Editor**: Daniel Mandell \\ 20. **Production Company**: Samuel Goldwyn Productions \\ 21. **FSK Film Rating (Germany)**: Not mentioned in available sources. \\\\ **Notes**: \\ - The...
}

\begin{tcolorbox}[colback=green!10!white, colframe=green!70!black, boxrule=1.5pt, arc=3mm, left=3mm, right=3mm, boxsep=2mm, breakable, title={\textbf{\textcolor{green!90!black}{Sub Agent: agent\_011}}}]

\textbf{Task: }
Retrieve detailed attributes for *The Best Years of Our Lives* (1946): film, color, country of origin, producer, review score, CNC film rating (France), assessment, narrative location, distributed by, original language, composer, distribution format, box office, director, Kijkwijzer rating, production designer, duration, filming location, film editor, production company, FSK film rating.

\hrule
\vspace{0.5em}

\textbf{\textcolor{green!80!black}{Tool Call 1:}}\\
\think{
Okay, I need to retrieve detailed attributes for the film "The Best Years of Our Lives" from 1946. Let me start by breaking down the user's query. They want information...
}

\search{
\{"name":"search","arguments":\{"query":"The Best Years of Our Lives 1946 detailed attributes","topk":10\}\}
}

\response{
...
}

\hrule
\vspace{0.5em}

\textbf{\textcolor{green!80!black}{Tool Call 2:}}\\
\think{
Okay, so the user wants detailed attributes for the film "The Best Years of Our Lives" from 1946. Let me start by looking through the search results provided. The first...
}

\search{
\{"name":"open\_page","arguments":\{"docid":"enwiki/42159"\}\}
}

\response{
...
}

\hrule
\vspace{0.5em}

\textbf{\textcolor{green!80!black}{Response}}\\
\think{
Okay, let's start by looking at the user's query. They want detailed attributes for "The Best Years of Our Lives" (1946). The attributes listed include film, color, country of origin,...
}

\agentresponse{
The Best Years of Our Lives (1946) is a 172-minute American drama film directed by William Wyler. Here are the detailed attributes based on available information: \\ \\1. **Film**: Drama \\2....
}

\end{tcolorbox}

\begin{tcolorbox}[colback=green!10!white, colframe=green!70!black, boxrule=1.5pt, arc=3mm, left=3mm, right=3mm, boxsep=2mm, breakable, title={\textbf{\textcolor{green!90!black}{Sub Agent: agent\_012}}}]

\textbf{Task: }
You have reached the token limit. Please summarize your relevant findings immediately based on existing information.

\hrule
\vspace{0.5em}

\textbf{\textcolor{green!80!black}{Tool Call 1:}}\\
\think{
Okay, I need to find detailed attributes for the 1937 film "The Life of Emile Zola." Let's start by breaking down the user's query. They want information on various aspects...
}

\search{
\{"name":"search","arguments":\{"query":"The Life of Emile Zola 1937 film details","topk":10\}\}
}

\response{
...
}

\hrule
\vspace{0.5em}

\textbf{\textcolor{green!80!black}{Tool Call 2:}}\\
\think{
Okay, so the user is asking for detailed attributes of the 1937 film "The Life of Emile Zola." Let me start by looking at the search results they provided. The...
}

\search{
\{"name":"open\_page","arguments":\{"docid":"enwiki/44210"\}\}
}

\response{
...
}

\hrule
\vspace{0.5em}

\textbf{\textcolor{green!80!black}{Tool Call 3:}}\\
\think{
Okay, let's tackle this task. The user wants detailed attributes for the 1937 film "The Life of Emile Zola." First, I need to go through the information provided in the...
}

\search{
\{"name":"open\_page","arguments":\{"docid":"enwiki/4970252"\}\}
}

\response{
...
}

\hrule
\vspace{0.5em}

\textbf{\textcolor{green!80!black}{Tool Call 4:}}\\
\think{
Okay, let's see. The user wants detailed attributes for the 1937 film "The Life of Emile Zola." So far, I've used the search tool and found the Wikipedia page for...
}

\search{
\{"name":"search","arguments":\{"query":"The Life of Emile Zola CNC rating France","topk":1\}\}
}

\response{
...
}

\hrule
\vspace{0.5em}

\textbf{\textcolor{green!80!black}{Tool Call 5:}}\\
\think{
Okay, let's see. The user is asking for detailed attributes of the 1937 film "The Life of Emile Zola." So far, I've used the search and open\_page tools to get...
}

\search{
\{"name":"search","arguments":\{"query":"The Life of Emile Zola CNC rating France","topk":1\}\}
}

\response{
...
}

\hrule
\vspace{0.5em}

\textbf{\textcolor{green!80!black}{Tool Call 6:}}\\
\think{
Okay, let's tackle this step by step. The user wants detailed attributes for the 1937 film "The Life of Emile Zola." So far, I've used the search and open\_page tools...
}

\search{
\{"name":"search","arguments":\{"query":"The Life of Emile Zola FSK rating","topk":1\}\}
}

\response{
...
}

\hrule
\vspace{0.5em}

\textbf{\textcolor{green!80!black}{Tool Call 7:}}\\
\think{
Okay, let me try to figure out how to get all the details for "The Life of Emile Zola" (1937). The user is asking for a bunch of attributes like...
}

\search{
\{"name":"search","arguments":\{"query":"The Life of Emile Zola Kijkwijzer rating","topk":1\}\}
}

\response{
...
}

\end{tcolorbox}

\begin{tcolorbox}[colback=green!10!white, colframe=green!70!black, boxrule=1.5pt, arc=3mm, left=3mm, right=3mm, boxsep=2mm, breakable, title={\textbf{\textcolor{green!90!black}{Sub Agent: agent\_012 (continued)}}}]

\hrule
\vspace{0.5em}

\textbf{\textcolor{green!80!black}{Tool Call 8:}}\\
\think{
Okay, let's tackle this task step by step. The user is asking for detailed attributes of the 1937 film "The Life of Emile Zola." They want information on film, color,...
}

\search{
\{"name":"search","arguments":\{"query":"The Life of Emile Zola CNC rating France","topk":1\}\}
}

\response{
...
}

\hrule
\vspace{0.5em}

\textbf{\textcolor{green!80!black}{Tool Call 9:}}\\
\think{
Okay, let's tackle this task step by step. The user wants detailed attributes for the 1937 film "The Life of Emile Zola." So far, I've used the search and open\_page...
}

\search{
\{"name":"search","arguments":\{"query":"The Life of Emile Zola box office","topk":1\}\}
}

\response{
...
}

\hrule
\vspace{0.5em}

\textbf{\textcolor{green!80!black}{Tool Call 10:}}\\
\think{
Okay, let's tackle this task step by step. The user wants detailed attributes for the 1937 film "The Life of Emile Zola." So far, I've used the search and open\_page...
}

\search{
\{"name":"search","arguments":\{"query":"The Life of Emile Zola FSK rating","topk":1\}\}
}

\response{
...
}

\hrule
\vspace{0.5em}

\textbf{\textcolor{green!80!black}{Tool Call 11:}}\\
\think{
Okay, so I need to retrieve detailed attributes for the 1937 film "The Life of Emile Zola." Let me start by listing all the required information: film, color, country of...
}

\search{
\{"name":"search","arguments":\{"query":"The Life of Emile Zola CNC rating France","topk":1\}\}
}

\response{
reached the token limit
}

\hrule
\vspace{0.5em}

\textbf{\textcolor{green!80!black}{Response}}\\
\think{
Alright, let's tackle this task step by step. The user wants detailed attributes for the 1937 film "The Life of Emile Zola." So far, I've used the search and open\_page...
}

\agentresponse{
\#\#\# Summary of **The Life of Emile Zola** (1937) Attributes: \\1. **Film**: Biographical drama based on the life of 19th-century French author Émile Zola. \\2. **Color**: Black-and-white (as it was...
}

\end{tcolorbox}

\colorbox{blue!10}{%
    \parbox{\dimexpr\textwidth-2\fboxsep}{%
    \centering
        \textbf{Round 6 (Main Agent + Sub Agents: agent\_013, agent\_014)}
    }
}

\textbf{\textcolor{blue!50!black}{Main Agent}}\\

\think{
Okay, so the user is asking for a list of films that were nominated for the Academy Award for Best Sound and also won the National Board of Review: Top Ten Films award, but they need to exclude any films that have Constantin Alexandrov as a cast member or Anna Hamilton Phelan as a screenwriter. Then, they want all those films with specific attributes like color, country of origin, etc. First, I need to make sure I have the correct list of films that meet both awards. From the previous reports, agents 003, 005, 006, 007, 008, 010, and 011 have provided some of these films. Let me check if there's any overlap between the Academy Award Best Sound nominees and the National Board of Review's Top Ten Films. Looking at the reports, agent\_003's list includes films like *It Happened One Night*, *Casablanca*, *The Grapes of Wrath*, *Citizen Kane*, *The Best Years of Our Lives*, and *The Life of Emile Zola*. These are the ones that are common in both lists. Now, I need to check if any of these films include Constantin Alexandrov or Anna Hamilton Phelan. Agent\_004's report says there's no mention of these names in the films listed, so that's a relief. But I should still double-check each film's cast and crew to be thorough. Next, for each film, I need to gather all the specified attributes. Agents 005, 006, 007, 008, 010, and 011 have provided detailed info on these films, but some details are missing...
}

\search{
\{"name":"create\_sub\_agent","arguments":\{"tasks":[\{"agent\_id":"agent\_013","task":"Retrieve detailed attributes for *A Star Is Born* (1937): film, color, country of origin, producer, review score, CNC film rating (France), assessment, narrative location, distributed by, original language, composer, distribution format, box office, director, Kijkwijzer rating, production designer, duration, filming location, film editor, production company, FSK film rating."\},\{"agent\_id":"agent\_014","task":"Retrieve detailed attributes for *The Hurricane* (1937): film, color, country of origin, producer, review score, CNC film rating (France), assessment, narrative location, distributed by, original language, composer, distribution format, box office, director, Kijkwijzer rating, production designer, duration, filming location, film editor, production company, FSK film rating."\}]\}\}
}
\vspace{0.5em}
\response{
Sub-agents for call chatcmpl-tool-0a2be8050ff54e8a8bd2deb5ac4f7056 completed. Reports: \\\\ --- Report from agent\_013 --- \\ \#\#\# Detailed Attributes for *A Star Is Born* (1937) \\ - **Film**: Drama/musical \\ - **Color**: Technicolor \\ - **Country of Origin**: United States \\ - **Producer**: David O. Selznick \\ - **Review Score**: 100\% on Rotten Tomatoes (based on 23 reviews, average 7.7/10). \\ - **CNC Film Rating (France)**: **Not found** in available sources. \\ - **Kijkwijzer Rating (Netherlands)**: **Not found** in available sources. \\ - **FSK Film Rating (Germany)**: **Not found** in available sources. \\ - **Assessment**: Praised as a groundbreaking color film and a powerful human interest drama. Won the Academy Award for Best Original Story. \\ - **Narrative Location**: Hollywood, California. \\ - **Distributed By**: United Artists \\ - **Original Language**: English \\ - **Composer**: Max Steiner \\ - **Distribution Format**: Blu-ray (restored 4K version released in 2022). \\ - **Box Office**: \$2.37 million in worldwide rentals. \\ - **Director**: William A. Wellman \\ - **Production Designer**: **Not explicitly mentioned** in the source material. \\ - **Duration**: 111 minutes \\ - **Filming Location**: Hollywood, California (implied by the plot and production details). \\ - **Film Editor**: James E. Newcom and Anson Stevenson (uncredited). \\ - **Production Company**: Selznick International Pictures \\ - **FSK Rating**: **Not found**; the film’s content likely aligns with general theatrical standards but lacks specific classification. \\\\ \#\#\# Notes: \\ - **CNC/FSK/Kijkwijzer Ratings**: No direct information was found in accessible sources. These ratings may not have been assigned or documented for this film. \\ - **Production Designer**: The Wikipedia page does not specify this detail, and further research beyond the current tools would be required. \\ - **Remakes**:...
}

\begin{tcolorbox}[colback=green!10!white, colframe=green!70!black, boxrule=1.5pt, arc=3mm, left=3mm, right=3mm, boxsep=2mm, breakable, title={\textbf{\textcolor{green!90!black}{Sub Agent: agent\_013}}}]

\textbf{Task: }
Retrieve detailed attributes for *A Star Is Born* (1937): film, color, country of origin, producer, review score, CNC film rating (France), assessment, narrative location, distributed by, original language, composer, distribution format, box office, director, Kijkwijzer rating, production designer, duration, filming location, film editor, production company, FSK film rating.

\hrule
\vspace{0.5em}

\textbf{\textcolor{green!80!black}{Tool Call 1:}}\\
\think{
Okay, I need to find detailed attributes for the 1937 film "A Star Is Born." Let me start by breaking down the task. The user is asking for a lot...
}

\search{
\{"name":"search","arguments":\{"query":"A Star Is Born 1937 detailed attributes","topk":10\}\}
}

\response{
...
}

\hrule
\vspace{0.5em}

\textbf{\textcolor{green!80!black}{Tool Call 2:}}\\
\think{
Okay, let me start by understanding the user's task. They want detailed attributes for the 1937 film "A Star Is Born." The attributes listed include film, color, country of origin,...
}

\search{
\{"name":"open\_page","arguments":\{"docid":"enwiki/3450569"\}\}
}

\response{
...
}

\hrule
\vspace{0.5em}

\textbf{\textcolor{green!80!black}{Tool Call 3:}}\\
\think{
Okay, let's tackle this query. The user wants detailed attributes for the 1937 film "A Star Is Born." First, I need to go through the information provided from the search...
}

\search{
\{"name":"search","arguments":\{"query":"A Star Is Born 1937 FSK rating","topk":1\}\}
}

\response{
...
}

\hrule
\vspace{0.5em}

\textbf{\textcolor{green!80!black}{Tool Call 4:}}\\
\think{
Okay, so I need to find detailed attributes for the 1937 film "A Star Is Born." Let's start by recalling the information I already have from the previous search results....
}

\search{
\{"name":"search","arguments":\{"query":"A Star Is Born 1937 FSK rating","topk":1\}\}
}

\response{
...
}

\hrule
\vspace{0.5em}

\textbf{\textcolor{green!80!black}{Tool Call 5:}}\\
\think{
Okay, so the user is asking for detailed attributes of the 1937 film "A Star Is Born." Let me start by going through the information I have. First, from the...
}

\search{
\{"name":"search","arguments":\{"query":"A Star Is Born 1937 production designer","topk":1\}\}
}

\response{
...
}

\hrule
\vspace{0.5em}

\textbf{\textcolor{green!80!black}{Tool Call 6:}}\\
\think{
Okay, so the user is asking for detailed attributes of the 1937 film "A Star Is Born." Let me start by going through the information I have. First, from the...
}

\search{
\{"name":"open\_page","arguments":\{"docid":"enwiki/3450569"\}\}
}

\response{
...
}

\hrule
\vspace{0.5em}

\textbf{\textcolor{green!80!black}{Response}}\\
\think{
Okay, so the user is asking for detailed attributes of the 1937 film "A Star Is Born." Let me start by going through all the information I have gathered so...
}

\agentresponse{
\#\#\# Detailed Attributes for *A Star Is Born* (1937) \\- **Film**: Drama/musical \\- **Color**: Technicolor \\- **Country of Origin**: United States \\- **Producer**: David O. Selznick \\- **Review Score**: 100\%...
}

\end{tcolorbox}

\begin{tcolorbox}[colback=green!10!white, colframe=green!70!black, boxrule=1.5pt, arc=3mm, left=3mm, right=3mm, boxsep=2mm, breakable, title={\textbf{\textcolor{green!90!black}{Sub Agent: agent\_014}}}]

\textbf{Task: }
Retrieve detailed attributes for *The Hurricane* (1937): film, color, country of origin, producer, review score, CNC film rating (France), assessment, narrative location, distributed by, original language, composer, distribution format, box office, director, Kijkwijzer rating, production designer, duration, filming location, film editor, production company, FSK film rating.

\hrule
\vspace{0.5em}

\textbf{\textcolor{green!80!black}{Tool Call 1:}}\\
\think{
Okay, I need to find detailed attributes for the 1937 film "The Hurricane." Let me start by breaking down the user's query. They want information on various aspects like film,...
}

\search{
\{"name":"search","arguments":\{"query":"The Hurricane 1937 film details","topk":10\}\}
}

\response{
...
}

\hrule
\vspace{0.5em}

\textbf{\textcolor{green!80!black}{Tool Call 2:}}\\
\think{
Okay, so the user is asking for detailed attributes of the 1937 film "The Hurricane." Let me start by looking at the search results they provided. The first result, DocID...
}

\search{
\{"name":"open\_page","arguments":\{"docid":"enwiki/1364234"\}\}
}

\response{
...
}

\hrule
\vspace{0.5em}

\textbf{\textcolor{green!80!black}{Response}}\\
\think{
Okay, let's start by looking at the user's task. They want detailed attributes for the 1937 film "The Hurricane." The attributes include film, color, country of origin, producer, review score,...
}

\agentresponse{
The Hurricane (1937) film details are as follows: \\ \\- **Film**: The Hurricane (1937 film) \\- **Color**: Black-and-white (implied by absence of color mention in sources). \\- **Country of Origin**: United...
}

\end{tcolorbox}

\colorbox{blue!10}{%
    \parbox{\dimexpr\textwidth-2\fboxsep}{%
    \centering
        \textbf{Round 7 (Main Agent)}
    }
}

\textbf{\textcolor{blue!50!black}{Main Agent}}\\

\think{
Okay, let me see. The user wants all films that were nominated for the Academy Award for Best Sound and received the National Board of Review: Top Ten Films award, excluding any films with Constantin Alexandrov as a cast member or Anna Hamilton Phelan as a screenwriter. Then, they need detailed attributes for each of those films. First, I need to find the intersection of the two lists: Academy Award Best Sound nominees and National Board of Review Top Ten Films. From the previous reports, agent\_003 provided a list of common films between the two, such as *It Happened One Night*, *Casablanca*, *The Grapes of Wrath*, *Citizen Kane*, *The Life of Emile Zola*, *The Best Years of Our Lives*, and *A Star Is Born* (1937), *The Hurricane* (1937). Next, I have to check if any of these films include Constantin Alexandrov or Anna Hamilton Phelan. Agent\_004's report said there's no evidence of either name in the listed films. But just to be thorough, maybe I should cross-verify a couple of the films mentioned. For example, *Casablanca* has a stellar cast like Bogart and Bergman, but no Alexandrov. *The Grapes of Wrath* has Henry Fonda, Jane Darwell, etc., also no Alexandrov. *Citizen Kane* has Orson Welles, but again, no mention of the names. So it seems safe to proceed without excluding any films. Now, the next step is to collect all the detailed attributes for each of these films. Agents 005 to 014 have provided information for *It Happened One Night*,...
}

\texttt{\textbf{<answer>}}

\begin{tcolorbox}[
    colback=white,      
    colframe=blue!50!black, 
    width=\linewidth,   
    arc=3mm             
]
    \centering
    \resizebox{\textwidth}{!}{%
    \begin{tabular}{|l|l|l|l|l|l|l|l|l|l|l|l|l|l|l|l|l|l|l|l|l|}
    \hline
    \textbf{Film} & \textbf{Color} & \textbf{Origin} & \textbf{Producer} & \textbf{Score} & \textbf{CNC} & \textbf{Assessment} & \textbf{Narrative Loc} & \textbf{Distributor} & \textbf{Lang} & \textbf{Composer} & \textbf{Format} & \textbf{Box Office} & \textbf{Director} & \textbf{Kijk} & \textbf{Designer} & \textbf{Dur} & \textbf{Filming Loc} & \textbf{Editor} & \textbf{Company} & \textbf{FSK} \\ \hline
    It Happened One Night & B\&W & USA & F. Capra, H. Cohn & 98\%; 9.1; 87 & / & Cultural milestone, 5 Oscars & FL, NYC & Columbia & En & H. Jackson & theatrical & \$2.5m & F. Capra & / & / & 105 & CA & G. Havlick & Columbia & / \\ \hline
    Casablanca & B\&W & USA & Hal B. Wallis & / & / & Cultural icon, 3 Oscars & Morocco & Warner Bros. & En & M. Steiner & theatrical & \$3.7m & M. Curtiz & / & / & 102 & CA & O. Marks & Warner Bros. & / \\ \hline
    The Grapes of Wrath & Color* & USA & D.F. Zanuck, N. Johnson & 96\%; 9/10 & / & Social commentary, 3 Oscars & OK, CA & 20th Century & En & A. Newman & theatrical & \$1.6m & J. Ford & / & / & 129 & OK, CA & R. Simpson & 20th Century & / \\ \hline
    Citizen Kane & B\&W & USA & Orson Welles & / & / & Innovation, 1 Oscar & Xanadu, FL & RKO Radio & En & B. Herrmann & theatrical & \$1.8m & O. Welles & / & P. Ferguson & 119 & CA & R. Wise & RKO Radio & / \\ \hline
    The Life of Emile Zola & B\&W & USA & Henry Blanke & 92\%; 7.5 & / & Historical, 3 Oscars & France & Warner Bros. & En & M. Steiner & theatrical & / & W. Dieterle & / & A. Grot & 116 & USA & W. Low & Warner Bros. & / \\ \hline
    Best Years of Our Lives & Color & USA & Samuel Goldwyn & 97\%; 8.9 & / & Post-war, 7 Oscars & Ohio & RKO Radio & En & H. Friedhofer & theatrical & \$23.7m & W. Wyler & / & / & 172 & CA & D. Mandell & Goldwyn Prod. & / \\ \hline
    A Star Is Born (1937) & Technicolor & USA & David O. Selznick & 100\%; 7.7 & / & Color film & CA & United Artists & En & M. Steiner & theatrical & \$2.37m & W. Wellman & / & / & 111 & CA & J.E. Newcom & Selznick Int. & / \\ \hline
    The Hurricane & B\&W & USA & Samuel Goldwyn & / & / & VFX, 1 Oscar & South Seas & United Artists & En & A. Newman & theatrical & \$3.2m & J. Ford & / & / & 110 & / & L. Nosler & Goldwyn Prod. & / \\ \hline
    \end{tabular}%
    }
\end{tcolorbox}
\begin{lstlisting}[breaklines=true]
### Notes:\n1. **Missing Ratings**: CNC, FSK, and Kijkwijzer ratings are not available in public sources for these films.  \n2. **Production Designer**: For *It Happened One Night*, *Casablanca*, *The Grapes of Wrath*, *The Life of Emile Zola*, *The Best Years of Our Lives*, *A Star Is Born* (1937), and *The Hurricane* (1937), specific names were not explicitly listed in sources.  \n3. **Box Office**: *The Grapes of Wrath* and *The Life of Emile Zola* lack precise figures, but *The Best Years of Our Lives* is noted as a top-grossing film.  \n4. **Color**: *A Star Is Born* (1937) is confirmed as Technicolor; others are assumed black-and-white unless noted (e.g., *The Grapes of Wrath* is color).  \n5. **Exclusions**: No films listed include Constantin Alexandrov or Anna Hamilton Phelan.
\end{lstlisting}
\texttt{\textbf{</answer>}}
\end{tcolorbox}
\end{document}